\documentclass{article}
\usepackage{arxiv}

\usepackage[utf8]{inputenc} 
\usepackage[T1]{fontenc}    
\usepackage{url}            
\usepackage{booktabs}       
\usepackage{amsfonts}       
\usepackage{nicefrac}       
\usepackage{microtype}      
\usepackage{lipsum}
\usepackage{cellspace}
\usepackage{alltt}
\usepackage{calc}
\usepackage{mathtools}
\usepackage{booktabs}
\usepackage{url}
\usepackage{caption}
\usepackage{longtable}
\usepackage{amsmath}
\usepackage{amsthm}
\usepackage{amssymb}
\usepackage{bm}
\usepackage{adjustbox}
\usepackage{wrapfig}
\usepackage{epstopdf}
\usepackage{tikz}
\usepackage{framed}
\usepackage{thmtools,thm-restate}
\usepackage{enumitem}
\usepackage{algorithmic}
\usepackage{algorithm}
\usepackage{soul}
\usepackage{tcolorbox}
\usepackage{multirow}
\usepackage{array}
\usepackage{makecell}
\usepackage[colorlinks=true,breaklinks]{hyperref}
\usepackage{cleveref}

\usetikzlibrary{arrows}
\usetikzlibrary{shapes,backgrounds}

\graphicspath{{./Images/}}

\newtheorem{theorem}{Theorem}
\newtheorem*{theorem*}{Theorem}

\newtheorem*{lemma*}{Lemma}
\newtheorem{lemma}[theorem]{Lemma}

\newtheorem{corollary}[theorem]{Corollary}

\newtheorem{definition}{Definition}
\makeatletter
\renewcommand{\th@definition}{%
  \normalfont
  \thm@preskip-2 \relax
  \thm@postskip-2 \relax
}
\makeatother

\title{A survey on domain adaptation theory:\\learning bounds and theoretical guarantees}

\author{Ievgen Redko, Emilie Morvant, Amaury Habrard, Marc Sebban\\
  Univ Lyon, UJM-Saint-Etienne, CNRS, Institut d'Optique Graduate School\\
  Laboratoire Hubert Curien UMR 5516, F-42023, Saint-Etienne, France\\
  \url{name.surname@univ-st-etienne.fr}
  \AND
  Youn{\`e}s Bennani\\
  Universit\'e Sorbonne Paris Nord, CNRS, Institut Galilée\\
  Laboratoire d'Informatique de Paris Nord UMR 7030, F-93430, Villetaneuse, France\\
  \url{name.surname@sorbonne-paris-nord.fr}}

\DeclareMathOperator{\tr}{tr}

\newcommand{\Acal}{{\cal A}}
\newcommand{\wbf}{{\bf w}}

\newcommand{\R}{\mathbb{R}}
\newcommand{\Hcal}{\mathcal{H}}
\newcommand{\hdh}{{\Hcal\!\Delta\!\Hcal}}
\newcommand{\X}{{\bf X}}
\newcommand{\Y}{Y}
\newcommand{\XY}{\X\times\Y}
\newcommand{\ms}{{m_S}}
\newcommand{\mt}{{m_T}}

\newcommand{\vect}[1]{\mathbf{#1}}
\newcommand{\mat}[1]{\mathbf{#1}}

\newcommand{\xbf}{\vect{x}}
\newcommand{\zbf}{\vect{z}}

\newcommand{\xb}{\xbf}

\newcommand{\target}{{\cal T}}
\newcommand{\source}{{\cal S}}
\newcommand{\sourceX}{{\source}_{\X}}

\newcommand{\targetX}{{\target}_{\X}}

\newcommand{\hatsourceX}{{\hat{\source}}_{\X}}
\newcommand{\hattargetX}{{\hat{\target}}_{\X}}

\newcommand{\TminusS}{{\target{\setminus}\source}}
\newcommand{\scriptsupport}{\mbox{\scriptsize\sc supp}}
\newcommand{\support}{\mbox{\small\sc supp}}
\newcommand{\task}{t}
\newcommand{\targettask}{\task_\target}
\newcommand{\sourcetask}{\task_\source}

\newcommand{\D}{{\cal D}}
\newcommand{\DX}{{\D}_{\X}}

\newcommand{\posterior}{\rho}
\newcommand{\prior}{\pi}

\newcommand{\risk}{{\rm R}}
\newcommand{\loss}{\ell}
\newcommand{\zoloss}{{\loss_{01}}}

\newcommand{\Remp}{\hat{\risk}}
\newcommand{\RD}{\risk_{\D}}
\newcommand{\RS}{\risk_{\source}}

\newcommand{\RT}{\risk_{\target}}

\newcommand{\RSemp}{\risk_{\hat{\source}}}
\newcommand{\RTemp}{\risk_{\hat{\target}}}

\newcommand{\sgn}[1]{\operatorname{sgn}\left(#1\right)}
\newcommand{\err}[3]{\epsilon^{#1}_{\mathcal #2}(#3)}

\newcommand{\dhdh}[2]{d_\hdh(#1,#2)}
\newcommand{\discl}[3]{disc_{#1}(#2,#3)}
\newcommand{\disc}{disc}

\DeclareMathOperator*{\KLop}{KL}
\newcommand{\KL}[2]{\KLop(#1|#2)}
\DeclareMathOperator*{\disPBop}{dis_{\posterior}}

\newcommand{\disPB}[2]{\disPBop(#1,#2)}

\newcommand{\bq}{\beta_q}
\newcommand{\binf}{\beta_{\infty}}
\newcommand{\BQ}{B_{\posterior}}
\newcommand{\GQ}{G_{\posterior}}
\newcommand{\dq}{d_{\DX}(\posterior)}

\newcommand{\dqS}{d_{\sourceX}(\posterior)}
\newcommand{\dqT}{d_{\targetX}(\posterior)}

\newcommand{\dqTemp}{d_{T}(\posterior)}
\newcommand{\eq}{e_{\D}(\posterior)}

\newcommand{\eqS}{e_{\source}(\posterior)}
\newcommand{\eqT}{e_{\target}(\posterior)}

\newcommand{\eqSemp}{e_{S}(\posterior)}

\newcommand{\I}[1]{\ensuremath{\mathbf{I}\left[#1\right]}}
\newcommand{\eqdef}{=} 
\newcommand{\transpose}[1]{{#1}^{\top}}
\newcommand{\sign}[1]{\ensuremath{\mathrm{sign}\left[#1\right]}}
\newcommand{\card}[1]{\left|#1\right|}
\newcommand{\trunc}{\ensuremath{\mathrm{tr}}}

\newcommand{\argmin}[1]{
    \underset{#1}{\mathrm{argmin}}\
}

\newcommand{\ie}{\textit{i.e.}}
\newcommand{\iid}{\textit{\it i.i.d.}}
\newcommand\norm[1]{\left\lVert#1\right\rVert}
\newcommand\abs[1]{\left\lvert#1\right\rvert}

\newcommand{\argmindevant}[1]{
    {\mathrm{argmin}}_{\,#1}\
}

\newcommand{\prob}[1]{
    \underset{#1}{\mathrm{\bf Pr}}\
}
\newcommand{\Prob}[1]{\prob{#1}}
\renewcommand{\Pr}[1]{\prob{#1}}

\newcommand{\esp}[1]{
    \underset{#1}{\mathrm{\bf E}}\
}

\newcommand{\espdevant}[1]{
    \mathrm{\mathrm{\bf E}}_{#1}\,
}

\newcommand{\eg}{{\it e.g.}}

\newcommand{\M}{\mathbf{M}}

\newcommand{\MS}{{\mathbf{M}_\mathcal{S}}}

\begin{document}
\maketitle

\begin{abstract}
All famous machine learning algorithms that comprise both supervised and semi-supervised learning work well only under a common assumption: the training and test data follow the same distribution.
When the distribution changes, most statistical models must be reconstructed from new collected data, which for some applications can be costly or impossible to obtain. 
Therefore, it has become necessary to develop approaches that reduce the need and the effort to obtain new labeled samples by exploiting data that are available in related areas, and using these further across similar fields. This has given rise to a new machine learning framework known as \textit{transfer learning}: a learning setting inspired by the capability of a human being to extrapolate knowledge across tasks to learn more efficiently.
Despite a large amount of different transfer learning scenarios, the main objective of this survey is to provide an overview of the state-of-the-art theoretical results in a specific, and arguably the most popular, sub-field of transfer learning, called \textit{domain adaptation}. In this sub-field, the data distribution is assumed to change across the training and the test data, while the learning task remains the same. We provide a first up-to-date description of existing results related to domain adaptation problem that cover learning bounds based on different statistical learning frameworks.  
\end{abstract}

\keywords{Transfer learning \and Domain adaptation \and Learning theory}

\begin{tcolorbox}
This survey is a shortened version of the recently published book \textbf{"Advances in Domain Adaptation Theory"} \cite{redko:hal-02286281} written by the authors of this survey. Its purpose is to provide a high-level overview of the book and to update it with some recent references. All of the proofs and most of the mathematical developments are omitted in this version, to keep the document to a reasonable length. For more details, we refer the interested reader to the original papers or to the full version of the book, available at \url{https://www.elsevier.com/books/advances-in-domain-adaptation-theory/redko/978-1-78548-236-6}.
\end{tcolorbox}

\section{Introduction}
The idea behind \emph{transfer learning} is inspired by the ability of human beings to learn with minimal or no supervision based on previously acquired knowledge. It is not surprising that this concept was not invented in the machine-learning community in the correct sense of the term, as the concept of "transfer of learning" had been used long before the construction of the first computer, and can be found in papers in the field of psychology from the early 20th century. From the statistical point of view, this learning scenario is different from supervised learning, as transfer learning does not assume that the training and test data have to be drawn from the same probability distribution. It was argued that this assumption is often too restrictive to hold in practice, as in many real-world applications a hypothesis is learned and deployed in environments that differ and exhibit an important shift.
A typical example often used in transfer learning is to consider a spam-filtering task where the spam filter is learned using an arbitrary classification algorithm for a corporate mailbox of a given user. In this case, the vast majority of the e-mails analyzed by the algorithm are likely to be of a professional character, with very few of them being related to the private life of the person considered. Imagine further a situation where this same user installs mailbox software on the personal computer and imports the settings of its corporate mailbox, with the hope that it will work equally well on this too.
However, this is not likely to be the case, as many personal e-mails may appear to be spam to an algorithm that has learned purely on professional communications, due to the differences in their content and attached files, as well as the nonuniformity of e-mail addresses. Another illustrative example is that of species classification in oceanographic studies, where experts rely on video coverage of a certain sea area to recognize species of the marine habitat. For instance, in the Mediterranean Sea and in the Indian Ocean, the species of fish that can be found on the recorded videos are likely to belong to the same family, even though their actual appearance might be quite dissimilar due to the different climate and evolutionary backgrounds. In this case, the learning algorithm trained on the video coverage of the Mediterranean Sea will most likely fail to provide correct classification of species in the Indian Ocean without being specifically adapted by an expert. 

For these kinds of applications, it might be desirable to find a learning paradigm that can remain robust to a changing environment and can adapt to a new problem at hand, by drawing parallels and exploiting the knowledge from the domain where it was learned initially.
In response to this problem, the quest for new algorithms that can learn on a training sample and then provide good performance on a test sample from a different, but related, probability distribution gave rise to a new learning paradigm, known as \textit{transfer learning}.
Its definition is given as follows.
\begin{definition}{(Transfer learning)}
\label{def: transfer_learning}
We consider a source data distribution $\source$ called the source domain, and a target data distribution $\target$ called the target domain. 
Let $\X_\source\times \Y_\source$ be the source input and output spaces associated to $\source$, and $\X_\target\times \Y_\target$ be the target input and output spaces associated to $\target$.
We use $\sourceX$ and $\targetX$ to denote the marginal distributions of $\X_\source$ and $\X_\target$, $\sourcetask$ and $\targettask$ to denote the source and target learning tasks depending on $\Y_\source$ and $\Y_\target$, respectively.
Then, transfer learning aims to help to improve the learning of the target predictive function $f_\target:\X_\target\to\Y_\target$ for $\targettask$ using knowledge gained from $\source$ and $\sourcetask$, where $\source \neq \target$. 
\end{definition}
Note that the condition $\source \neq \target$ implies either $\sourceX \neq \targetX$ (\ie, $\X_\source \neq \X_\target$ or $\sourceX(\X) \neq \targetX(\X)$) or $\sourcetask \neq \targettask$ (\ie, $\Y_\source \neq \Y_\target$ or $\source(\Y\vert\X) \neq \target(\Y\vert\X)$).
\begin{figure}[!t]
\centering
\includegraphics[width=0.75\textwidth]{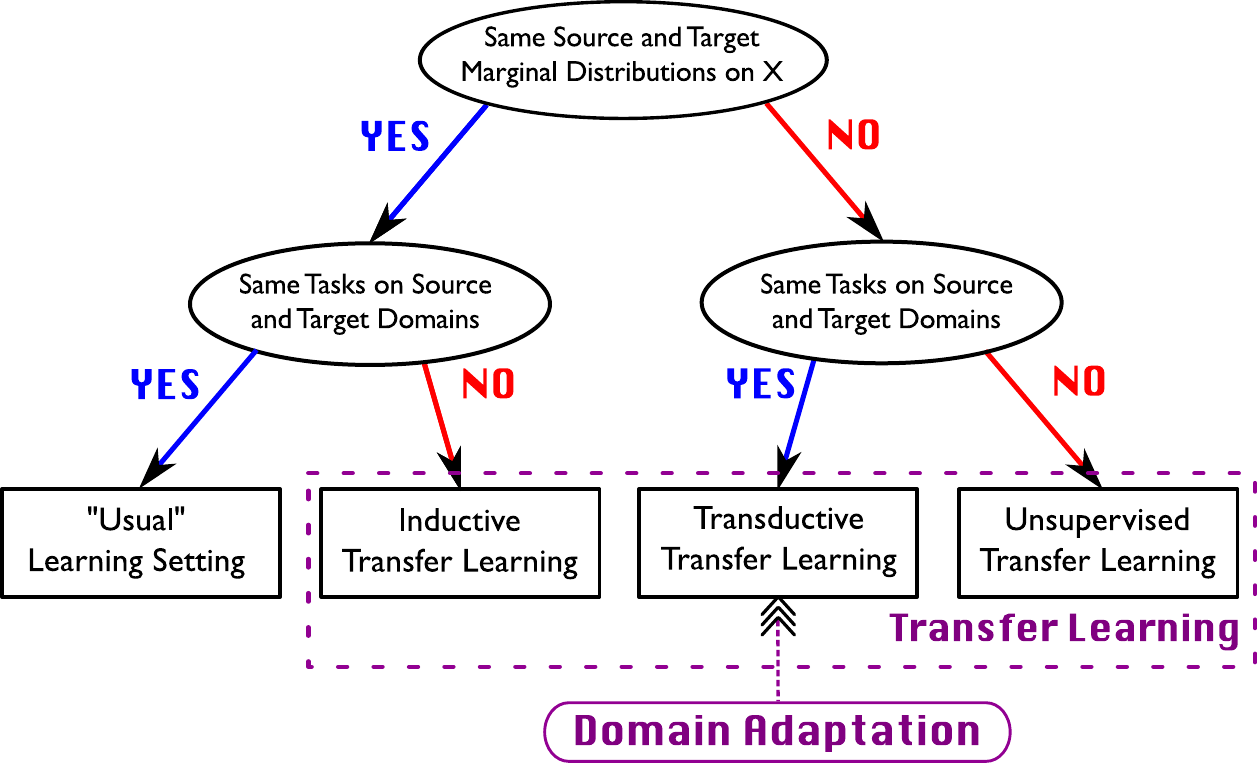}
\caption{Comparison of standard supervised learning, transfer learning, and positioning of the domain adaptation.\label{img:tl}}
\end{figure}
In transfer learning, three possible learning settings are often distinguished based on these different relationships (illustrated in Figure~\ref{img:tl}):
\begin{enumerate}
\item \textbf{Inductive transfer learning} where $\sourceX = \targetX$ and $\sourcetask \neq \targettask$;\\
\textit{For example, $\sourceX$ and $\targetX$ are the distributions of the data collected from the mailbox of one particular user, where $\sourcetask$ is the task of detecting spam, while $\targettask$ is the task of detecting a hoax};
\item \textbf{Transductive transfer learning} where $\sourceX \neq \targetX$ but $\sourcetask = \targettask$;\\
\textit{For example, in the spam filtering problem, $\sourceX$ is the distribution of the data collected for one user, $\targetX$ is the distribution of the data of another user, and $\sourcetask$ and $\targettask$ are both the task of detecting spam};
\item \textbf{Unsupervised transfer learning} where $\sourcetask \neq \targettask$ and $\sourceX \neq \targetX$;\\
 \textit{For example, $\sourceX$ generates the data collected from one user and $\targetX$ generates the content of web-pages collected on the web, where $\sourcetask$ is to filter out spams, while $\targettask$ is to detect hoaxes}.
\end{enumerate}
Arguably, the vast majority of situations where transfer learning is most needed fall into the second category. This second category has the name of \emph{domain adaptation}, where we suppose that the source and the target tasks are the same, but where we have a source dataset with an abundant amount of labeled observations and a target dataset with no (or few) labeled instances. In this survey, we concentrate on theoretical advances related to the latter case, and we highlight their differences with respect to the traditional supervised learning paradigm. A brief overview of the contributions presented is given in Tables 1 and 2 for learning bounds and hardness results, respectively.

\setlength{\aboverulesep}{0pt}
\setlength{\belowrulesep}{0pt}
\setlength\LTleft{0pt}
\setlength\LTright{0pt}
\begin{longtable}{@{\extracolsep{\fill}}>{\footnotesize}c>{\footnotesize}c>{\footnotesize}c>{\footnotesize}c>{\footnotesize}c>{\footnotesize}c>{\footnotesize}c}
\caption{Summary of the learning bounds presented in this survey for domain adaptation. \textbf{(Task)} refers to the considered learning problem; \textbf{(Framework)} specifies the statistical learning framework used in the analysis; \textbf{(Divergence)} is the metric used to compare the source and target distributions; \textbf{(Link)} represents the dependence between the source error and the divergence term; \textbf{(Non-estim.)} indicates the presence of a nonestimable term in the bounds.} 
\endfirsthead
\endhead
\toprule
\addlinespace[.1cm]
\multirow{3}{*}{\footnotesize{\textbf{\textsc{Reference}}}}& \multicolumn{5}{c}{\footnotesize{\textbf{\textsc{Learning bounds}}}}\\
\addlinespace[.1cm]\cline{2-6}
\addlinespace[.1cm]
& \textbf{\textsc{Task}} & \textbf{\textsc{Framework}} & \textbf{\textsc{Divergence}} &	\textbf{\textsc{Link}} & \textbf{\textsc{Non-estim.}}\\
\addlinespace[.1cm]
\toprule
\addlinespace[.1cm]
\makecell{\cite{BenDavid07analysisof}\\\cite{Blitzer07Learning}\\\cite{bendavidth}} & \makecell{Binary\\ classification} & VC & $L^1$, $\hdh$ & Add. & $+$\\
\hline
\addlinespace[.1cm]
\cite{MansourMR09colt} & \makecell{Classification/\\ Regression} & Rademacher & Discrepancy & Add. & $+$\\
\addlinespace[.1cm]
\hline
\addlinespace[.1cm]
\cite{KurokiCBHSS19} & \makecell{Classification} & Rademacher & (S-)Discrepancy & Add. & $+$\\
\addlinespace[.1cm]
\hline
\addlinespace[.1cm]
\makecell{\cite{CortesMM10}\\\cite{CortesM14}\\\cite{CortesMM15}} & Regression & Rademacher & \makecell{(Generalized)\\ Discrepancy} & Add. & $+$\\
\hline
\addlinespace[.1cm]
\cite{MansourMR08multiple} & \makecell{Classification/\\ Regression} & -- & -- & -- & --
\\
\addlinespace[.1cm]
\hline
\addlinespace[.1cm]
\makecell{\cite{MansourMR09}\\\cite{HoffmanMZ18}} & \makecell{Classification/\\ Regression} & -- & R\'enyi & Mult. & --\\
\addlinespace[.1cm]
\hline
\addlinespace[.1cm]
\cite{DhouibR18} & \makecell{Binary classification/\\Similarity learning} & -- & $L^1$, $\chi^2$ & Mult. & $+$\\
\addlinespace[.1cm]
\hline
\addlinespace[.1cm]
\cite{RedkoCFT19} & Binary classification & Rademacher & Discrepancy & Add. & $+$\\
\addlinespace[.1cm]
\hline
\addlinespace[.1cm]
\cite{ZhangZY12} & \makecell{Regression/ \\ Classification} & \makecell{Uniform \\entropy number} & IPM & Add. & --\\
\addlinespace[.1cm]
\hline
\addlinespace[.1cm]
\cite{Redko15} & Regression & Rademacher & IPM/MMD & Add. & $+$\\
\addlinespace[.1cm]
\hline
\addlinespace[.1cm]
\cite{RedkoHS16} & Regression & -- & IPM/Wassertein & Add. & $+$\\
\addlinespace[.1cm]
\hline
\addlinespace[.1cm]
\cite{zhang_bridging_2019} & \makecell{Large-margin\\classification} & Rademacher & IPM & Add. & $+$\\
\addlinespace[.1cm]
\hline
\addlinespace[.1cm]
\cite{dhouib_margin_2020} & \makecell{Large margin\\Binary classification} & -- & \makecell{IPM/minimax Wasserstein} & Add. & +\\
\addlinespace[.1cm]
\hline
\addlinespace[.1cm]
\cite{JohanssonSR19} & \makecell{Classification} & -- & IPM & Add. & $+$\\
\addlinespace[.1cm]
\hline
\addlinespace[.1cm]
\cite{ShenQZY18} & Classification & -- & Wasserstein & Add. & $+$\\
\addlinespace[.1cm]
\hline
\addlinespace[.1cm]
\cite{CourtyFHR17} & Classification & -- & Wasserstein & Add. & $+$\\
\addlinespace[.1cm]
\hline
\addlinespace[.1cm]
\cite{pbda} & \makecell{Classification} & PAC-Bayes & Domain disagreement & Add. & $+$\\
\addlinespace[.1cm]
\hline
\addlinespace[.1cm]
\cite{dalc} & \makecell{Classification} & PAC-Bayes & $\beta$-divergence & Mult. & $+$\\
\addlinespace[.1cm]
\hline
\addlinespace[.1cm]
\cite{li2007bayesian} & Classification & PAC-Bayes & -- & Add. & --\\
\addlinespace[.1cm]
\hline
\addlinespace[.1cm]
\cite{McNamaraB17} & \makecell{Binary classification} & VC/PAC-Bayes & -- & Add. & --\\
\addlinespace[.1cm]
\hline
\addlinespace[.1cm]
\cite{MansourS14} & Classification & Robustness & $\lambda$-shift & Add. & --\\
\addlinespace[.1cm]
\hline
\addlinespace[.05cm]
\makecell{\cite{KuzborskijO13}\\\cite{KuzborskijO17}\\\cite{simonHTL}} & Regression & Stability & -- & -- & --\\
\addlinespace[.05cm]
\hline
\addlinespace[.1cm]
\cite{PerrotH15ICML} & \makecell{Classification/\\Similarity learning} & Stability & -- & -- & --\\
\addlinespace[.1cm]
\hline
\addlinespace[.1cm]
\cite{MorvantKAIS12} & \makecell{Classification/\\Similarity learning} & Robustness/VC & $\hdh$ & Add. & +\\
\bottomrule
\end{longtable}
\begin{longtable}{@{\extracolsep{\fill}}>{\footnotesize}c>{\footnotesize}c>{\footnotesize}c>{\footnotesize}c>{\footnotesize}c>{\footnotesize}c>{\footnotesize}c}
\caption{Summary of the contributions presented in this survey for hardness results in domain adaptation. \textbf{(Type)} is the type of result obtained; \textbf{(Setting)} indicates the presence or absence of target data (either labelled or unlabelled); \textbf{(Assumptions)} indicates the assumptions considered (individual or combined); \textbf{(Proper)} specifies whether the learned model is required to belong to a predefined class; \textbf{(Constr.)} indicates whether the result is of a constructive nature.} 
\endfirsthead
\endhead
\toprule
\addlinespace[.1cm]
\multirow{3}{*}{\footnotesize{\textbf{\textsc{Reference}}}}& \multicolumn{5}{c}{\footnotesize{\textbf{\textsc{Hardness results}}}}\\
\addlinespace[.1cm]\cline{2-6}
\addlinespace[.1cm]
& \textbf{\textsc{Type}} & \textbf{\textsc{Setting}} & \textbf{\textsc{Assumptions}} &	\textbf{\textsc{Proper}} & \textbf{\textsc{Constr.}}\\
\addlinespace[.1cm]
\toprule
\addlinespace[.1cm]
\cite{BenDavidLLP10} & \makecell{Impossibility/\\Sample compl.} & Unlabelled target & \makecell{Cov. shift,\\$\hdh$, $\lambda_\Hcal$} & -- & $+$\\
\addlinespace[.1cm]
\hline
\addlinespace[.1cm]
\cite{BenDavidSU12} & \makecell{Impossibility/\\Sample compl.} & \makecell{No target/\\Unlabelled target} & \makecell{Cov. shift,\\$C_{\cal B}$, Lipscht.} & $+$ & $+$/--\\
\addlinespace[.1cm]
\hline
\addlinespace[.1cm]
\cite{BenDavidU12} & \makecell{Impossibility/\\Sample compl.} & \makecell{Unlabelled target} & \makecell{Cov. shift,\\$C_{\cal B}$, Realizab.} & -- & --\\
\addlinespace[.1cm]
\hline
\addlinespace[.1cm]
\cite{RedkoHS19} & \makecell{Estimation/\\Sample compl.} & \makecell{Labelled target} & -- & -- & --\\
\addlinespace[.1cm]
\hline
\addlinespace[.1cm]
\cite{CZG19} & \makecell{Impossibility} & \makecell{Unlabelled target} & \makecell{Cov. shift,\\$\hdh$, $\lambda_\Hcal$} & -- & $+$\\
\addlinespace[.1cm]
\hline
\addlinespace[.1cm]
\cite{JohanssonSR19} & \makecell{Impossibility} & \makecell{Unlabelled target} & \makecell{Cov. shift,\\$\hdh$, $\lambda_\Hcal$} & -- & $+$\\
\addlinespace[.1cm]
\hline
\addlinespace[.1cm]
\cite{hanneke19} & \makecell{Sample compl.} & \makecell{Labelled target} & \makecell{Relaxed cov. shift,\\Noise cond.} & -- & --\\
\bottomrule
\end{longtable}

The rest of this survey is organized as follows. In \Cref{chap:sota}, we briefly present the traditional statistical learning frameworks that are referred to throughout the survey. In \Cref{chap:div_based}, we present the first theoretical results of the domain adaptation theory from the seminal studies of \cite{BenDavid07analysisof,MansourMR09colt,CortesM11} that rely on the famous $\hdh$ and discrepancy distances. We further turn our attention to hardness results for the domain adaptation problem in \Cref{chap:imposs}. \Cref{chap:ipms} presents several studies that establish the generalization bounds for domain adaptation based on the popular integral probability metrics (IPMs). In \Cref{chap:dalc}, we highlight several learning bounds defined using the PAC-Bayesian framework. Finally, in \Cref{chap:algorithmic}, we give an overview of the contributions that take the actual learning algorithm into account when deriving the learning bounds, and we conclude the survey in \Cref{chap:conclusions}.

\section{Preliminary knowledge}
\label{chap:sota}
Below we recall the usual supervised learning set-up and the different quantities used to derive generalization bounds in this context. This includes the concepts of Vapnik-Chervonenkis (VC) \cite{vapnik2006estimation,VapnikC71} and Rademacher complexities ~\cite{KoltchinskiiP99}, the definitions related to the PAC-Bayesian theory~\cite{McAllester99}, and those from the more recent algorithmic stability~\cite{BousquetE02} and algorithmic robustness~\cite{XuM10} frameworks.

\subsection{Definitions}
Let a pair $(\X,\Y)$ define the input and the output spaces where $\X$ is described by real-valued vectors of finite dimension $d$, \ie, $\X \subseteq \mathbb{R}^d$, and for $\Y$ we distinguish between two possible scenarios: 1) when $\Y$ is continuous, \eg, $ \Y = \left[ -1, 1 \right] $ or $\Y = \mathbb{R}$, we talk about regression;
2) when $\Y$ is discrete and takes values from a finite set, we talk about classification.
Two important cases of classification are binary classification and multi-class classification, where $\Y = \left\lbrace -1, 1\right\rbrace $ (or  $\Y=\left\lbrace 0, 1\right\rbrace$) and $\Y = \left\lbrace 1, 2, \dots, C \right\rbrace $ with $C > 2$, respectively.

We assume that $\X\times\Y$ is drawn from an unknown joint probability distribution $\D$ and that we observe them through a finite training sample (also called the \textit{learning sample}) $S =\{(\xbf_i,y_i)\}_{i=1}^m \sim (\D)^m$ of $m$ independent and identically distributed (\iid) pairs (also called examples or data instances). We further use $\Hcal = \lbrace  h \vert h: \X \rightarrow \Y \rbrace$ to denote a \textit{hypothesis space} (also called the \textit{hypothesis class}) that consists of functions that map each element of $\X$ to $\Y$. These functions $h$ are usually called hypotheses, or more specifically classifiers or regressors, depending on the nature of $\Y$.

Let us now consider a loss function $\loss: \Y \times \Y \rightarrow [0, 1]$ that gives a cost of $h(\xbf)$ deviating from the true output $y\in\Y$. We can define the \textit{true risk} and the empirical risk with respect to $\D$ and $S$, respectively, as follows.
\begin{definition}{(True risk)}
Given a loss function $\loss: \Y \times \Y \rightarrow [0, 1]$, the true risk (also called the generalization error) $\risk^\loss_{\D}(h)$ 
for a given hypothesis $h \in \Hcal$ on a distribution $\D$ over $\X\times\Y$ is defined as
\begin{align*}
 \risk^\loss_{\D}(h)\ =\ \esp{(\xbf,y)\sim \D} \loss(h(\xbf),y).
\end{align*}
By abuse of notations, for a given pair of hypotheses $(h,h') \in \Hcal^2$, we can write
\begin{align*}
 \risk^\loss_{\D}(h,h')\ =\ \esp{(\xbf,y)\sim \D} \loss(h(\xbf),h'(\xbf)).
\end{align*}
\label{def:true_risk}
\end{definition}
\begin{definition}{(Empirical risk)}
Given a loss function $\loss: \Y \times \Y \rightarrow [0, 1]$ and a training sample $S =\{(\xbf_i,y_i)\}_{i=1}^m$, where each example is drawn \iid\ from $\D$, the empirical risk $\risk^\loss_{\hat{\D}}(h)$ for a given hypothesis $h \in \Hcal$ is defined as
\begin{align*}
 \risk^\loss_{\hat{\D}}(h) = \frac{1}{m} \sum_{i=1}^m \loss(h(\xbf_i),y_i)\,,
 \end{align*}
 where $\hat{\D}$ is the empirical distribution associated to the sample $S$.
\label{def:emp_risk}
\end{definition}
\setlength{\intextsep}{0pt}%
The most natural loss function that can be used to count the number of errors committed by hypothesis $h\in \Hcal$ on the distribution $\D$ is the $0-1$ loss function $\loss_{0-1}: \Y \times \Y \rightarrow \{0,1\}$, which is defined for a training example $(\xbf,y)$ as
\begin{align}
\zoloss(h(\xbf),y) &= \mathbf{I}\left[h(\xbf) \neq y\right]  =\left\{
                \begin{array}{ll}
                  1, \ \text{if} \ h(\xbf)\neq y\,,\\
                  0 ,\ \text{otherwise}.
                \end{array}
              \right. \label{eq:zeroone}
\end{align} 
\begin{wrapfigure}[15]{r}{0.3\textwidth}
\centering
\includegraphics[width=\linewidth]{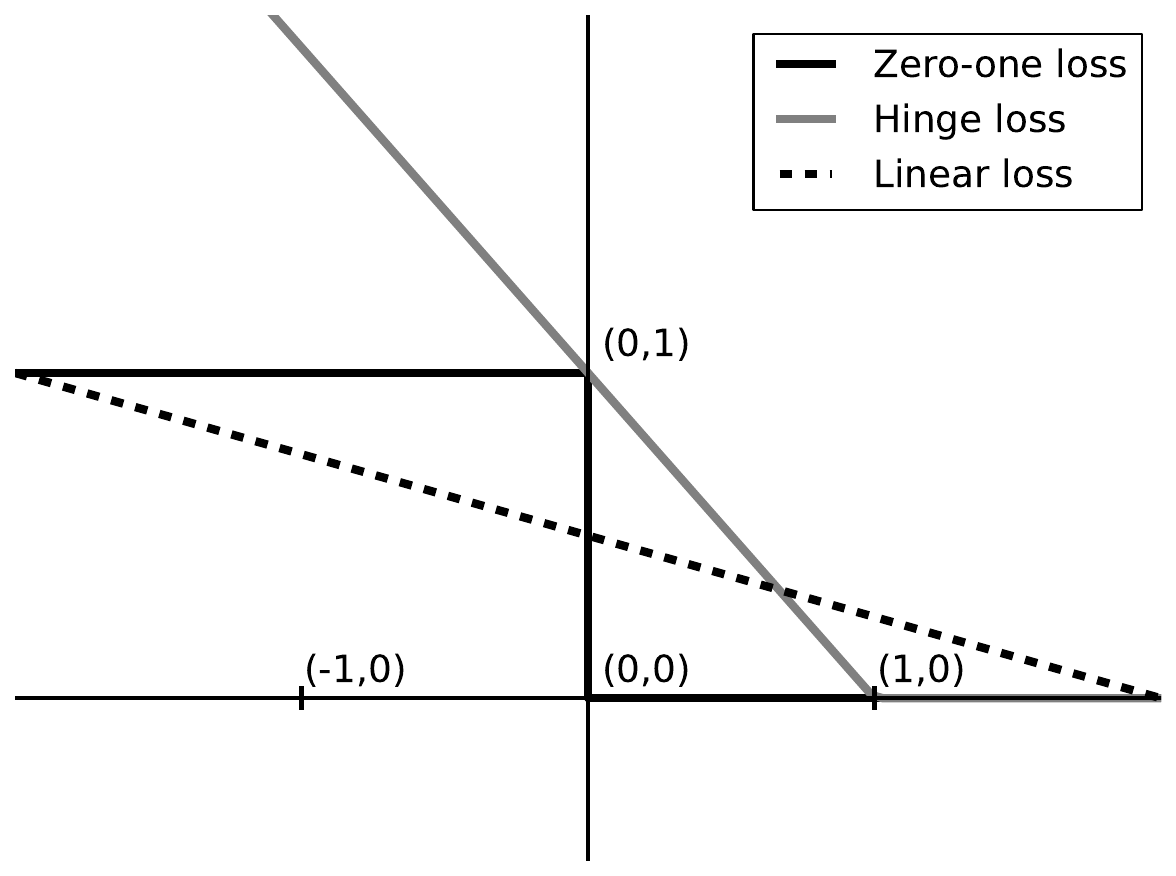}
\caption{\label{fig:loss}Illustration of different loss functions.}
\end{wrapfigure}
A popular proxy to this nonconvex function is the hinge loss defined for a given pair $(\xbf,y)$ by
\begin{align*}
\loss_{\text{hinge}}(h(\xbf),y) &= \left[1-yh(\xbf) \right]_+ = \max\left(0,1-yh(\xbf)\right).
\end{align*}
Another loss function often used in practice that extends the $0-1$ loss to the case of real values is the linear loss $\loss_{\text{lin}}:\mathbb{R}\times \mathbb{R} \rightarrow [0,1]$, defined by:
\begin{align*}
\loss_{\text{lin}}(h(\xbf),y) = \frac{1}{2}\left(1-yh(\xbf)\right).
\end{align*}
The three above-mentioned loss functions are illustrated in Figure~\ref{fig:loss}. Note that in Figure~\ref{fig:loss}, the X-axis are $yh(\xbf)$ values, as $h(\xbf)=y$ is equivalent to $yh(\xbf)\geq 0$ when $\Y = \{-1,1\}$.
\paragraph{Notations} Below, we present the notations that are used throughout the survey. 
\begin{center}
\setlength{\aboverulesep}{4pt}
\setlength{\belowrulesep}{4pt}
\setlength\LTleft{0pt}
\setlength\LTright{0pt}
\begin{longtable}{ll}
\toprule
$\X$ & Input space\\
$\Y$ & Output space\\
$\D$ & A domain: a yet unknown distribution over $\XY$ \\
$\DX$ & Marginal distribution of $\D$ on $\X$ \\
$\hat{\D}_\X$ & Empirical distribution associated with a sample drawn from $\DX$ \\
$\support (\D) $ & Support of distribution $\D$ \\
$\mathrm{\bf Pr}(\cdot)$ & Probability of an event\\
$\mathbb{E}(\cdot)$ & Expectation of a random variable\\
$\xbf\! \eqdef\! \transpose{(x_1,\dots,x_d)}\! \in\! \R^d$ & A \mbox{$d$-dimensional} real-valued vector \\
$(\xbf,y)\sim \D$ & $(\xbf,y)$ is drawn {\it i.i.d.} from $\D$ \\
$ S\! =\!\{(\xbf_i,y_i)\}_{i=1}^m \!\sim\! (\D)^m$ & Labeled learning sample constituted by $m$ examples drawn {\it i.i.d.} from $\D$ \\
$S_u\! =\!\{(\xbf_i)\}_{i=1}^m \!\sim\! (\DX)^m$ & Unlabeled learning sample constituted by $m$ examples drawn {\it i.i.d.} from $\DX$ \\
$\card{S} $ & Size of the set $S$ \\
$\Hcal$ & Hypothesis space \\
$\I{a}$ & Indicator function: returns $1$ if $a$ is true, $0$ otherwise\\
$\sign{a}$ & Return the sign of $a$: $1$ if $a\geq0$, $-1$ otherwise\\
$\mat{M}$ & An arbitrary matrix \\
$\transpose{\mat{M}} $ & Transpose of the matrix $\mat{M}$ \\
$\mat{0}$ & Null vector (matrix)\\
$\|\cdot\|_{1}$ & $L_1$-norm\\
$\|\cdot\|_{\infty}$ & $L_\infty$-norm\\
\bottomrule
\end{longtable}
\end{center}
\vspace{-.9cm}

\subsection{Probably approximately correct setting}
\label{sec:generalizationbounds}
Statistical learning theory~\cite{Vapnik95} provides us with results regarding the conditions that ensure the convergence of the empirical risk to the true risk for a given hypothesis class. These results are known as the \emph{generalization bounds}, and they are usually expressed in the form of probably approximately correct (PAC) inequalities~\cite{Valiant84} that have the following form:
\begin{align*}
\Prob{S\sim (\D)^m}\left\lbrace \vert \risk^\loss_{S}(h)-\risk^\loss_{\D}(h) \vert \leq \varepsilon \right\rbrace  \geq 1-\delta,
\end{align*}
where $\varepsilon >0$ and $\delta \in (0,1]$. 
This expression essentially tells us that we want to upper-bound the gap between the true risk and its estimated value by the smallest possible value of $\varepsilon$ and with a high probability over the random choice of the training sample $S$.
The major question now is to understand whether $\risk^\loss_{S}(h)$ converges to $\risk^\loss_{\D}(h)$ with an increasing size of the learning sample, and what is the speed of this convergence. We now proceed to a presentation of several theoretical paradigms that were proposed in the literature to show the different characteristics of a learning model or a data sample that this speed can depend on.

\subsection{Vapnik-Chervonenkis complexity}
\label{sec:VC}
Vapnik-Charvonenkis (VC) bounds~\cite{VapnikC71,vapnik2006estimation} are based on the original definition that allows quantification of the complexity of a given hypothesis class.
This concept of complexity is captured by the famous VC dimension that is defined as follows.
\begin{definition}{(VC dimension)}
  The VC dimension $\text{VC}(\Hcal)$ of a given hypothesis class $\Hcal$ for the problem of binary classification is defined as the largest possible cardinality of some subset $\X' \subset \X$ for which there exists a hypothesis $h \in \Hcal$ that perfectly classifies elements from $\X'$ whatever their labels are.
More formally, we have
\begin{align*}
\text{VC}(\Hcal) = \max \lbrace \vert \X' \vert: \forall y_i \in \lbrace -1, +1\rbrace ^{\vert \X' \vert}, \exists h \in \Hcal \text{ so that } \forall \xbf_i \in \X', h(\xbf_i) = y_i  \rbrace.
\end{align*}
\end{definition}
\begin{wrapfigure}{r}{.3\linewidth}
\centering
\begin{minipage}{0.45\linewidth}
  \includegraphics[width=\linewidth]{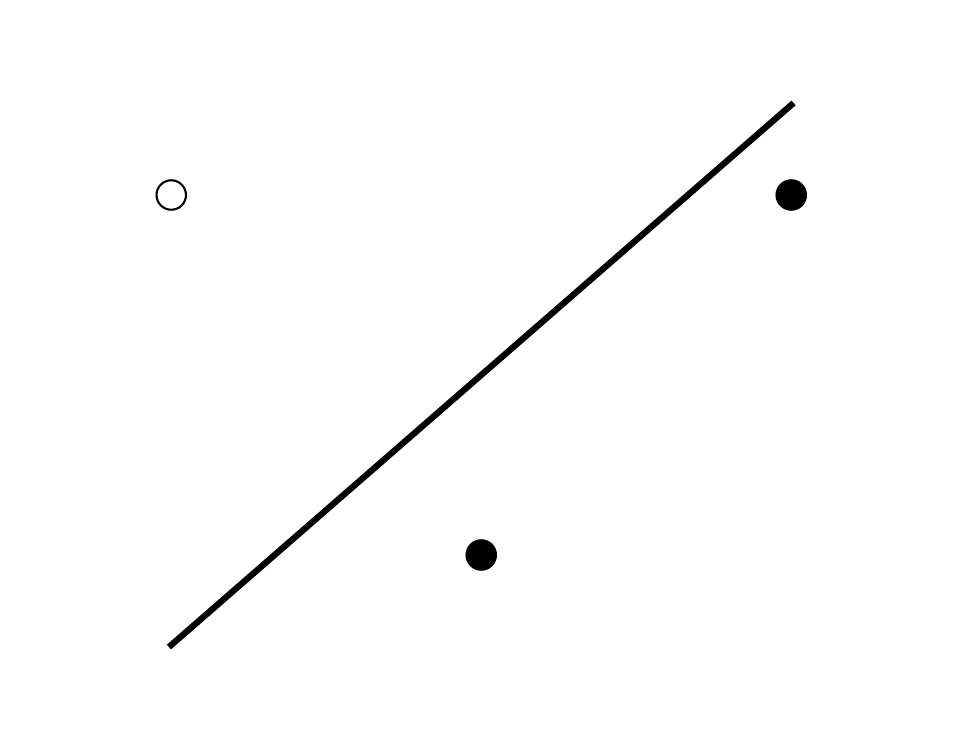}
  \end{minipage}%
\begin{minipage}{0.45\linewidth}
  \includegraphics[width = \linewidth]{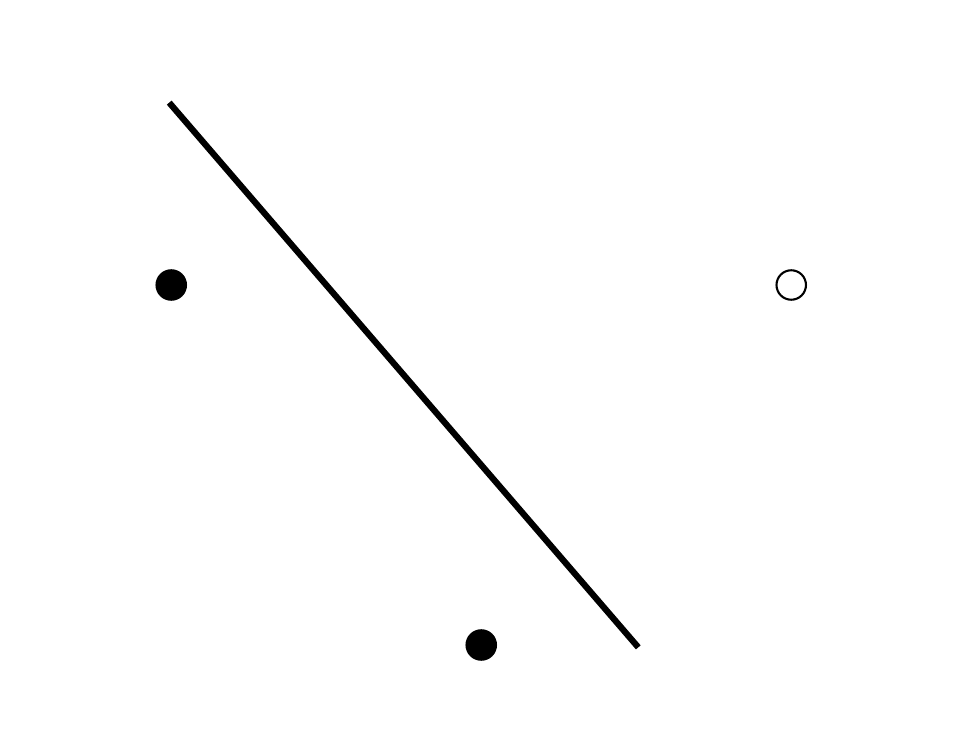}
\end{minipage}
\begin{minipage}{0.45\linewidth}
  \includegraphics[width=\linewidth]{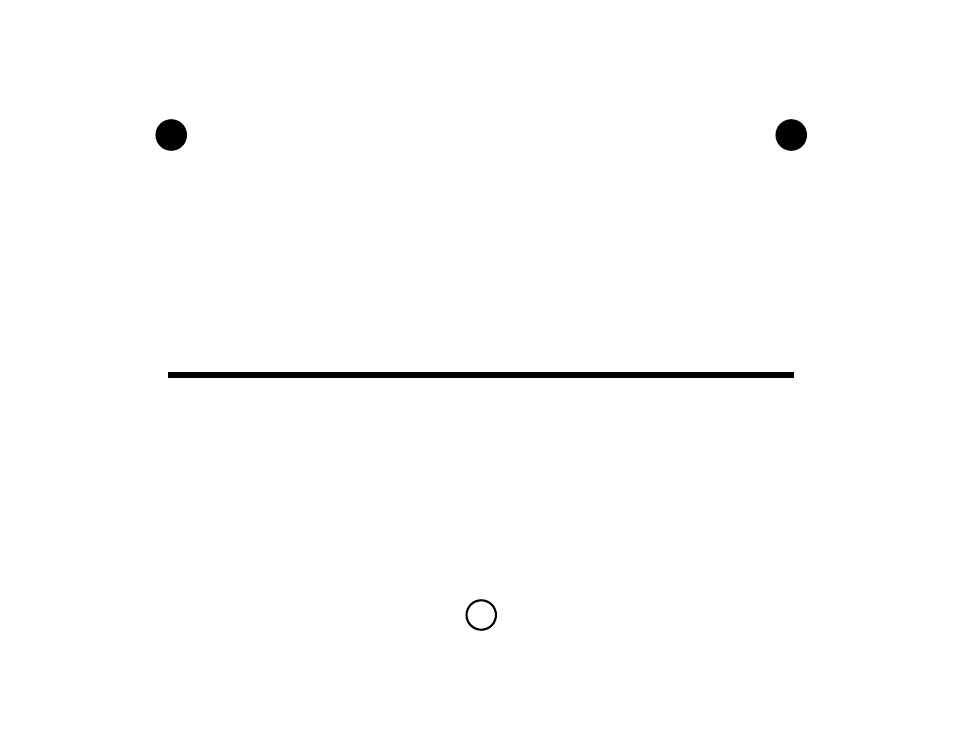}
  \end{minipage}%
\begin{minipage}{0.45\linewidth}
  \includegraphics[width = \linewidth]{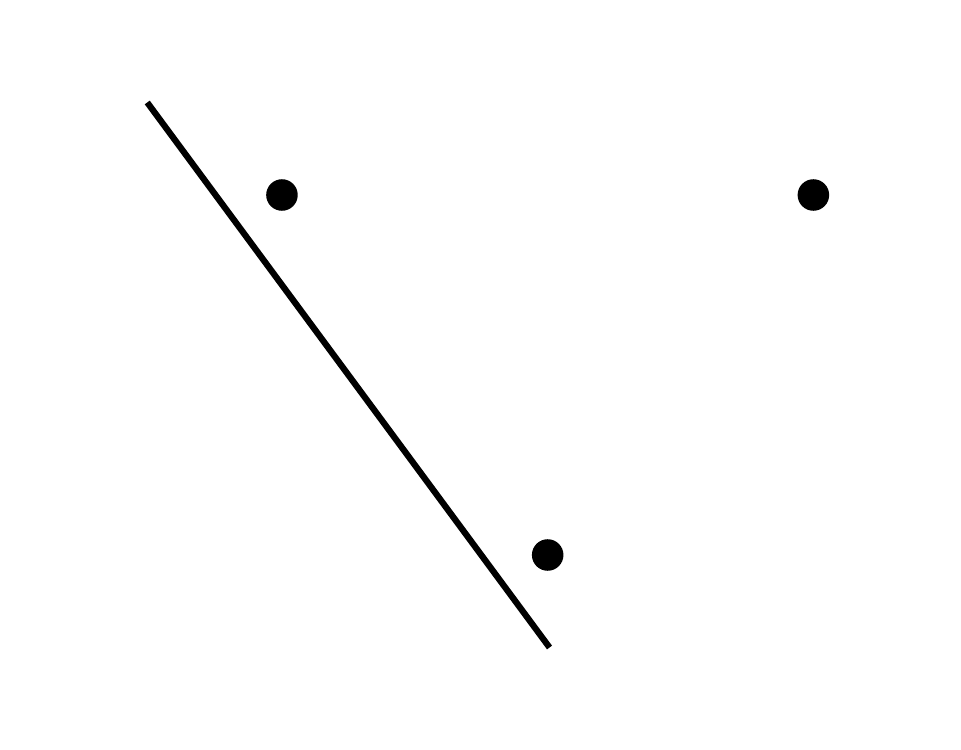}
\end{minipage}
\caption{Illustration of the Vapnik-Charvonenkis (VC) dimension. Here, half-planes in $\mathbb{R}^d$ with $d=2$ can correctly classify at most three points for all possible $2^3$ labelings. The VC dimension here is $2+1$.}
  \label{fig:ex_vcdim}
\end{wrapfigure}
As follows from the definition, the VC dimension is the cardinality of the biggest subset of a given sample that can be subject to perfect classification provided by a hypothesis from $\Hcal$ for all possible labelings of its observations.
To illustrate this, we can consider the classical example given in Figure~\ref{fig:ex_vcdim}, where the hypothesis class $\Hcal$ consists of half-planes in $\mathbb{R}^d$.
In this particular case with $d=2$, we can perfectly classify only $d+1$ elements, regardless their labeling, as for the case with $d+2$ points this will no longer be possible.
This means that the VC dimension of the class of half-planes in $\mathbb{R}^d$ is $d+1$. Note that the result obtained reveals that in this particular scenario, the VC dimension is equal to the number of parameters needed to define the function of the hypothesis plane. This, however, is not true in general, as some classes might have an infinite VC dimension despite the finite number of parameters needed to define the hypothesis class. A common example used in the literature to show this is given by $$\Hcal = \{h_\theta(\xbf): \X \rightarrow \{0,1\}: h_\theta(\xbf) = \frac{1}{2}\sin(\theta \xbf), \theta \in \mathbb{R}\}.$$ 
It can be proven that the VC dimension of this class is infinite. 

The following theorem uses the VC dimension of a hypothesis class to upper-bound the gap between the true and the empirical error for a given loss function and a finite sample of size $m$.
\begin{theorem}
Let $\X$ be an input space, $\Y  = \lbrace -1, +1\rbrace$ the output space, and $\D$ their joint distribution. 
Let $S$ be a finite sample of size $m$ drawn \iid\ from $\D$, and $\Hcal = \lbrace h:X \rightarrow Y \rbrace$ be a hypothesis class of VC dimension $\text{VC}(\Hcal)$. Then for any $\delta \in (0,1]$ with probability of at least $1-\delta$ over the random choice of the training sample $S \sim (\D)^m$, the following holds
\begin{align*}
\forall h \in \Hcal, \quad \risk^\loss_{\D}(h) \leq  \risk^\loss_{S}(h)  + \sqrt{\frac4m\left(\text{VC}(\Hcal) \ln\frac{2em}{\text{VC}(\Hcal)}+\ln\frac{4}{\delta}\right)}.
\end{align*}
\label{trm:VC1}
\end{theorem}

\subsection{Rademacher complexity}
\label{sec:rademacher}
Intuitively, the Rademacher complexity measures the capacity of a given hypothesis class to resist against noise that might be present in the data. This, in turn, was shown to lead to more accurate bounds than those based on the VC dimension~\cite{KoltchinskiiP99}. 
To present the Rademacher bounds, we first provide a definition of a Rademacher variable.
\begin{definition}{(Rademacher variable)}
  A random variable $\kappa$ defined as 
$$    \kappa=\left\{
                \begin{array}{ll}
                  1, \ \text{with probability} \ \frac{1}{2}\,\\
                  -1,\ \text{otherwise}\,,
                \end{array}
              \right.
$$
is called the Rademacher variable.
\end{definition}
From this definition, a Rademacher variable defines a random binary labeling as it takes values $-1$ and $1$ with equal probability and allows the introduction of the Rademacher complexity for an unlabeled sample of size $m$, as follows.
\begin{definition}{(Rademacher complexity)}
For a given unlabeled sample $S =\{(\xbf_i)\}_{i=1}^m$ and a given hypothesis class $\Hcal$, the Rademacher complexity is defined as follows:
\begin{align*}
\mathcal{R}_S(\Hcal) = \esp{\bm{\kappa}} \left[ \sup_{h \in \Hcal} \frac{2}{m}\sum_{i=1}^m \kappa_i h(\xbf_i)\right]\,, 
\end{align*}
where $\bm{\kappa}$ is a vector of $m$ independent Rademacher variables.
The Rademacher complexity for the whole hypothesis class is thus defined as the expected value of $\mathcal{R}_S(\Hcal)$ by
\begin{align*}
\mathcal{R}_m(\Hcal) = \esp{S\sim (\D)^m}\mathcal{R}_S(\Hcal).
\end{align*}
\end{definition}
In this definition, $\mathcal{R}_S(\Hcal)$ encodes the complexity of a given hypothesis class $\Hcal$ based on the observed sample $S$, while $\mathcal{R}_m(\Hcal)$ is the expected value of this complexity over all possible samples that were drawn from some joint probability distribution. 
Contrary to the VC dimension, this complexity measure is defined in terms of the expected value over all labelings, and not only the worst one. 
The following theorem presents the Rademacher-based generalization bound~\cite{KoltchinskiiP99,BartlettM02}.
\begin{theorem}
Let $S=\{(\xbf_i,y_i)\}_{i=1}^m$  be a finite sample of $m$ examples drawn \iid\ from $\D$, and $\Hcal = \lbrace h:\X \rightarrow \Y \rbrace$ be a hypothesis class. 
Then, for any $\delta \in (0,1]$ with probability of at least $1-\delta$ over the choice of the sample $S \sim (\D)^m$, the following holds
\begin{align*}
  \forall h \in \Hcal, \quad &\risk^\loss_{\D}(h)\ \leq\  \risk^\loss_{S}(h)  + \mathcal{R}_m(\Hcal)+\sqrt{\frac{\ln\frac{1}{\delta}}{2m}}.
\end{align*}
\label{thr:radem1}
\end{theorem}

\subsection{PAC-Bayesian bounds}
\label{sec:pac_bayes_SoA}
The PAC-Bayesian approach~\cite{TaylorWilliamson1997,McAllester99} provides generalization bounds for a hypothesis expressed as a weighted majority vote over the hypothesis space $\Hcal$, as, for instance, in ensemble methods~\cite{Dietterich00,ReV12}. 
In this section, we recall the general PAC-Bayesian generalization bound as presented in ~\cite{GermainLLMR15} in the setting of binary classification, where $\Y = \left\lbrace -1, 1\right\rbrace$ with the $0-1$ loss or the linear loss. 
To derive such a generalization bound, a prior distribution $\prior$ over $\Hcal$ is assumed, which models an {\it a-priori} belief on the hypotheses from $\Hcal$ before the observation of the training sample $S\sim (\D)^m$. 
Given $S$, the learner aims to find a posterior distribution $\posterior$ over $\Hcal$ that leads to a well-performing \mbox{$\posterior$-weighted} majority vote $\BQ(\xbf)$ (also called the Bayes classifier), defined as
\begin{align*}
\BQ(\xbf) = \sign{\esp{ h \sim \posterior} h(\xbf)}.
\end{align*} 
In other words, rather than finding the best hypothesis from $\Hcal$, we want to learn $\posterior$ over $\Hcal$, such that this minimizes the true risk $\RD(\BQ)$ of the \mbox{$\posterior$-weighted} majority vote.
However, PAC-Bayesian generalization bounds do not directly focus on the risk of the deterministic \mbox{$\posterior$-weighted} majority vote $\BQ$, but on giving an upper bound over the expectation over $\posterior$ of all of the individual hypothesis true risks, called the \emph{Gibbs risk}: $\espdevant{h \sim \posterior} \risk^\loss_{\D}(h)$.
The Gibbs risk is associated to a stochastic classifier, called the Gibbs classifier, which draws a hypothesis $h$ from $\Hcal$ according to the posterior distribution $\posterior$, and predicts the label of $\xbf$ given by $h(\xbf)$.
An important behavior of the Gibbs risk is that it is closely related to the deterministic \mbox{$\posterior$-weighted} majority vote.
Indeed, if $\BQ$ miss-classifies $\xbf\in \X$, then at least half of the classifiers (under measure $\posterior$) make a prediction error on $\xbf$.
Therefore, we have
\begin{align}
\label{eq:gibbsrelation}
 \risk^\loss_{\D}(\BQ) \leq 2\, \esp{h \sim \posterior} \risk^\loss_{\D}(h). 
\end{align}
Thus,  an upper bound on $\esp{h \sim \posterior} \risk^\loss_{\D}(h)$ 
provides an upper bound on $ \risk^\loss_{\D}(\BQ)$ as well.

Note that PAC-Bayesian generalization bounds do not directly take into account the complexity of the hypothesis class $\Hcal$, contrary to the Rademacher complexity or the VC dimension, but they measure the deviation between the prior distribution $\prior$ and the posterior distribution $\posterior$ on $\Hcal$ through the Kullback-Leibler divergence:
\begin{align*}
\KL{\posterior}{\prior}\ =\ \esp{h \sim \posterior} \ln \frac{\posterior(h)}{\prior(h)}.
\end{align*}

The result that follows is a general PAC-Bayesian theorem that takes the form of an upper bound on the deviation between the true and empirical Gibbs risks when measured by a convex function $D : [0 , 1]  \times  [0, 1] \to \R$.
\begin{theorem}[\cite{GermainLLM09,GermainLLMR15}]
\label{theo:PB}
For any distribution $\D$ on $\X \times \Y$, for any hypothesis class $\Hcal$, for any prior distribution $\prior$ on $\Hcal$, for any $\delta  \in  (0,1]$, for any convex function $D : [0 , 1]  \times  [0, 1] \to \R $, with a probability of at least $1 - \delta$ over the random choice of $S \sim (\D)^m$, we have, for all posterior distribution $\posterior$ on $\Hcal$,
\begin{align*}
D \left(  \esp{h \sim \posterior} \risk^\loss_{S}(h), \esp{h \sim \posterior} \risk^\loss_{\D}(h)  \right) \leq\  \frac{1}{m} \bigg[  \KL{\posterior}{\prior}   +  \ln \bigg( \frac{1}{\delta} \esp{S \sim (\D)^m}\, \esp{h \sim \prior}  e^{m\,D\left( \risk^\loss_{S}(h),  \risk^\loss_{\D}(h) \right)}   \bigg)   \bigg].
\end{align*}
\end{theorem}
By upper-bounding $\esp{S \sim (\D)^m}\, \esp{h \sim \prior}  e^{m\,D(\risk^\loss_{S}(h))}$ and by selecting a well-suited deviation function $D$, we can retrieve the classical versions of the PAC-Bayesian theorem (\ie, \cite{McAllester99,Seeger02,catoni2007pac}).

\subsection{Uniform stability}
\label{sec:UniformStability}
As the complexity of the hypothesis class intuitively depends directly on the properties of a learning algorithm, it might be desirable to have the generalization bounds that manifest this relationship explicitly.
\cite{BousquetE02} introduced generalization bounds that provide a solution to this problem based on the concept of uniform stability of a learning algorithm. We now give its definition.
\begin{definition}{(Uniform stability)}
An algorithm $\mathcal{A}$ has uniform stability $\beta$ with respect to the loss function $\loss$ if the following holds
\begin{align*}
\forall S \in \lbrace \X \times \Y\rbrace^m, \forall i \in \lbrace 1, \dots, m\rbrace, \sup_{(\xbf,y) \in S} \abs{\loss(h_S(\xbf),y)-\loss(h_{S^{\setminus i}}(\xbf),y)} \leq \beta\,,
\end{align*}
where the hypothesis $h_S$ is learned on the sample $S$ while $h_{S^{\setminus i}}$ is obtained on $S$ with its $i^{\text{th}}$ observation being deleted.
\end{definition}
The intuition behind this definition is to say that an algorithm that is expected to generalize well should be robust to small perturbations in the training sample. Consequently, stable algorithms should have an empirical error that remains close to their generalization error. 
This idea is confirmed by the following theorem.
\begin{theorem}
Let $\mathcal{A}$ be an algorithm with uniform stability $\beta$ with respect to a loss function $\loss$, such that $0 \leq \loss(h_S(\xbf,y) \leq M$, for all $(\xbf,y) \in (\X \times \Y)$ and all sets $S$. 
Then, for any $m \geq 1$, and any $\delta \in (0,1]$, the following bound holds with probability of at least $1-\delta$ over the random choice of the sample $S$,
\begin{align*}
\risk^\loss_{\D}(h_S) \leq  \risk^\loss_{S}(h_S)  + 2\beta + (4m\beta+M)\sqrt{\frac{\ln\frac{1}{\delta}}{2m}}.
\end{align*}
\end{theorem}

\subsection{Algorithmic robustness}
\label{sec:robustness}
The main underlying idea of algorithmic robustness~\cite{XuM10,XuM12} is to say that a robust algorithm should have similar performance in terms of the classification error for testing and training samples that are close. 
The measure of similarity used to define whether two samples are close or not relies on partitioning the joint space $\X \times \Y$ in a way that puts two similar points of the same class in the same partition. This partition is further defined using the concept of covering numbers~\cite{KolmogorovT59}, as introduced below.
\begin{definition}{(Covering number)} Let $(Z,\varrho)$ denote a metric space with metric $\varrho(\cdot)$ defined on $Z$.
  For $Z' \subset Z$, we say that $\hat{Z}'$ is a $\gamma$ covering of $Z'$, if for any element $t \in Z'$ there is an element $\hat{t} \in \hat{Z}'$ such that $\varrho(t,\hat{t})\leq \gamma$.
  Then the number of $\gamma$ covering of $Z'$ is expressed as
\begin{align*}
N(\gamma, Z', \varrho) = \min \left\lbrace \abs{\hat{Z}'} :  \hat{Z}' \ \text{is a } \gamma \text{-covering of }Z'  \right\rbrace.
\end{align*}
\label{def:cov_numb}
\end{definition}
In the case where $\X$ is a compact space, its covering number $N(\gamma, \X, \varrho)$ is finite. Furthermore, for the product space $\X \times \Y$, the number of $\gamma$-covering is also finite and is equal to $\abs{\Y}N(\gamma, \X, \varrho)$. 
As previously explained, the above partitioning ensures that two points from the same subset are from the same class and are close to each other with respect to metric $\varrho$. 
Bearing this in mind, the algorithmic robustness is defined as follows.
\begin{definition}(Algorithmic robustness)
  \label{theo:robustness}
  Let $S$ be a training sample of size $m$ where each example is drawn from the joint distribution $\D$ on $\X \times \Y$.
  An algorithm $\mathcal{A}$ is \mbox{$(M,\epsilon(\cdot))$-robust} on $\D$ with respect to a loss function $\loss$ for $M \in \mathbb{N}$ and $\epsilon(\cdot): (\X \times \Y)^m \rightarrow \mathbb{R}$ if $\X \times \Y$ can be partitioned into $M$ disjoint subsets denoted by $\{ Z_j \}_{j=1}^M$, so that for all $(\xbf,y) \in \X \times \Y$, $(\xbf',y')$ drawn from $\D$ and $j \in \{1, \dots, M \}$ we have
\begin{align*}
\big((\xbf,y),(\xbf',y')\big) \in Z_j^2 \quad \longrightarrow \quad \abs{\loss(h_S(\xbf),y)-\loss(h_S(\xbf'),y')}\leq \epsilon(S)\,,
\end{align*}
where $h_S$ is a hypothesis learned by $\mathcal{A}$ on $S$.
\end{definition}
We are now ready to present the generalization guarantees that characterize robust algorithms that verify the definition presented above. 
\begin{theorem}
Let $S$ be a finite sample of size $m$ drawn \iid\ from $\D$, $\mathcal{A}$ be $(M,\epsilon(\cdot))$-robust on $\D$ with respect to a loss function $\loss(\cdot,\cdot)$, such that $0 \leq \loss(h_S(\xbf),y) \leq M_\loss$, for all $(\xbf,y) \in (\X \times \Y)$. 
Then, for any $\delta \in (0,1]$, the following bound holds with probability of at least $1-\delta$ over the random draw of the sample $S \sim (\D)^m$,
  \begin{align*}
    \risk^\loss_{\D}(h_S) \leq  \risk^\loss_{S}(h_S)  + \epsilon(S) + M_\loss\sqrt{\frac{2M\ln2  + 2\ln\frac{1}{\delta}}{m}}\,,
  \end{align*}
where $h_S$ is a hypothesis learned by $\mathcal{A}$ on $S$.
\end{theorem}
Note that the algorithmic robustness focuses on measuring the divergence between the costs associated to two similar points, assuming that the learned hypothesis function should be locally consistent.
Uniform stability, in turn, explores the variation in the cost due to perturbations of the training sample, and thus assumes that the learned hypothesis does not change much.

\section{Seminal divergence-based learning bounds}
\label{chap:div_based}
In this section, we provide the description of domain adaptation generalization bounds that laid the foundation of this field. These seminal bounds mainly relied on traditional divergence measures between the probability distributions, to relate the source and target domains.
\subsection{Learning bound based on the $L^1$-distance}
From a theoretical point of view, the domain adaptation problem was rigorously investigated for the first time by ~\cite{BenDavid07analysisof} and \cite{bendavidth}\footnote{Note that in \cite{bendavidth}, the authors presented an extended version of the results previously published in \cite{BenDavid07analysisof} and \cite{Blitzer07Learning}.}. 
The authors of these papers focused on the domain adaptation problem following VC theory (recalled in Section~\ref{sec:VC}) and considered the $0-1$ loss (Equation~\ref{eq:zeroone}) function in the setting of binary classification with $\Y=\{-1,+1\}$. 
They further proposed to make use of the \emph{$L^1$-distance}, the definition of which is given below. 
\begin{definition}{($L^1$-distance)}
  \label{def:L1dist}
Let $\mathcal{B}$ denote the set of measurable subsets under two probability distributions $\D_1$ and $\D_2$. The $L^1$-distance or the total variation distance between $\D_1$ and $\D_2$ is defined as
\begin{align*}
d_1(\D_1,\D_2) = 2\sup_{B \in \mathcal{B}} \left\vert \Prob{\D_1}(B) - \Prob{\D_2}(B)\right\vert.
\end{align*}
\end{definition}
The $L^1$-distance is a proper metric on the space of probability distributions that informally quantifies the largest possible difference between the probabilities that the two probability distributions $\D_1$ and $\D_2$ can assign to the same event $B$.
This distance is relatively popular in many real-world applications, such as image denoising or numerical approximations of partial derivative equations.  

Starting from Definition \ref{def:L1dist}, the first important result from their work was formulated as follows.
\begin{restatable}[\cite{BenDavid07analysisof}]{theorem}{bdun}
Given two domains $\source$ and $\target$ over $\X\times\Y$ and a hypothesis class $\Hcal$, the following holds 
\begin{align*}
\forall h\in\Hcal,\quad \risk^\zoloss_{\target} (h) \leq \risk^\zoloss_{\source}(h) + d_1(\sourceX,\targetX) + \min \left\lbrace \esp{\xbf\sim\sourceX}\left[\vert f_\source(\xbf) - f_\target(\xbf) \vert\right], \esp{\xbf\sim\targetX}\left[\vert f_\target(\xbf) - f_\source(\xbf) \vert\right] \right\rbrace\,,
\end{align*}
where $f_\source(\xbf)$ and $f_\target(\xbf)$ are the source and target true labeling functions associated to $\source$ and $\target$, respectively.
\label{trm:bd1}
\end{restatable}
This theorem presents the first result that relates the performance of a given hypothesis function with respect to two different domains. 
It implies that the error achieved by a hypothesis in the source domain upper-bounds the true error on the target domain where the tightness of the bound depends on the distance between their distributions and that of the labeling functions.

\subsection{Learning bound based on $\hdh{}{}$-divergence} Despite being the first result of this kind proposed in the literature, the idea of bounding the error in terms of the \mbox{$L^1$-distance} between the marginal distributions of the two domains includes two important restrictions: 1) the $L^1$-distance cannot be estimated from finite samples for arbitrary probability distributions; and 2) it does not allow the divergence measure to be linked to the considered hypothesis class, and thus leads to very loose inequality. 

To address these issues, the authors further defined the \mbox{$\hdh$-divergence} based on the $\mathcal{A}$-divergence introduced in \cite{KiferBG04} for detection of changes in data streams. We give its definition below.
\begin{definition}[Based on \cite{KiferBG04}]
Given two domains' marginal distributions $\sourceX$ and $\targetX$ over the input space $\X$, let $\Hcal$ be a hypothesis class, and let $\hdh$ denote the symmetric difference hypothesis space defined as
$$h \in \hdh \Longleftrightarrow\ h(\xbf) = g(\xbf) \oplus g'(\xbf)\,,$$
for some $(g,g')^2 \in \Hcal^2$, where $\oplus$ stands for the XOR operation.
Let $I(h)$ denote the set for which $h \in \hdh$ is the characteristic function, \ie, $\xbf \in I(h) \Leftrightarrow g(\xbf) = 1$. 
The \mbox{$\hdh$-divergence} between $\sourceX$ and $\targetX$ is defined as:
  \begin{align*}
  \dhdh{\sourceX}{\targetX} = 2\sup_{h \in \hdh} \left\vert \Prob{\sourceX}(I(h)) - \Prob{\targetX}(I(h))\right\vert.
  \end{align*}
\end{definition}
The $\hdh$-divergence solves both problems associated with the \mbox{$L^1$-distance}. First, from its definition, we can see that $\hdh$-divergence explicitly takes into account the considered hypothesis class. This ensures that the bound remains meaningful and directly related to the learning problem at hand.
On the other hand, the \mbox{$\hdh$-divergence} for any class $\Hcal$ is never larger than the \mbox{$L^1$-distance}, and thus can lead to a tighter bound. Finally, for a given hypothesis class $\Hcal$ of finite VC dimension, the \mbox{$\hdh$-divergence} can be estimated from finite samples using the following lemma. 
\begin{lemma}
  Let $\Hcal$ be a hypothesis space of VC dimension $\text{VC}(\Hcal)$.
  If $S_u$, $T_u$ are unlabeled samples of size $m$ each, drawn independently from $\sourceX$ and $\targetX$, respectively, then for any $\delta \in (0,1)$ with probability of at least $1-\delta$ over the random choice of the samples, we have
  \begin{align}
    \label{eq:hdh_VC}
    d_\hdh(\sourceX,\targetX) \leq \hat{d}_\hdh(S_u,T_u)  + 4\sqrt{\frac{2\,\text{VC}(\Hcal)\log (2m)+\log(\frac{2}{\delta})}{m}}\,,
    \end{align}
where $\hat{d}_\hdh(S_u,T_u)$ is the empirical \mbox{$\hdh$-divergence} estimated on $S_u$ and $T_u$.
\label{trm:hdiv}
\end{lemma}

\begin{figure}
\centering
\begin{minipage}{0.8\linewidth}
  \includegraphics[width=\linewidth]{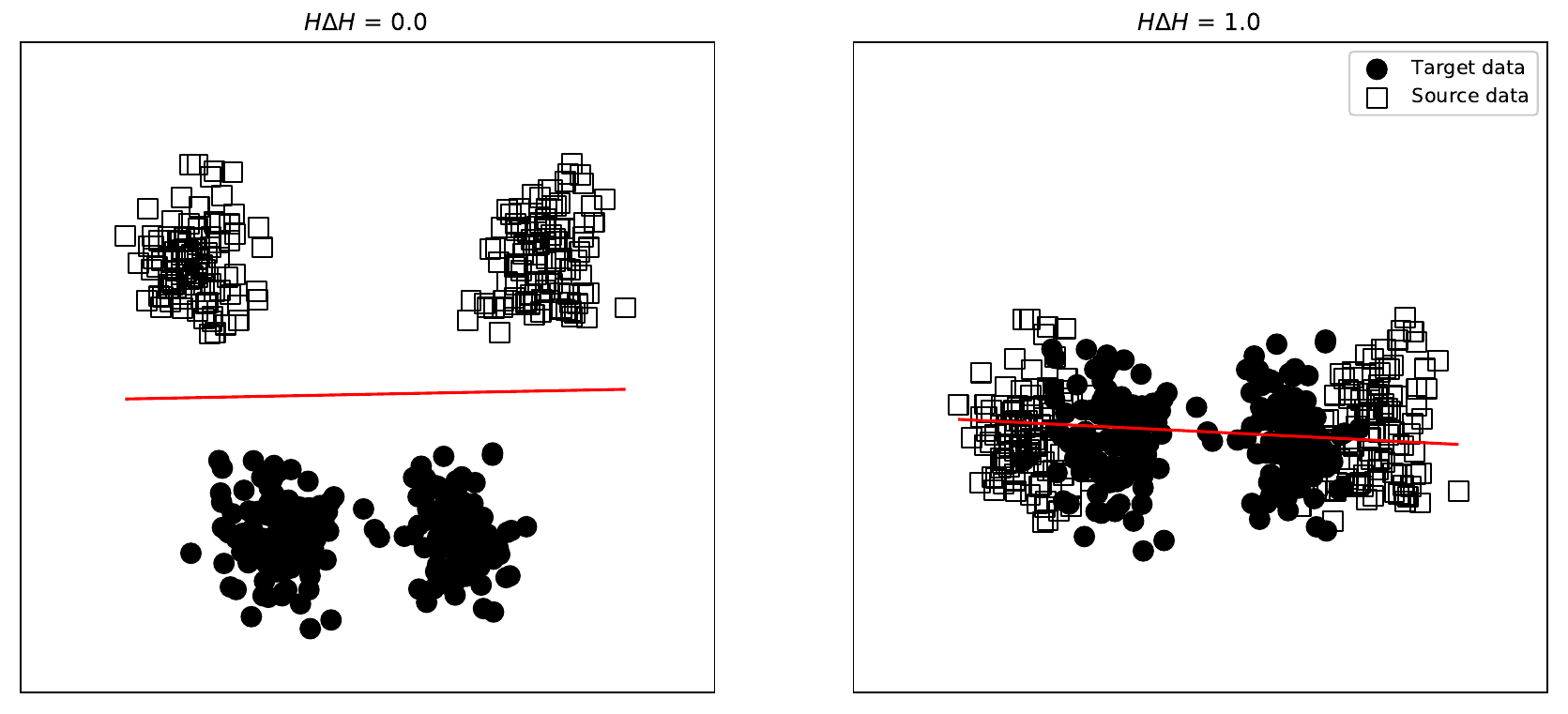}
  \end{minipage}
  \\
\begin{minipage}{0.8\linewidth}
  \includegraphics[width = \linewidth]{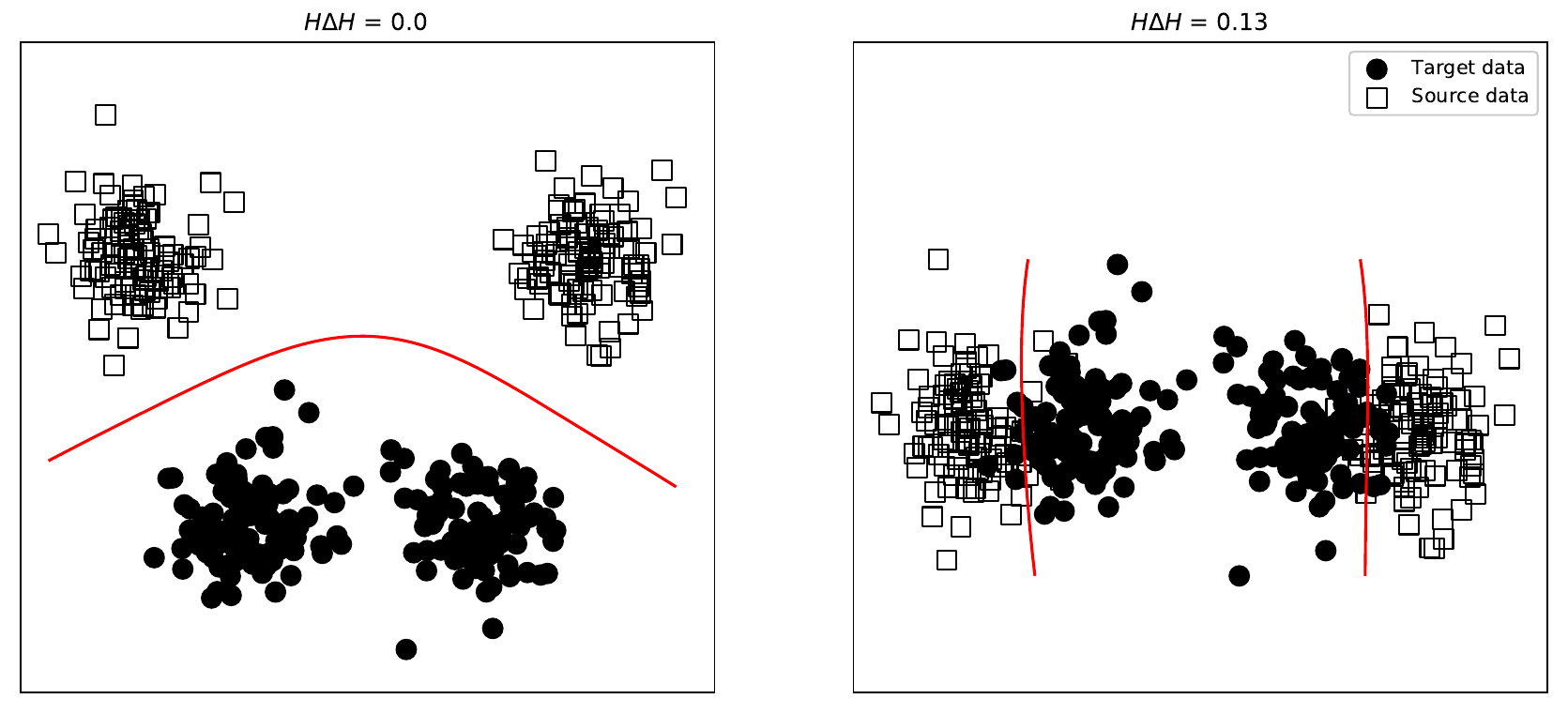}
\end{minipage}
\caption{Illustration of the $\hdh$-divergence when the hypothesis class consists of linear (\textbf{top row}) and nonlinear (\textbf{bottom row}) classifiers. Note that the indicated value of $\hdh$ is the error of the obtained classifier without subtracting 1 and multiplying the result by two, as in \Cref{trm:hdiv_emp}.}
\label{fig:ex_hdh}
\end{figure}

Inequality~\eqref{eq:hdh_VC} shows that with an increasing number of instances and for a hypothesis class of finite VC dimension, the empirical \mbox{$\hdh$-divergence} can be a good proxy for its true counterpart.
The former can be further calculated thanks to the following result.
\begin{restatable}[\cite{bendavidth}]{lemma}{hdivemp}
  Let $\Hcal$ be a hypothesis space.
  Then, for two unlabeled samples $S_u$, $T_u$ of size $m$, we have
$$\hat{d}_\hdh(S_u,T_u) = 2\left(1\! -\!\!\! \min_{h \in \hdh} \left[\frac{1}{m} \sum_{\xbf: h(\xbf) = 0}\!\! \I{\xbf\in S_u} + \frac{1}{m} \sum_{\xbf: h(\xbf) = 1}\!\! \I{\xbf\in T_u}\right] \right).$$
\label{trm:hdiv_emp}
\end{restatable}
It can be noted that the expression of the empirical \mbox{$\hdh$-divergence} given above is essentially the error of the best classifier for the binary classification problem of distinguishing between the source and target instances pseudo-labeled with $0$'s and $1$'s.
In practice, this means that the value of the \mbox{$\hdh$-divergence} depends explicitly on the hypothesis class used to produce such a classifier. This dependence and the intuition behind the \mbox{$\hdh$-divergence} are illustrated in Figure~\ref{fig:ex_hdh}.
In Figure~\ref{fig:ex_hdh}, we consider two different domain adaptation problems, where for one of them the source and target samples are well separated, while for the other, the source and target data are mixed together.
To calculate the value of the \mbox{$\hdh$-divergence}, we need to choose a hypothesis class used to produce a classifier that distinguishes between them.
Here, we consider two different families of classifiers: a linear support vector machine classifier, and its nonlinear version with radial basis function kernels.
For each solution, we also plot the decision boundaries to see how the source and target instances are classified in both cases.
From the visualization of the decision boundaries, we note that the linear classifier fails to distinguish between the mixed source and target instances, while the nonlinear classifier manages to do this relatively well.
This is reflected by the value of the \mbox{$\hdh$-divergence}, which is zero in the first case for both classifiers, and is drastically different for the second adaptation problem. Having two different divergence values for the same adaptation problem might appear surprising at first sight, but this has a simple explanation.
By choosing a richer hypothesis class composed of nonlinear functions, we have increased the VC complexity of the considered hypothesis space, and have thus increased the complexity term in Lemma~\ref{trm:hdiv}.
This shows the trade-off that has to be borne in mind when the \mbox{$\hdh$-divergence} is calculated in the same way as is suggested by general VC theory.

At this point, we already have a "reasonable" version of the \mbox{$L^1$-distance} used to derive the first seminal result. We have also presented its finite sample approximation, but we have not yet applied this to relate the source and target error functions.
The next lemma gives the final key needed to obtain a learning bound for domain adaptation that is linked to a specific hypothesis class and is derived for the available source and target finite size samples. This reads as follows.
\begin{restatable}[\cite{bendavidth}]{lemma}{hdhun}
Let $\source$ and $\target$ be two domains on $\X\times\Y$. For any pair of hypotheses $(h,h') \in \hdh^2$, we have
$$\left\vert \RT^\zoloss(h,h')  - \RS^\zoloss(h,h') \right\vert\ \leq\ \frac{1}{2}\,\dhdh{\sourceX}{\targetX}.$$
\label{trm:hdh1}
\end{restatable}
Note that in this lemma, the source and target risk functions are defined for the same pairs of hypotheses, while the true risk should be calculated based on a given hypothesis and the corresponding labeling function. This result presents the complete learning bound for domain adaptation with \mbox{$\hdh$-divergence}, and it is established by means of the following theorem. 
\begin{restatable}[\cite{bendavidth}]{theorem}{bddeux}
Let $\Hcal$ be a hypothesis space of VC dimension $\text{VC}(\Hcal)$. 
If $S_u$, $T_u$ are unlabeled samples of size $m'$ each, which are drawn independently from $\sourceX$ and $\targetX$, respectively, then for any $\delta \in (0,1)$ with probability of at least $1-\delta$ over the random choice of the samples, then for all $h\in \Hcal$
$$ \RT^\zoloss(h) \ \leq \RS^\zoloss(h) + \tfrac{1}{2}\hat{d}_{\hdh}(S_u,T_u) + 4\sqrt{\frac{2\,\text{VC}(\Hcal)\log (2m')+\log(\frac{2}{\delta})}{m'}} +\lambda\,,$$
where $\lambda$ is the combined error of the ideal hypothesis $h^*$ that minimizes $\RS(h)+\RT(h)$.
\label{trm:bd2}
\end{restatable}
As indicated at the beginning of this section, a meaningful domain adaptation generalization bound should include two terms that reflect both the divergence between the marginal distribution of the source and target domains, and the divergence between their labeling functions. The first term here is obviously reflected by the \mbox{$\hdh$-divergence} between the observable samples, while the second term is given by the $\lambda$ term, as it depends on the true labels (and can be seen as a measure of capacity to adapt). The presence of the trade-off between source risk, divergence, and capability to adapt is a very important phenomenon in domain adaptation.
Indeed, it shows that the reduction in the divergence between the samples can be insufficient when there is no hypothesis that can achieve a low error on both the source and target samples. 

\paragraph{The semi-supervised case}
In the unsupervised case that we have considered previously, it is assumed that there is no access to labeled instances in the target domain that can help to guide adaptation. For this case, the main strategy that leads to an efficient adaptation is to have a classifier learned on a target-aligned labeled sample from the source domain, and to apply it directly in the target domain afterwards. While this situation occurs relatively often in practice, many applications can be found where several labeled target instances are available during the learning stage. In what follows, we consider this situation and give a generalization bound for it, which shows that the error obtained by a classifier that has been learned on a mixture of source and target labeled data can be upper-bounded by the error of the best classifier learned using the target domain data only. 

To proceed, let us now assume that we have $\beta m$ instances drawn independently from $\target$ and $(1-\beta)m$ instances drawn independently from $\source$ and labeled by $f_\source$ and $f_\target$, respectively. A natural goal for this setting is to use the available labeled instances from the target domain to find a trade-off between minimizing the source and the target errors depending on the number of instances available in each domain and the distance between them.
In this case, we can consider the empirical combined error \cite{Blitzer07Learning} defined as a convex combination of errors on the source and target training data for $\alpha \in [0,1]$:
$$\Remp^\alpha(h) = \alpha \RTemp^\zoloss(h) + (1-\alpha)\RSemp^\zoloss(h).$$

The use of the combined error is motivated by the fact that if the number of instances in the target sample is small compared to the number of instances in the source domain (which is usually the case in domain adaptation), minimizing only the target error might not be appropriate.
Instead, there might be the need to find a suitable value of $\alpha$ that ensures the minimum of $\risk^\alpha(h)$ with respect to a given hypothesis $h$.
Note that in this case, the shape of the generalization bound that we are interested in becomes different.
Indeed, in all previous theorems the goal was to upper-bound the target error by the source error, while in this case we would like to know whether learning a classifier minimizing the combined error is better than minimizing the target error using the available labeled instances alone.
The answer to this question is given by the following theorem. 
\begin{restatable}[\cite{Blitzer07Learning,bendavidth}]{theorem}{combined}
Let $\Hcal$ be a hypothesis space of VC dimension $\text{VC}(\Hcal)$.
Let $\source$ and $\target$ be the source and target domains, respectively, defined on $\X\times\Y$.
Let $S_u$, $T_u$ be unlabeled samples of size $m'$ each, drawn independently from $\sourceX$ and $\targetX$, respectively. 
Let $S$ be a labeled sample of size $m$ generated by drawing $\beta\, m$ points from $\target$ ($\beta \in [0,1]$) and $(1-\beta)\,m$ points from $\source$ and labeling them according to $f_\source$ and $f_\target$, respectively.
If $\hat{h} \in \Hcal$ is the empirical minimizer of $\Remp^\alpha(h)$ on $S$ and $h_T^* = \argmin{h \in \Hcal} \RT^\zoloss(h)$ then for any $\delta \in (0,1)$, with probability of at least $1-\delta$ over the random choice of the samples, we have
 $$\RT^\zoloss(\hat{h})\ \leq\ \RT^\zoloss(h_T^*) + c_1 + c_2\,,$$
where 
\begin{align} 
&c_1\ =\ 4 \sqrt{\frac{\alpha^2}{\beta} + \frac{(1-\alpha)^2}{1-\beta}} \sqrt{\frac{2\,\text{VC}(\Hcal)\log (2(m+1))+2\log (\frac{8}{\delta})}{m}}\,,\nonumber\\
\mbox{and}\ &c_2 \ =\  2(1-\alpha) \left(  \frac{1}{2}\dhdh{S_u}{T_u}+ 4\sqrt{\frac{2\,\text{VC}(\Hcal) \log(2m')+\log(\frac{8}{\delta})}{m'}} +\lambda\right).
\label{eq:theo:combined}
\end{align}
\label{theo:combined}
\end{restatable}
This theorem presents an important result that reflects the usefulness of the combined minimization of the source and target errors based on the available labeled samples in both domains compared to the minimization of the target error only. This essentially shows that the error achieved by the best hypothesis of the combined error in the target domain is always upper-bounded by the error achieved by the hypothesis of the best target domain.
Furthermore, this indicates two important consequences: 
\begin{enumerate}
\item if $\alpha = 1$, the term related to the \mbox{$\hdh$-divergence} between the domains disappears, as in this case we have enough labeled data in the target domain and a low-error hypothesis can be produced solely from the target data; 
\item if $\alpha = 0$, the only way to produce a low-error classifier on the target domain is to find a good hypothesis in the source domain while minimizing the \mbox{$\hdh$-divergence} between the domains. In this case, it has also to be assumed that $\lambda$ is low, so that the adaptation is possible. 
\end{enumerate}

Additionally, Theorem~\ref{theo:combined} can provide some insights into the optimal mixing value of $\alpha$ depending on the quantity of labeled instances in the source and target domains. To illustrate this, the right-hand side of Equation~\eqref{eq:theo:combined} can be rewritten
as a function of $\alpha$, to understand when this function is minimized. This gives
\begin{align*}
f(\alpha) = 2B \sqrt{\frac{\alpha^2}{\beta} + \frac{(1-\alpha)^2}{1-\beta}} + 2(1-\alpha)A,
\end{align*}
where $$B = \sqrt{\frac{2\,\text{VC}(\Hcal)\log (2(m+1))+2\log (\frac{8}{\delta})}{m}}$$ is a complexity term that is approximately equal to $\sqrt{\text{VC}(\Hcal)/m}$ and 
$$A =  \frac{1}{2}\hat{d}_\hdh(S_u,T_u)+ 4\sqrt{\frac{2\,\text{VC}(\Hcal) \log(2m')+\log(\frac{8}{\delta})}{m'}} +\lambda$$ is the total divergence between the two domains. 

It then follows that the optimal value $\alpha^*$ is a function of the number of target examples $m_T = \beta m$, the number of source examples $m_S = (1-\beta)m$, and the ratio $D = \sqrt{\text{VC}(\Hcal)}/A$: 
\begin{align*}
\alpha^*(m_S,m_T,D) = \left\{
\begin{array}{ll}
                  1, \ m_T \geq D^2\\
                  \min(1,\nu), \ m_T \leq D^2
                \end{array}
                \right.
\end{align*}
where $$\nu = \frac{m_T}{m_T+m_S}\left( 1 + \frac{m_S}{\sqrt{D^2(m_S+m_T)-m_Sm_T}}  \right).$$

\begin{figure}[!ht]
\centering
\includegraphics[width = 0.8\linewidth]{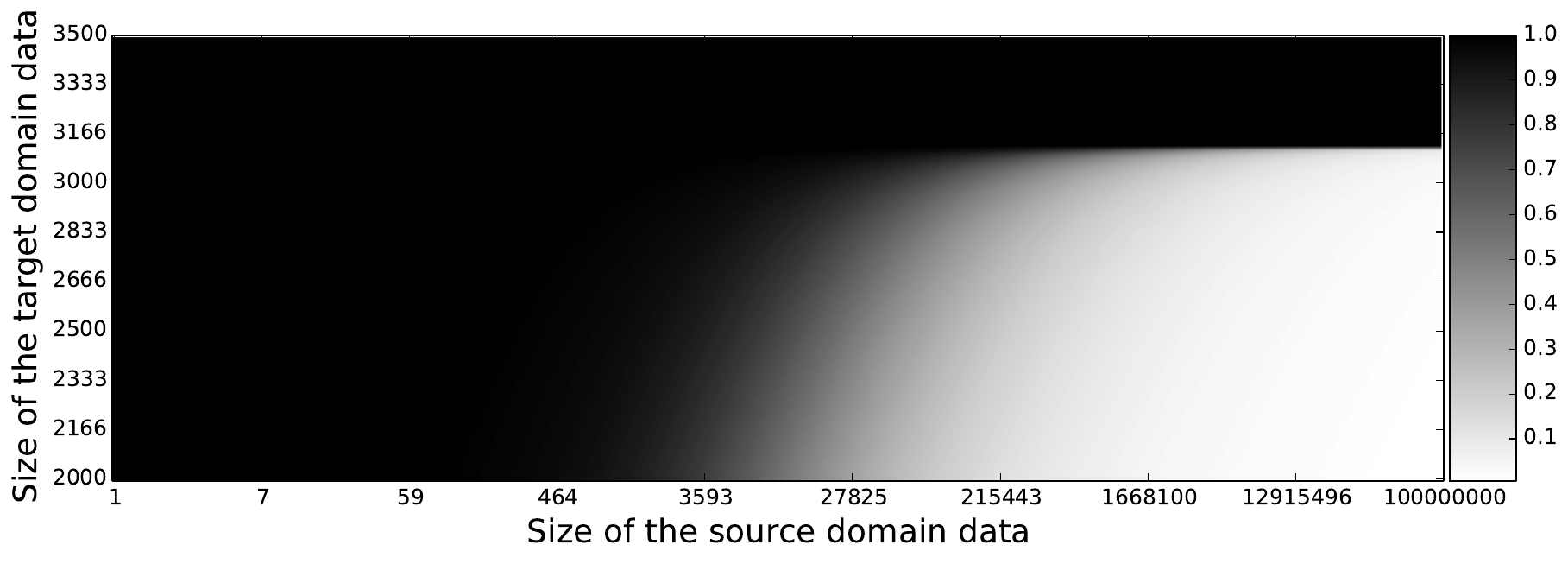}
\caption{Illustration of the optimal value for $\alpha$ as a function of the number of source and target labeled instances.}
\label{fig:comb_error}
\end{figure}

As mentioned in \cite{bendavidth}, this reformulation offers two interesting insights. First, if $m_T = 0$ ({\it $\beta = 0$}) then $\alpha^* = 0$, and if $m_S = 0$ (\ie, $\beta = 1$) then $\alpha^* = 1$. As mentioned above, this implies that if we have only source or only target labeled data, the most appropriate choice is to use them for learning directly. Secondly, if the divergence between two domains is zero, then the optimal combination is to use the training data with uniform weighting of the examples. On the other hand, if there are enough target data, \ie, $m_T \geq D^2 = \text{VC}(\Hcal)/A^2$, then no source data are required for efficient learning, and using it will be detrimental to the overall performance. This is because the possible error decrease as a result of using additional source data is always subject to its increase due to the increasing divergence between the source and target data. Secondly, for a few target examples, we might not have enough source data to justify its use. In this case, the sample of the source domain can be simply ignored. Finally, once we have enough source instances combined with a few target instances, $\alpha^*$ takes on intermediate values. This analysis is illustrated in Figure~\ref{fig:comb_error}.

\subsection{Generalization bounds based on a discrepancy distance}
\label{sec:mansour}
One important limitation of the \mbox{$\hdh$-divergence} is its explicit dependence on a particular choice of a loss function, which is taken to be a $0-1$ loss. In general, however, it would be preferred to have generalization results for a more general domain adaptation setting, where any arbitrary loss function $\loss$ with some reasonable properties can be considered. In this section, we present a series of results that allow the first theoretical analysis of domain adaptation presented in the previous section to be extended to any arbitrary loss function. 
As we will show, the new divergence measure considered in this section is not restricted to be used exclusively for the task of binary classification, but can also be used for large families of regularized classifiers and regression. Moreover, the results in this section use the concept of the Rademacher complexity, as recalled in Section~\ref{chap:sota}. This particular improvement will lead to data-dependent bounds that are usually tighter that the bounds obtained using the VC theory. 

\paragraph{Discrepancy distance}
We start with the definition of the new divergence measure that was first introduced in \cite{MansourMR09colt}. As they mentioned, its name, the \emph{discrepancy distance}, is due to the relationship between this concept and the discrepancy problems that arise in combinatorial contexts. 
\begin{definition}[\cite{MansourMR09colt}]
Given two domains $\source$ and $\target$ over $\X\times\Y$, let $\Hcal$ be a hypothesis class,
and let $\loss: \Y\times \Y \rightarrow \mathbb{R}_+ $ define a loss function. 
The discrepancy distance $\disc_\loss$ between the two marginals $\sourceX$ and $\targetX$ over $\X$ is defined by
$$\discl{\loss}{\sourceX}{\targetX} = \sup_{(h,h') \in \Hcal^2} \ \abs{ \esp{\xbf\sim\sourceX} \left[\loss\left(h'(\xbf),h(\xbf)\right)\right] -  \esp{\xbf\sim \targetX} \left[\loss\left(h'(\xbf),h(\xbf)\right)\right] }.$$
\end{definition}
We note that the \mbox{$\hdh$-divergence} and the discrepancy distance are related. First, for the $0-1$ loss, we have
\begin{align*}
\discl{\zoloss}{\sourceX}{\targetX}\  =\ \frac12\, d_\hdh(\sourceX,\targetX)\,,
\end{align*}
which shows that in this case the discrepancy distance coincides with the \mbox{$\hdh$-divergence} that appears in Theorems~\ref{trm:bd2} and~\ref{theo:combined}, and it suffers from the same computational restrictions as the latter. 
Furthermore, their tight connection is illustrated by the following proposition.
\begin{restatable}[\cite{MansourMR09colt}]{proposition}{lundisc}
Given two domains $\source$ and $\target$ over $\X\times\Y$, let $\Hcal$ be a hypothesis class, and let $\loss: \Y\times \Y \rightarrow \mathbb{R}_+ $ define a loss function that is bounded, $\forall(y,y')\in\Y^2,\ \loss(y,y') \leq M$ for some $ M > 0$. 
Then, for any hypothesis $h \in \Hcal$, we have 
$$\discl{\loss}{\sourceX}{\targetX}\ \leq\ M\,d_1(\sourceX,\targetX).$$
\end{restatable}
This proposition establishes a link between the seminal results~\cite{bendavidth} presented in the previous section, and shows that for a loss function bounded by $M$, the discrepancy distance can be upper-bounded in terms of the \mbox{$L^1$-distance}. 
\paragraph{Learning bounds} To present a generalization bound, we first need to understand how the discrepancy distance can be estimated from finite samples. To this end,
\cite{MansourMR09colt} proposed the following lemma that bounds the discrepancy distance using the Rademacher complexity (see Section~\ref{sec:rademacher}) of the hypothesis class.
\begin{restatable}[\cite{MansourMR09colt}]{lemma}{mansourun}
Let $\Hcal$ be a hypothesis class, and let $\loss: \Y\times \Y \rightarrow \mathbb{R}_+ $ define a loss function that is bounded, $\forall(y,y')\in\Y^2,\ \loss(y,y') \leq M$ for some $ M > 0$ and let $L_\Hcal = \{\xbf \rightarrow \loss(h'(\xbf), h(\xbf)): h, h' \in \Hcal\}$. Let $\DX$ be a distribution over $\X$, and let $\hat{\mathcal{\D}}_\X$ denote the corresponding empirical distribution for a sample $S = (\xbf_1, \dots , \xbf_m)$. Then, for any $\delta \in (0,1)$, with probability of at least $1 - \delta$ over the choice of sample $S$, we have
\begin{align*}
\discl{\loss}{\DX}{\hat{\mathcal{\D}}_\X} \leq \mathcal{R}_S(L_\Hcal) + 3M \sqrt{\frac{\log\frac{2}{\delta}}{2m}},
\end{align*}
where $\mathcal{R}_S(L_\Hcal)$ is the empirical Rademacher complexity of $L_\Hcal$ based on the observations from $S$.
\label{trm:mansour1}
\end{restatable}
It can be noted that this lemma looks very much like the usual generalization inequalities obtained using the Rademacher complexities presented in Section~\ref{sec:rademacher}. Using this result, we can further prove the following corollary for the case of more general loss functions defined as $\forall (y,y')\in\Y^2,\ \loss_q(y,y') = \vert y-y'\vert^q$ for some $q$. This parametric family of functions is a common choice of a loss function for a regression task.
\begin{restatable}[\cite{MansourMR09colt}]{corollary}{mansourdeux}Let $\source$ and $\target$ be the source and target domains over $\X\times\Y$, respectively.
Let $\Hcal$ be a hypothesis class, and let $\loss_q: \Y\times \Y \rightarrow \mathbb{R}_+ $ be a loss function that is bounded, $\forall(y,y')\in\Y^2,\ \loss_q(y,y') \leq M$ for some $ M > 0$, and defined as $\forall (y,y')\in\Y^2,\ \loss_q(y,y') = \vert y-y'\vert^q$ for some $q$. 
Let $S_u$ and $T_u$ be samples of size $m_s$ and $m_t$ drawn independently from $\sourceX$ and $\targetX$ and let $\hatsourceX, \hattargetX$ denote the empirical distributions corresponding to $\sourceX$ and $\targetX$. Then, for any $\delta \in (0,1)$, with probability of at least $1-\delta$ over the random choice of the samples, we have
\begin{align*}
\discl{\loss_q}{\sourceX}{\targetX} \ \leq\ \discl{\loss_q}{\hatsourceX}{\hattargetX} + 4q\left( \mathcal{R}_{S_u}(\Hcal)+\mathcal{R}_{T_u}(\Hcal)\right)
+ 3M \left( \sqrt{\frac{\log(\frac{4}{\delta})}{2m_s}} + \sqrt{\frac{\log(\frac{4}{\delta})}{2m_t}}\right).
\end{align*}
\label{trm:mnsr2}
\end{restatable}
This result highlights one of the major differences between the approach of~\cite{bendavidth} and that of~\cite{MansourMR09colt}, which arises from the way that they estimate the introduced distance.
While \Cref{trm:bd2} relies on the VC dimension 
to bound the true \mbox{$\hdh$-divergence} by its empirical counterpart, $\disc_\loss$ is estimated using the quantities based on the Rademacher complexity. To illustrate what this implies for the generalization guarantees, we now present the analog of \Cref{trm:bd2}, which relates the source and target error functions using the discrepancy distance, and compare this to the original result.
\begin{restatable}[\cite{MansourMR09colt}]{theorem}{mansourtrois}
Let $\source$ and $\target$ be the source and target domains over $\X\times\Y$, respectively.
Let $\Hcal$ be a hypothesis class, and let $\loss: \Y\times \Y \rightarrow \mathbb{R}_+ $ be a loss function that is symmetric, obeys the triangle inequality, and is bounded, $\forall(y,y')\in\Y^2,\ \loss(y,y') \leq M$ for some $ M > 0$. 
Then, for $h_\source^*\ =\ \argmin{h \in \Hcal}\ \risk^\loss_{\source}(h)$ and $h_\target^*\ =\ \argmin{h \in \Hcal}\ \risk^\loss_{\target}(h)$ denoting the ideal hypotheses for the source and target domains, we have
\begin{align*}
\forall h\in\Hcal,\ \risk^\loss_{\target}(h) \ \leq\   \risk^\loss_{\source}(h,h_S^*) + \discl{\loss}{\sourceX}{\targetX} + \epsilon\,,\\ 
\end{align*}
where $\risk^\loss_{\source}(h,h_\source^*) = \esp{\xbf\sim\sourceX} \loss\left(h(\xbf),h_\source^*(\xbf)\right) \text{ and } \epsilon\ =\  \RT^\loss(h_T^*) + \RS^\loss(h_T^*,h_\source^*).$
\label{trm:mnsr3}\label{trm:mnsr1}
\end{restatable}
\paragraph{Comparison with the $\hdh{}{}$-divergence} As pointed out by the authors, this bound is not directly comparable to \Cref{trm:bd2}, but involves similar terms and reflects a very common trade-off between them. Indeed, the first term of this bound stands for the same source risk function as that in the work of~\cite{bendavidth}. 
The second term here captures the deviation between the two domains through the discrepancy distance similar to the \mbox{$\hdh$-divergence} used before. Finally, the last term $\epsilon$ can be interpreted as the capacity to adapt, and it is very close in spirit to the $\lambda$ term seen previously. 

Despite these similarities, the closer comparison made by \cite{MansourMR09colt} revealed that the bound based on the discrepancy distance can be tighter in some plausible scenarios. For instance, in a degenerate case where there is only one hypothesis $h \in \Hcal$ and a single target function $f_\target$, the bounds of Theorem \ref{trm:mnsr1} and of Theorem \ref{trm:bd2} with true distributions give $\RT^\loss(h,f) +  \discl{\loss}{\sourceX}{\targetX}$ and $\RT^\loss(h,f) + 2\RS^\loss(h,f)+\discl{\loss}{\sourceX}{\targetX}$, respectively. In this case,  the latter expression is obviously larger when $\RS^\loss(h,f) \leq \RT^\loss(h,f)$. The same kind of result can also be shown to hold under the following plausible assumptions:
\begin{enumerate}
\item When $h^* = h_\source^* = h_\target^*$, the bounds of Theorems~\ref{trm:mnsr1} and \ref{trm:bd2} respectively boil down to
\begin{align}
\RT^\loss(h) \leq   \RT^\loss(h^*) + \RS^\loss(h,h^*) + \discl{\loss}{\sourceX}{\targetX}\,,
\label{eq:comparison1a}
\end{align}
and 
\begin{align}
\RT^\loss(h) \leq   \RT^\loss(h^*) + \RS^\loss(h^*) + \RS^\loss(h) + \discl{\loss}{\sourceX}{\targetX}\,,
\label{eq:comparison1b}
\end{align}
where the right-hand side of Equation \ref{eq:comparison1b} includes the sum of three errors and is always larger than the right-hand side of Equation~\ref{eq:comparison1a}, due to the triangle inequality.
\item When $h^* = h_\source^* = h_\target^*$ and $ \discl{\loss}{\sourceX}{\targetX} = 0$, Theorems~\ref{trm:mnsr1} and \ref{trm:bd2} give 
$$\RT^\loss(h) \leq   \RT^\loss (h^*) + \RS^\loss (h,h^*) \qquad \text{ and }\qquad \RT^\loss(h) \leq   \RT^\loss (h^*) + \RS^\loss (h^*) + \RS^\loss (h)\, ,$$
where the former coincides with the standard generalization bound, while the latter does not.
\item Finally, when $f_\target \in \Hcal$, Theorem~\ref{trm:bd2} simplifies to 
$$\vert \RT^\loss(h) - \RS^\loss(h)\vert \leq \discl{\zoloss}{\sourceX}{\targetX}\,,$$
which can be straightforwardly obtained from Theorem~\ref{trm:mnsr1}.
\end{enumerate}
All of these results show a tight link that can be observed in different contributions of the domain adaptation theory. This relation illustrates that the results of ~\cite{MansourMR09colt} strengthen the previous contributions on the subject, but retain a tight connection to them.

\subsection{Generalization bounds based on the discrepancy distance for regression}
\begin{wrapfigure}[13]{r}{.3\linewidth}
    \centering
    \includegraphics[width = \linewidth]{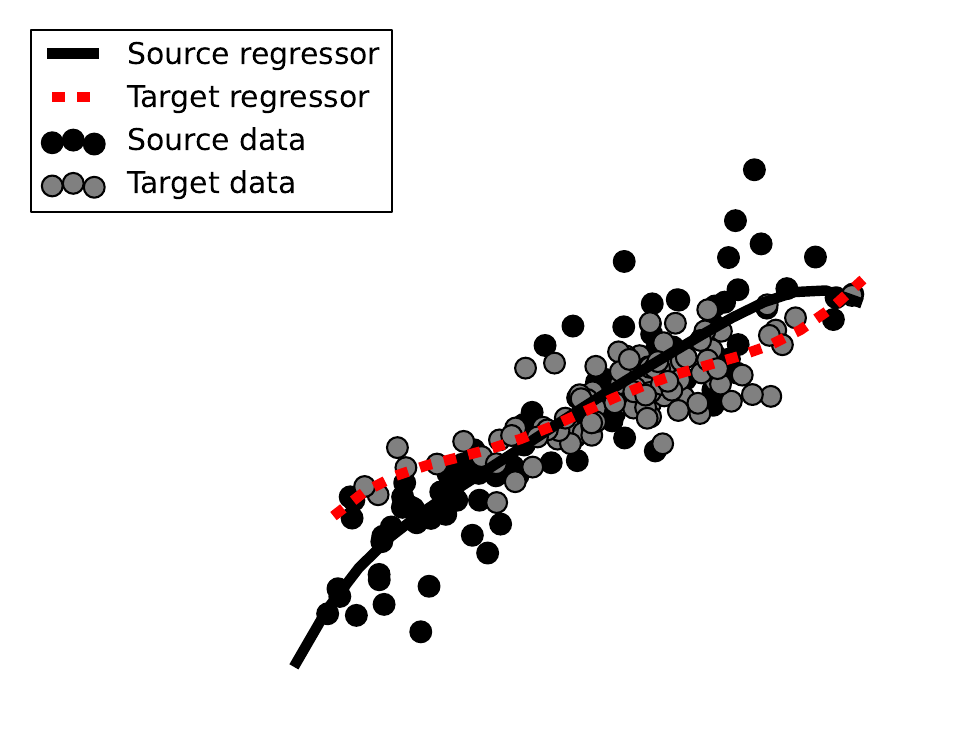}
    \caption{Domain adaptation problem for a regression task.}
    \label{fig:regression_da}
\end{wrapfigure}
As mentioned at the beginning of this section, the discrepancy distance not only extends the first theoretical results obtained for domain adaptation, but also allows new point-wise guarantees to be derived for other learning scenarios, such as, for instance, the regression task, where contrary to classification, the output variable $\Y$ is continuous. The domain adaptation problem for regression is illustrated in \Cref{fig:regression_da}.

To address this scenario, another type of theoretical result based on the discrepancy distance was proposed by~\cite{CortesM11}. These authors considered the case where the hypothesis set $\Hcal$ as a subset of the reproducing kernel Hilbert space (RKHS) $\mathbb{H}$ associated to a positive definite symmetric kernel $K: \Hcal = \lbrace h \in \mathbb{H}: \norm{h}_K \leq \Lambda \rbrace$, where  $\Vert \cdot \Vert_K$ denotes the norm defined by the inner product on $\mathbb{H}$ and $\Lambda \geq 0$. We shall assume that there exists $R > 0$ such that $K(\xbf,\xbf) \leq R^2$ for all $\xbf \in \X$. 
By the reproducing property, for any $h \in \Hcal$ and $\xbf \in \X$, $h(\xbf) = \langle h, K (\xbf, \cdot)\rangle_K$, and thus this implies that $\vert h(\xbf) \vert \leq \Vert h \Vert_K  \sqrt{K(\xbf, \xbf)} \leq \Lambda R$.

In this setting, the authors further presented point-wise loss guarantees in domain adaptation for a broad class of kernel-based regularization algorithms. Given a learning sample $S$, where $\forall (\xbf,y) \in S, \xbf \sim \D_\X,\ y = f_\D(\xbf)$, these algorithms are defined by the minimization of the following objective function:
\begin{align*}
F_{\hat{\D}_\X} (h) = \risk^\loss_{\hat{\D}_\X}(h,f_\D) + \beta  \Vert h \Vert_K^2,
\end{align*}
where $\beta > 0$ is a trade-off parameter.
This family of algorithms includes support vector machines, support vector regression~\cite{Vapnik95}, kernel ridge regression (KRR)~\cite{SaundersGV98}, and many other methods. Finally, the loss function $\loss$ is also assumed to be $\mu$-admissible following the definition given below.
\begin{definition}[$\mu$-admissible loss] 
A loss function $\loss:\Y\times\Y \to \R$ is \mbox{$\mu$-admissible} if it is symmetric and convex with respect to both of its arguments, and for all $\xbf \in \X$ and $y \in \Y$ and $(h, h') \in \Hcal^2$, it verifies the following Lipschitz condition for some $\mu > 0$:
\begin{align*}
\vert \loss(h'(\xbf), y) - \loss(h(\xbf), y) \vert \leq \mu\vert h'(\xbf) - h(\xbf)\vert.
\end{align*}
\end{definition}
The family of $\mu$-admissible losses includes the hinge loss and all $\loss_q(y,y') = \vert y-y'\vert^q$ with $q \geq 1$, in particular the squared loss, when the hypothesis set and the set of output labels are bounded.

With the assumptions made previously, the following results can be proven.
\begin{restatable}[\cite{CortesM11,CortesM14}]{theorem}{thmunregression}
Let $\source$ and $\target$ be the source and target domains on $\X\times\Y$, let $\Hcal$ be a hypothesis class, and let $\loss$ be a \mbox{$\mu$-admissible} loss. We assume that the 
target labeling function $f_\target$ belongs to $\Hcal$, and let $\eta$ denote $\max\lbrace \loss(f_\source(\xbf),f_\target(\xbf)): \xbf \in \support(\hatsourceX)\rbrace$. 
Let $h'$ be the hypothesis that minimizes $F_{\hattargetX}$ and $h$ the one returned when $F_{\hatsourceX}$ is minimized. 
Then, for all $(\xbf,y)\in\X\times\Y$, we have
\begin{align*}
\vert \loss(h'(\xbf), y) - \loss(h(\xbf), y) \vert \leq \mu\, R\, \sqrt{\frac{\discl{\loss}{\hatsourceX}{\hattargetX} + \mu \eta}{\beta}}.
\end{align*}
\label{cortes:thm1_regression}
\end{restatable}
This theorem shows that the difference between the errors achieved by the optimal hypotheses learned on the source and target samples is proportional to the distance between the samples plus a term that reflects the worst value that a loss function can achieve for some instance that belongs to the support of $\hatsourceX$. 

A similar theorem can be proven when $f_\source \in \Hcal$ and not $f_\target$ is assumed. Moreover, the authors indicated that these theorems can be extended to the case where neither the target function $f_\target$ nor $f_\source$ belong to $\Hcal$, by replacing $\eta$ in the statement of the theorem with 
\begin{align*}
\eta' = \underset{\xbf \in \support(\hatsourceX)}{\max}\lbrace \loss(h^*_\target(\xbf),f_\source(\xbf))\rbrace + \underset{\xbf \in \support(\hattargetX)}{\max}\lbrace \loss(h^*_\target(\xbf),f_\target(\xbf))\rbrace,
\end{align*}
where $h^*_\target \in \argmin{h \in \Hcal} \loss (h(\xbf), f_\target)$. In both cases, when $\eta$ is assumed to be small, \ie \ $\eta \ll 1$, the key term of the obtained bound is the empirical discrepancy distance $\discl{\loss}{\hatsourceX}{\hattargetX}$. In the extreme case when $f_\target = f_\source = f$, we obtain $\eta = 0$, and the problem reduces to the covariate shift adaptation scenario that is characterized by the same labeling function in both domains, and is analyzed in more in detail in the following section. In general, a parallel can be drawn between the $\eta$ term that appears in this bound and the other so-called adaptation capacity terms, such as the $\lambda$ term in the bound of Ben-David ~{\textit et al.} from Theorem~\ref{trm:bd2}. 

The result given by Theorem~\ref{cortes:thm1_regression} can be further strengthened when the considered loss function is assumed to be the squared loss $\loss_2= (y-y')^2$ for some $(y, y') \in \Y^2$, and when the kernel-based regularization algorithm described above coincides with the KRR. 
In what follows, the term $\eta$ will be replaced by a finer quantity defined as
\begin{align*}
\delta_\Hcal (f_\source, f_\target) = \inf_{h \in \Hcal} \Vert \esp{\xbf \sim \hatsourceX}\left[\Delta(h, f_\source)\right] -  \esp{\xbf \sim \hattargetX}\left[\Delta(h, f_\target)\right]\Vert,
\end{align*}
where $\Delta(h, f) = \big(f(\xbf)-h(\xbf)\big)\Phi(\xbf)$ with $\Phi(\xbf)$ is  associated to {the} kernel $K$ feature vector, such that $K(\xbf, \xbf') = \langle  \Phi(\xbf), \Phi(\xbf')\rangle$. Using this quantity, the following guarantee holds.
\begin{restatable}[\cite{CortesM14}]{theorem}{thmdeuxkrr}Let $\loss$ be a squared loss bounded by some $M >0$, and let $h'$ be the hypothesis that minimizes $F_{\hattargetX}$, and $h$ the one returned when $F_{\hatsourceX}$ is minimized. Then, for all $(\xbf,y) \in \X\times\Y$, we have:
\begin{align*}
\vert \loss(h(\xbf), y) - \loss(h'(\xbf), y) \vert \! \leq\!
 \frac{R\,\sqrt{M}}{\beta}
\left(\delta_\Hcal (f_\source, f_\target)\! +\! \sqrt{\delta^2_\Hcal (f_\source, f_\target) \!+\! 4\,\beta\,\discl{\loss}{\hatsourceX}{\hattargetX} }\right).
\end{align*}
\label{cortes:thm2_krr}
\end{restatable}
As indicated by the authors, the main advantage of this result is its expression in terms of $\delta_\Hcal (f_\source, f_\target)$ instead of $\eta_\Hcal (f_\source, f_\target)$. It can be noted that $\delta_\Hcal (f_\source, f_\target)$ is defined as a difference, and thus it becomes zero for $\sourceX = \targetX$ , which does not hold for $\eta_\Hcal (f_\source , f_\target)$.
Furthermore, when the covariate-shift assumption holds for some shared labeling function $f$ such that $f_\source = f_\target = f$, $\delta_\Hcal (f, f)$ can be upper-bounded using the following result.
\begin{restatable}[\cite{CortesM14}]{theorem}{thmtroisregression}
Assume that for all $\xbf \in \X$, $K(\xbf,\xbf) \leq R^2$ for some $R > 0$. Let $\mathcal{A}$ denote the union of the supports of $\hatsourceX$  and $\hattargetX$. Then, for any $p>1$ and $q > 1$, with $1/p + 1/q = 1$,
\begin{align*}
\delta_\Hcal (f, f) \leq d_p(f_{\vert \mathcal{A}},\Hcal_{\vert \mathcal{A}})\loss_q(\hatsourceX,\hattargetX),
\end{align*}
\label{cortes:thm3_krr}
where for any set $\mathcal{A} \subseteq \X$, $f_{\vert \mathcal{A}}$ (resp. $\Hcal_{\vert \mathcal{A}}$) denote the restriction of $f$ (resp. $h$) to $\mathcal{A}$ and $d_p(f_{\vert \mathcal{A}},\Hcal_{\vert \mathcal{A}}) = \inf_{h \in \Hcal} \Vert f -  h\Vert_p$.
\end{restatable}
In particular, the authors show that for a labeling function $f$ that belongs to the closure of $\Hcal_{\vert \mathcal{A}}$, $\delta_\Hcal (f)=0$ when the KRR algorithm is used with normalized Gaussian kernels. For this specific algorithm that is often used in practice, the bound of the theorem then reduces to the simpler expression:
\begin{align*}
\vert \loss(h(\xbf), y) - \loss(h'(\xbf), y) \vert\ \leq\ 2\, R\, \sqrt{\frac{M\discl{\loss}{\hatsourceX}{\hattargetX}}{\beta}}.
\end{align*}

\paragraph{Generalized discrepancy} The above-mentioned results can be further strengthened using a recently proposed notation of the generalized discrepancy introduced by \cite{CortesMM15}. To introduce this distance, we can first note that a regression task in the domain adaptation context can be seen as an optimal approximation of an ideal hypothesis $h^*_\target = \argmin{h \in \Hcal} \RT^\loss(h, f_\target)$ by another hypothesis $h$ that ensures the closeness of the losses $\RT^\loss(h^*, f_\target)$ and $\RT^\loss(h, f_\target)$. As we do not have access to $f_\target$, but only to the labels of the source sample $S$, the main idea is to define for any $h \in \Hcal$, a reweighting function
$Q_h: S \rightarrow \mathbb{R}$ such that the objective function $G$ that is defined for all $h \in \Hcal$ by
\begin{align*}
G(h) = \risk^\loss_{Q_h}(h) + \beta  \Vert h \Vert_K^2,
\end{align*}
remains uniformly close to $F_{\hattargetX}(h)$ defined over the target sample $T_u$. As indicated by the authors, this idea introduces a different learning concept, as instead of reweighting the training sample with some fixed set of weights, the weights are allowed to vary as a function of the hypothesis $h$, and are not assumed to sum to 1 or to be nonnegative. Based on this construction, the optimal reweighting can be obtained by solving:
\begin{align*}
&Q_h = \argmin{q \in \mathcal{F}(\sourceX,\mathbb{R})} \vert \risk^\loss_{\hattargetX}(h, f_\target) - \risk^\loss_{q}(h,f_\source)\vert,
\end{align*} 
where $\mathcal{F}(\sourceX,\mathbb{R})$ is the set of real-valued functions defined over $\support (\sourceX)$.

We can note that, in practice, we might not have access to labeled target instances, which implies that we cannot estimate $f_\target$.
To solve this problem, the authors proposed to consider a nonempty convex set of candidate hypotheses $\Hcal'' \subseteq \Hcal$ that can contain a good approximation of $f_\target$. Using $\Hcal''$ as a set of surrogate labeling functions, the previous optimization problem becomes:
\begin{align*}
Q_h = \argmin{q \in \mathcal{F}(\sourceX,\mathbb{R})} \max_{h'' \in \Hcal}\vert \risk^\loss_{\hattargetX}(h, h'') - \risk^\loss_{q}(h,f_\source)\vert.
\end{align*} 
The risk obtained using the solution of this optimization problem given by $Q_h$ can be equivalently expressed as follows:
\begin{align*}
\risk^\loss_{Q_h}(h, f_\source) = \frac{1}{2}\left( \max_{h'' \in \Hcal} \risk^\loss_{\hattargetX}(h, h'')+ \min_{h'' \in \Hcal}\risk^\loss_{\hattargetX}(h, h'')\right).
\end{align*} 
This, in its turn, allows us to reformulate $G(h)$, which can now become:
\begin{align*}
G(h) =  \frac{1}{2}\left( \max_{h'' \in \Hcal} \risk^\loss_{\hattargetX}(h, h'')+ \min_{h'' \in \Hcal}\risk^\loss_{\hattargetX}(h, h'')\right) + \beta  \Vert h \Vert_K^2.
\end{align*}
The proposed optimization problem should have the same point-wise guarantees as those established in Theorem \ref{cortes:thm2_krr}, but based on a new notation of the distance between the probability distributions that can be seen as a generalization of the discrepancy distance used before. To introduce this, we now define $A(\Hcal)$ as a set of functions $U: h \rightarrow U_h$ that map $\Hcal$ to $\mathcal{F}(\sourceX,\mathbb{R})$, such that for all $h \in \Hcal$, $h \rightarrow \loss_{U_h}(h, f_\source)$ is a convex function. The set $A(\Hcal)$ contains all of the constant functions $U$ such that $U_h = q$ for all $h \in \Hcal$, where $q$ is a distribution over $\sourceX$. The definition of the generalized discrepancy can thus be given as follows.
\begin{definition} For any $U \in A(\Hcal)$, the generalized discrepancy between $U$ and $\hattargetX$ is defined as  
$$\text{DISC}(\hattargetX,U) = \max_{h \in \Hcal,h'' \in \Hcal''} \ \abs{ \risk^\loss_{\hattargetX}(h, h'') - \risk^\loss_{U_h}(h, f_\source)}.$$
\end{definition}
In addition, the authors defined the following distance of $f$ to $\Hcal''$ over the support of $\hattargetX$:
\begin{align*}
d_\infty^{\hat{\target}_\X} (f_\target,\Hcal'') = \min_{h_0 \in \Hcal''} \max_{\xbf \in \support (\hattargetX)} \vert h_0(\xbf) - f_\target (\xbf) \vert.
\end{align*}
Using the above-defined quantities, the following point-wise guarantees can be given. 
\begin{restatable}[\cite{CortesMM15}]{theorem}{pointwise}
Let $h^*$ be a minimizer of $\risk^\loss_{\hat{\target}_\X}(h,f_\target) + \beta  \Vert h \Vert_K^2$, and $h_\text{Q}$ be a minimizer of $\risk^\loss_{Q_h}(h, f_\source) + \beta  \Vert h \Vert_K^2$. Then, for $\text{Q}:h \rightarrow Q_h$ and $\forall \xbf \in \X, \ y \in \Y$, the following holds 
\begin{align*}
\vert \loss(h_\text{Q}(\xbf),y) -  \loss(h^*(\xbf),y)  \vert \leq \mu R \sqrt{\frac{\mu d_\infty^{\hat{\target}_\X} (f_\target,\Hcal'') + \text{DISC}(\text{Q},\hattargetX)}{\beta}}.
\end{align*}
\label{thm:pointwise}
\end{restatable}

Furthermore, this inequality can be equivalently written in terms of the risk functions as 
\begin{align*}
\RT^\loss(h_\text{Q},f_\target) \leq \RT^\loss(h^*,f_\target) + \mu R \sqrt{\frac{\mu d_\infty^{\hat{\target}_\X} (f_\target,\Hcal'') + \text{DISC}(\text{Q},\hattargetX)}{\beta}}.
\end{align*}
The result of Theorem~\ref{thm:pointwise} suggests the selection of $\Hcal''$ to minimize the right-hand side of the last inequality. In particular, the authors provided further evidence that if the space over which $\Hcal''$ is searched is the family of all of the balls centered in $f_\source$ defined in terms of $l_{q^*}$, \ie, $\Hcal'' = \{h'' \in \Hcal \vert l_q(h'', f_Q) \leq r \}$ for some distribution $q$ over the space of the reweighted source samples, then the proposed algorithm based on the generalized discrepancy gives demonstrably better results compared to the original algorithm.  

\paragraph{Semi-supervised case} When labeled sample $T$ from the target domain is available, part of it can actually be used to find an appropriate value of $r$. To support this statement, let us consider the following set $S'=S\cup T$ and an empirical distribution $\hatsourceX'$ over it, and use $q'^{*}$ to denote the distribution that minimizes the discrepancy between $\hatsourceX'$ and $\hattargetX$. 
Now, as $\support(\hatsourceX)$ is included in that of $\support(\hatsourceX')$, the following inequality can be obtained
\begin{align*}
 \discl{\loss}{\hattargetX}{q'^{*}}  &=  \min_{\support(q) \subseteq \support(\hatsourceX')} \discl{\loss}{\hattargetX}{q} \\
 &\leq  \min_{\support(q) \subseteq \support(\hatsourceX)} \discl{\loss}{\hattargetX}{q} = \discl{\loss}{\hattargetX}{q^*}.
\end{align*}
Consequently, in view of Theorem~\ref{thm:pointwise}, for an appropriate choice of $\Hcal''$, the learning guarantee for adaptation algorithms based on the generalized discrepancy is more favorable when some labeled data from the target domain are used. Thus, use of the limited amount of labeled points from the target distribution can improve the performance of their proposed algorithm.

\subsection{Other relevant contributions}
\paragraph{\cite{MansourMR08multiple}} In this paper, the authors considered the multi-source domain adaptation problem, and introduced the learning bounds in two different adaptation settings. For the first one, they assumed that $\targetX = \sum_{i=1}^N \alpha_i\sourceX^i$, and studied the performance of a hypothesis defined as $h_\alpha = \sum_{i=1}^N \alpha_i h_i$, where $\sourceX^i$ is the marginal distributions of the $i$\textsuperscript{th} source domain, and $\forall i, \alpha_i\geq0, \ \sum_{i=1}^N\alpha_i=1$. In this scenario, the authors proved that there exists a domain adaptation problem such that $\RT(h_\alpha)=\frac{1}{2}$ even when $\forall i,\ \risk_{\sourceX^i}(h_i)=0$. This prompted them to consider a different combined hypothesis defined as 
$$h_\alpha^\D = \sum_{i=1}^N \frac{\alpha_i\sourceX^i}{\sum_{i=1}^N \alpha_i\sourceX^i}h_i.$$
In this case, the authors proved that $\RT(h_\alpha^\D)\leq \varepsilon$ when $\forall i,\ \risk_{\sourceX^i}(h_i)\leq \varepsilon$.
\paragraph{\cite{MansourMR09}} This work extends the contribution of \cite{MansourMR08multiple} by analyzing arbitrary target distributions that are not necessarily represented by a weighted mixture of source distributions. The authors proposed domain adaptation learning bounds of the following form:
$$\RT(h_\alpha^\D) \leq (\varepsilon d_\alpha(\targetX|\sourceX))^\frac{\alpha-1}{\alpha} M^{\frac{1}{\alpha}},$$
where $d_\alpha(\targetX|\sourceX)=\left(\int_\X\frac{\targetX^\alpha}{\sourceX^{(\alpha-1)}}\right)^\frac{1}{\alpha-1}$ is the exponential of the $\alpha$-R{\'e}nyi divergence, $\risk_{\sourceX^i}(h_i)\leq \varepsilon$, and $M\geq0$ is a constant that bounds the loss function used in the definition of $\risk_\D$.
\paragraph{\cite{HoffmanMZ18}} In this work, the authors extend the analysis of \cite{MansourMR09} to account for cross-entropy and other similar losses not considered in previous work. They also propose a principal way of determining the coefficients $\alpha_i$ ensuring efficient adaptation and extend their analysis to the scenario of non-deterministic labeling.
\paragraph{\cite{DhouibR18}} In this work, the authors proposed a learning bound for hypotheses associated to a general family of similarity functions introduced in \cite{Balcan2008_2}. The proposed bounds rely on $L^1$ and $\chi^2$ divergences and similar to \cite{MansourMR09} present a multiplicative dependence of the source error on the divergence term. 
\paragraph{\cite{RedkoCFT19}} Finally, in this work the authors introduced a bound for the multi-source domain adaptation based on the discrepancy of \cite{MansourMR09colt} for the target shift scenario where the inequality between $\source$ and $\target$ is due to the drift between the marginal distributions of $\Y$ in each domain. 
\paragraph{\cite{KurokiCBHSS19}} This paper proposes source-guided discrepancy (S-disc) that has a virtue of being much easier to estimate in case of $\zoloss$ than the discrepancy proposed by \cite{MansourMR09colt}. The authors also derive a generalization error bound based on S-disc and show that it is never looser than the original bound proposed by \cite{MansourMR09colt}.

\subsection{Summary}
This section presents several cornerstone results of the domain adaptation theory, including those proposed by Ben-David {\it et al.} based on the \mbox{$\hdh$-divergence}, and a variety of results based on the discrepancy distance proposed by Mansour {\it et al.} and Cortes {\it et al.} for the tasks of classification and regression. 
As can be noted, the general ideas used to prove generalization bounds for domain adaptation are based on the definition of a relation between the source and target domains through a divergence that allows us to upper-bound the target risk by the source risk, and on the theoretical results presented in \Cref{chap:sota}, and their properties. Unsurprisingly, this trend is usually maintained regardless of the considered domain adaptation scenario or the learning algorithm analyzed. The overall form of the presented generalization bound on the error of a hypothesis calculated with respect to the target distribution appears to contain, inevitably, the following important terms:
\begin{enumerate}
    \item The source error of the hypothesis measured with respect to some loss function;
    \item The divergence term between the marginal distributions of the source and target domains. In the case of Ben-David {\it et al.}, this term is explicitly linked to the hypothesis space that induces a complexity term that is related to its Vapnik-Chervonenkis dimension; in the case of Mansour {\it et al.} and Cortes {\it et al.}, the divergence term depends on the hypothesis space, but the complexity term is data dependent and is linked to the Rademacher complexity of the hypothesis space; 
    \item The nonestimable term that reflects the {\it a-priori} hardness of the domain adaptation problem. This last usually requires at least some target labeled data to be quantified. 
\end{enumerate}
The terms that appear in the bounds show us that in the case where two domains are almost indistinguishable, the performance of a given hypothesis across these will remain largely similar. When this is not the case, the divergence between the source and target domain marginal distributions starts to have a crucial role in the assessment of the proximity of two domains. For both of the set of results presented, the actual value of this divergence can be consistently calculated using the available finite (unlabeled) samples, thus providing us with a first estimate of the potential success of adaptation. Finally, the last term tells us that even when the divergence between the marginal distributions is taken to zero across two domains, this might not suffice for efficient adaptation. This last point can be summarized by the following statement, as made by Ben-David in~\cite{bendavidth}:
\begin{quotation}
    "When the combined error of the ideal joint hypothesis is large, then there is no classifier that performs well on both the source and target domains, so we cannot hope to find a good target hypothesis by training only on the source domain."
\end{quotation}
This statement brings us to another important question regarding the conditions that need to be verified to make sure that the adaptation is successful. This question stimulates a cascade of other relevant questions, such as what is the actual size of the source and target unlabeled samples needed for the adaptation to be efficient? Are target labeled data needed for an efficient adaptation, and if yes, can we prove formally that it leads to better results? And finally, what are the pitfalls of domain adaptation when even strong prior knowledge regarding the adaptation problem does not guarantee that it has a solution? All these question are answered by the so-called "hardness theorems" that we present in the following section.

\section{Hardness results for domain adaptation}
\label{chap:imposs}
This section is devoted to a series of results that prove the so-called "hardness or impossibility theorems" for domain adaptation. These latter statements show the extent to which the domain adaptation problem can be hard to solve, or the conditions when it is provably unsolvable under some common assumptions. These theorems are very important, as they highlight that in some cases it will not be possible to adapt well even with a prohibitively large amount of data from both domains, or when the adaptation task might be trivial.

\subsection{Problem set-up}
Before presenting the main theoretical results, we first introduce the necessary preliminary definitions that formalize the concepts used afterwards. These definitions are then followed by a set of assumptions that are commonly considered to have a direct influence on the potential success of domain adaptation.

\paragraph{Definitions}
We have seen from the  previous sections that the adaptation efficiency is directly correlated with two main terms that inevitably appear in almost all analyses: one term that depicts the divergence between the domains, and the other term that stands for the existence and the error achieved by the best hypothesis across the source and target domains. 
The authors of \cite{BenDavidLLP10} proposed to analyze the presence of these two terms in the bounds by answering the following questions: 
\begin{enumerate}
    \item Is the presence of these two terms inevitable in the domain adaptation bounds?
    \item Is there a way to design a more intelligent domain adaptation algorithm that uses not only the labeled training sample, but also the unlabeled sample of the target data distribution?
\end{enumerate}
These two questions are very important, as answering them can help us to obtain an exhaustive set of conditions that theoretically ensure efficient adaptation with respect to a given domain adaptation algorithm. 
Before proceeding to the presentation of the main results, the authors first defined several quantities that they used later. 
The first one is the formalization of an unsupervised domain adaptation algorithm \cite{BenDavidLLP10}.
\begin{definition}[domain adaptation learner]
A domain adaptation learner is a function 
\begin{align*}
{\cal A}: \bigcup_{m=1}^\infty \bigcup_{n=1}^\infty (\X \times \{0,1\})^m \times \X^n \rightarrow \{0,1\}^\X.
\end{align*}
\end{definition}
As before, the standard notation for the performance of the learner is given by the error function used. When the error is measured with respect to the best hypothesis in some hypothesis class $\Hcal$, we use the notation $\risk_\D(\Hcal) = \inf_{h \in \Hcal} \risk_\D(h)$. Using this notation, the authors further defined the learnability, as follows.
\begin{definition}[$(\varepsilon, \delta, m, n)$-learnability]
Let $\source$ and  $\target$ be distributions over $\X\times \{0, 1\}$, $\Hcal$ a hypothesis class, ${\cal A}$ a domain adaptation learner, $\varepsilon>0$, $\delta > 0$, and $m$, $n$ positive integers. 
We say that ${\cal A}(\varepsilon, \delta, m, n)$-learns $\target$ from $\source$ relative to $\Hcal$, if when given access to a labeled sample $S$ of size $m$, generated \iid by $\source$, and an unlabeled sample $T_u$ of size $n$, generated \iid\ by $\targetX$, with probability of at least $1-\delta$ (over the choice of the samples $S$ and $T_u$), the learned classifier does not exceed the error of the best classifier in $\Hcal$ by more than $\varepsilon$, \ie, 
\begin{align*}
\Pr{\substack{S \sim (\source)^m\\ T_u \sim (\targetX)^n}} \Big[\risk_\target({\cal A}(S,T_u)) \leq \risk_\target(\Hcal) + \varepsilon \Big]\ \geq\ 1- \delta. 
\end{align*}
\label{def:learnability1}
\end{definition}
This definition gives us a criterion that we can use to judge whether a particular algorithm has strong learning guarantees, which consists in finding an optimal trade-off between both $\varepsilon$ and $\delta$ in the above definition. 
We further introduce two alternative definitions of domain adaptation learnability for the proper learning setting and when the best error of a classifier in $\Hcal$ is scaled by an additional constant $c$. 
\begin{definition}[$(c, \varepsilon, \delta, m, n)$-proper learnability]
With the notations from Definition \ref{def:learnability1}, we say that $A(c, \varepsilon, \delta, m, n)$-solves a proper domain adaptation for the class $\mathcal{W}$ relative to $\Hcal$, if $\Acal$ outputs an element $h$ of $\Hcal$ with 
\begin{align*}
\Pr{\substack{S \sim (\source)^m\\ T_u \sim (\targetX)^n}} \Big[\risk_\target({\cal A}(S,T_u)) \leq c\risk_\target(\Hcal) + \varepsilon \Big]\ \geq\ 1- \delta. 
\end{align*}\label{def:learn_da}
\end{definition}
In other words, this definition says that the proper solving of the domain adaptation problem is achieved when the error of the returned hypothesis \textit{from a fixed hypothesis class} {\it w.r.t.} the target distribution is bounded by $c$ times the error of the best hypothesis on the target distribution plus a constant $\varepsilon$. Obviously, efficient solving of the proper domain adaptation is characterized by small $\delta$ and $\epsilon$, and $c$ close to $1$. We also note that for both of the definitions given above, the inequality event can be reduced to $\risk_\target({\cal A}(S,T_u)) \leq \varepsilon$ when the hypothesis class $\Hcal$ contains a zero-error hypothesis, \ie, $\risk_\target(\Hcal) = 0$.

Finally, we will also need a definition that was introduced in \cite{BenDavidU12} that expresses the capacity of a hypothesis class to produce a zero-error classifier with margin $\gamma$.
\begin{definition}
Let $\X \subseteq \mathbb{R}^d$, $\DX$ be a distribution over $\X$, $h: \X \rightarrow \{0,1\}$ be a classifier, and $B_\gamma(\xbf)$ be the ball of radius $\gamma$ around some domain point $\xbf$. We say that $h$ is a $\gamma$-margin classifier with respect to $\DX$ if for all $\xbf \in \X$ whenever $\DX(B_\gamma(\xbf)) > 0$, then $h(y) = h(z)$ holds for all $y,z \in B_\gamma(\xbf)$.
\end{definition}
In \cite{BenDavidU12}, it was also noted that when $h$ is a \mbox{$\gamma$-margin} classifier with respect to $\DX$, this is equivalent to $h$ satisfying the Lipschitz-property with Lipschitz constant $\tfrac{1}{2\gamma}$ on the support of $\DX$. Thus, we can refer to this assumption as the Lipschitzness assumption. 
For the sake of completeness, we present the original definition of the probabilistic Lipschitzness below.
\begin{definition}
Let $\phi: \mathbb{R} \rightarrow [0, 1]$. 
We say that $f: \X \rightarrow \mathbb{R}$ is \mbox{$\phi$-Lipschitz} with respect to a distribution $\DX$ over $\X$ if, for all $\lambda$ > 0, we have
$$\Pr{\xbf \sim \DX} \Big[\exists \xbf': \vert f(\xbf)-f(\xbf')\vert > \lambda \mu(\xbf,\xbf') \Big] \leq \phi(\lambda),$$
where $\mu:\X \times \X \rightarrow \mathbb{R}_+$ is some metric over $\X$.
\end{definition}

\paragraph{Common assumptions in domain adaptation}  
We now proceed to recall the most common assumptions that were considered in the literature as those that ensure efficient adaptation.
\begin{itemize}
    \item[1.] \textbf{Covariate shift.} This assumption is among the most popular ones, and it has been extensively studied in a series of theoretical studies on the subject (see, for instance, \cite{Sugiyama08directimportance}, and the references therein). While in domain adaptation we generally assume $\source \neq \target$, this can be further understood as $\sourceX(\X)\source(\Y\vert \X) \neq \targetX(\X)\target(\Y \vert \X)$, where  $\source(\Y\vert \X) = \target(\Y \vert \X)$ while $\sourceX \neq \targetX$ is generally called the covariate shift assumption.
    \item[2a.] \textbf{Similarity of the (unlabeled) marginal distributions.} 
    \cite{BenDavidLLP10} considered the \mbox{$\hdh$-distance} between $\sourceX$ and $\targetX$ to assess the impossibility of domain adaptation, and assumed that it remains low between these two domains. This is the most straightforward assumption that directly follows from all of the proposed generalization bounds for domain adaptation. We refer the reader to  Section~\ref{chap:div_based} for the details. 
    \item[2b.] \textbf{Weight-ratio of the (unlabeled) marginal distributions.} The weight-ratio assumption was introduced in \cite{CortesMM10}, and further studied in \cite{BenDavidU12} as a stronger concept of similarity between two marginal distributions. This is defined as: 
    $$C_{\cal B}(\sourceX,\targetX)= \underset{\substack{b\in {\cal B}\\ \targetX(b)\ne 0}}{\inf} \frac{\sourceX(b)}{\targetX(b)}$$
    with respect to a collection of input space subsets ${\cal B}\subseteq 2^\X$.
    \item[3.] \textbf{Ideal joint error.} Finally, this last important assumption is the one that states that there should exist a low-error hypothesis for both domains.
    As explained in Section~ \ref{chap:div_based}, this error can be defined as a so-called $\lambda_\Hcal$ term, as follows:
    $$\lambda_{\Hcal} = \min_{h \in \Hcal} \RS(h)+\RT(h).$$
\end{itemize}
These three assumptions are at the heart of impossibility theorems, where they are usually analyzed in a pair-wise fashion.
\subsection{Constructive impossibility theorems} 
In what follows, we present a series of so-called impossibility results related to the domain adaptation problem. These results are then illustrated based on some concrete examples that highlight the pitfalls of domain adaptation algorithms.

To proceed, we present a theorem showing that some of the intuitive assumptions presented above do not suffice to guarantee the success of domain adaptation.
More precisely, among the three assumptions that have been rapidly discussed -- covariate shift, small \mbox{$\hdh$}-distance between the unlabeled distributions, and the existence of the hypothesis that achieves low error on both the source and target domains (small $\lambda_{\Hcal}$) -- these last two are both necessary (and, as we know from previous results, are also sufficient).
\begin{restatable}[Necessity of a small $\hdh$-distance \cite{BenDavidLLP10}]{theorem}{imposs}
Let $\X$ be some domain set, and $\Hcal$ a class of functions over $\X$. 
Assume that, for some ${\cal A} \subseteq \X$, we have that \mbox{$\{h^{-1}(1)\cap {\cal A}: h \in \Hcal\}$} contains more than two sets and is linearly ordered by inclusion. 
Then, the conditions "covariate shift" plus "small $\lambda_\Hcal$" do not suffice for domain adaptation. 
In particular, for every $\epsilon > 0$, there exists probability distributions $\source$ over $\X \times \{0, 1\}$, and $\targetX$ over $\X$ such that for every domain adaptation learner $\cal A$, every integer $m>0$, $n>0$, there exists a labeling function $f: \X \rightarrow \{0, 1\}$ such that
\begin{enumerate}
\item $\lambda_\Hcal \leq \epsilon$ is small;
\item $\source$ and $\target_f$ satisfy the covariate shift assumption;
\item $\Pr{\substack{S \sim (\source)^m\\ T_u \sim (\targetX)^n}} \left[\risk_{\target_f}({\cal A}(S,T_u)) \geq \tfrac{1}{2} \right] \geq \frac{1}{2}$,
\end{enumerate}
where the distribution $\target_f$ over $\X\times\{0,1\}$ is defined as $\target_f\{1\vert \xbf\in \X\}=f(\xbf)$.
\end{restatable}
This result highlights the importance of the need for small divergence between the marginal distributions of the domains, as even when the covariate shift assumption is satisfied and $\lambda_\Hcal$ is small, the error of the classifier returned by a domain adaptation learner can be larger than $\tfrac{1}{2}$ with a probability that exceeds this same value. 
We now proceed to the symmetric result that shows the necessity for a small joint error between the two domains expressed by the $\lambda_\Hcal$ term.
\begin{restatable}[Necessity for a small $\lambda_\Hcal$ \cite{BenDavidLLP10}]{theorem}{impossdeux}
Let $\X$ be some domain set, and $\Hcal$ a class of functions over $\X$ where the VC dimension is much smaller than $\vert \X \vert$ (for instance, any $\Hcal$ with a finite VC dimension over an infinite $\X$). 
Then, the conditions covariate shift plus small $\hdh$-divergence do not suffice for domain adaption. 
In particular, for every $\epsilon > 0$ there exist probability distributions $\source$ over $\X \times \{0, 1\}$, $\targetX$ over $\X$, such that for every domain adaptation learner $\cal A$, every integer $m,n>0$, there exists a labeling function $f: \X \rightarrow \{0, 1\}$ such that
\begin{enumerate}
\item $\dhdh{\targetX}{\sourceX} \leq \epsilon$ is small;
\item The covariate shift assumption holds;
\item $\Pr{\substack{S \sim \source^m\\ T_u \sim (\targetX)^n}} \Big[\risk_{\target_f}(\mathcal{A}(S,T_u)) \geq \tfrac{1}{2} \Big] \geq \frac{1}{2}.$
\end{enumerate}
\end{restatable}
Once again, this theorem shows that small divergence combined with a satisfied covariate shift assumption can lead to an error of the hypothesis returned by a domain adaptation learner that exceeds $\tfrac{1}{2}$ with high probability. 
Consequently, the main conclusion of these two theorems can be summarized as follows: among the studied assumptions, neither the assumption combination 1. and 3., nor 2a. and 3., suffice for successful domain adaptation in the unsupervised case. Another important conclusion that should be underlined here is that all generalization bounds for domain adaptation with a distance term and a joint error term introduced throughout this survey indeed imply learnability, even with the most straightforward learning algorithm. On the other hand, the covariate shift assumption is not really necessary: it cannot replace any of the other assumptions, and it becomes redundant when the other two assumptions hold. 
This study, however, needs further investigation, as in the case of semi-supervised domain adaptation, the situation can be drastically different. 

\paragraph{Case of proper domain adaptation learning} Below, we turn our attention to the impossibility results established in \cite{BenDavidSU12} for the case where the output of the given domain adaptation algorithm should be a hypothesis that belongs to some predefined hypothesis class. This particular constraint easily justifies itself in practice, where we may need to find a hypothesis as quickly as possible from a predefined set of hypotheses, at the expense of a higher error rate. 
The following result was obtained by \cite{BenDavidU12} in this setting.
\begin{restatable}[\cite{BenDavidSU12}]{theorem}{hardcinq}
Let domain $\X = \left[ 0,1\right]^d$, for some $d$. Consider the class $\Hcal$ of half-spaces as the target class. Let $\xbf$ and $\zbf$ be a pair of antipodal points on the unit sphere, and let $\mathcal{W}$ be a set that contains two pairs $(\source, \target)$ and $(\source', \target')$ of distributions with:
\begin{enumerate}
\item both pairs satisfy the covariate shift assumption;
\item $f(\xbf) = f({\bf z}) = 1$ and $f(\overline{0}) = 0$ for their common labeling function f;
\item $\sourceX(\xbf) = \targetX({\bf z}) = \sourceX(\overline{0}) = \frac{1}{3}$;
\item $\targetX(\xbf)=\targetX(\overline{0})=\frac{1}{2}$ or $\targetX'({\bf z})=\targetX'(\overline{0})=\frac{1}{2}$.
\end{enumerate}
Then, for any number $m$, any constant $c$, no proper domain adaptation learning algorithm can $(c, \varepsilon, \delta, m, 0)$ solve the domain adaptation learning task for $\mathcal{W}$ with respect to $\Hcal$, if $\varepsilon < \frac{1}{2}$ and $\delta < \frac{1}{2}$. In other words, every learner that ignores unlabeled target data fails to produce a zero-risk hypothesis with respect to $\mathcal{W}$.
\end{restatable}
This theorem shows that having some amount of data generated by the target distribution is crucial for the learning algorithm to estimate whether the support of the target distribution is $\xbf$ and $\overline{0}$, or ${\bf z}$ and $\overline{0}$. Surprisingly, there is no possible way to obtain this information without having access to a sample drawn from the target distribution event if the point-wise weight-ratio is assumed to be as large as $\frac{1}{2}$. Thus, no amount of labeled source data can compensate for having a sample from the target marginal distribution.

\paragraph{Illustrative examples}
Now as the main impossibility theorems are stated, it can be useful to give an illustrative example of situations where different assumptions and different learning strategies might fail or succeed. 
To this end, \cite{BenDavidLLP10} considered several examples that showed the inadequacy of the covariate shift assumption explained above, as well as the limits of the reweighting scheme. 

In what follows, the considered hypothesis class is restricted to the space of threshold functions on $[0,1]$, where a threshold function $h_t(\xbf)$ is defined for any $t \in [0,1]$ as $h_t(\xbf)=1$ if $\xbf<t$, and $0$ otherwise. In this case, the set $\hdh$ becomes the class of half-open intervals.

\begin{figure}[!ht]
\begin{center}
   \begin{tikzpicture}
    \draw (0,0) -- (5.75,0);
    \draw (0,-1.1) -- (0,-0.9);

    \fill[black] (0,0) circle (0.6 mm) node[above] {$0$};
    \foreach \i in {2,4,6,8,10} 
      \fill[black] (\i/2,0) circle (0.6 mm) node[above] {$\i\xi$};
      \node[anchor=west, right] at (5.85,0)
        {$y=1$};
      \node[anchor=east] at (-0.2,0)
        {\large $\source$};
    
    \draw (0,-1) -- (5.75,-1);
    \fill[black] (0.5,-1) circle (0.6 mm) node[above] {$\xi$};
    \foreach \i in {3,5,...,11} 
      \fill[black] (\i/2,-1) circle (0.6 mm) node[above] {$\large\i\xi$};
      \node[anchor=west, right] at (5.85,-1)
    {$\large y=0$};
    \node[anchor=east] at (-0.2,-1)
        {$\large\target$};
\end{tikzpicture}
\caption{This scheme illustrates the considered source and target distributions that satisfy the covariate shift assumption with $\xi = \frac{2}{23}$.}
\label{fig:imposs:cov_shift} 
\end{center}
\end{figure}
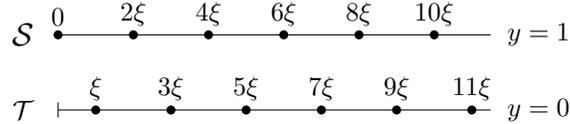

\textbf{Inadequacy of the covariate shift.} 
Let us consider the following construction: for some small fixed $\xi \in \{0;1\}$, let $\target$ be a uniform distribution over $\{2k\xi: k \in \mathbb{N}, 2k\xi \leq 1\} \times \{1\}$, and let the source distribution $\source$ be the uniform distribution over $\{(2k+1)\xi: k \in \mathbb{N}, (2k + 1)\xi \leq 1\} \times \{0\}$. 
The illustration of these distributions is given in Figure~\ref{fig:imposs:cov_shift}.

\noindent{}For this construction, the following holds.
\begin{enumerate}
\item The covariate shift assumption holds for $\target$ and $\source$;
\item The distance $d_\hdh(\source,\target)=\xi$, and thus it can be arbitrarily small; 
\item The errors $\risk_{\source}(\Hcal)$ and $\risk_{\target}(\Hcal)$ are zero;
\item $\lambda_\Hcal(\source,\target) =1 - \xi$ and $\risk_{\target}(h^*_\source) \geq 1-\xi$ are large.
\end{enumerate}
From this example it can instantly be seen that even if the covariate shift assumption is combined with a small \mbox{$\hdh$-divergence} between domains, this still results in a large joint error, and consequently in complete failure of the best source classifier when applied to the target distribution.

\textbf{Reweighting method.} A reweighting method in domain adaptation consists of the determination of a vector of weights $\wbf = \{w_1, w_2, \dots, w_m\}$ that are used to reweight the unlabeled source sample $S_u$ generated by $\sourceX$, to built a new distribution $\target_\wbf^{S_u}$ such that $d_\hdh(\target_\wbf^{S_u}, \targetX)$ is as small as possible. 
In what follows, we denote this reweighted distribution $\target^\source$. 
This new sample is then fed to any available supervised learning algorithm at hand, to produce a classifier that is expected to have a good performance when applied subsequently in the target domain.
As this method has a very important role in the domain adaptation, the authors also gave two intrinsically close examples that show both its success and failure under the standard domain adaptation assumptions. 

\begin{figure}[!ht]
\centering
\begin{tikzpicture}
\draw[->,>=stealth] (0,0)--(5,0) ;
\draw[->,>=stealth] (0,0)--(0,2) ;
\draw[ultra thick] (0,1) -- (3,1);
\draw[ultra thick] (3,1) -- (3,0);
\draw[ultra thick] (3,0) -- (3.5,0);
\draw[ultra thick] (3.5,0) -- (3.5,1);
\draw[ultra thick] (3.5,1) -- (4.5,1);

\node[anchor=west, below] at (0,0) {$0$};
\node[anchor=east] at (-0.2,0.1) {$y=0$};
\node[anchor=east] at (-0.2,1) {$y=1$};
\node[anchor=west, rotate=60, below] at (2.5,-0.4) {$1-3\epsilon$};
\node[anchor=west, rotate=60, below] at (3,-0.4) {$1-\epsilon$};
\node[anchor=west, below] at (4.5,0) {$1$};
\draw (4.5,-0.1) -- (4.5,0.1);
\node[anchor=west, right] at (5,0.3) {$f(\xbf)$};

\draw[ultra thick] (8.5,1) -- (10,1);
\node[anchor=east, above] at (8,1) {$1-\epsilon$};
\node[anchor=east, above] at (10,1) {$1$};
\draw[thick] (8.5,0.9) -- (8.5,1.1);
\draw[thick] (10,0.9) -- (10,1.1);
\node[anchor=west] at (10.2,1) {$\targetX$};

\draw[ultra thick] (6.5,0.1) -- (10,0.1);
\draw[thick] (6.5,0) -- (6.5,0.2);
\draw[thick] (10,0) -- (10,0.2);
\node[anchor=west, below] at (6.5,0.1) {$0$};
\node[anchor=west, below] at (10,0.1) {$1$};

\node[anchor=west] at (10.2,0.1) {$\sourceX$};

\end{tikzpicture}
\caption{Illustration of the reweighting scenario. The source and target distributions satisfy the covariate shift assumption where $f$ is their common conditional distribution. The marginal $\sourceX$ is the uniform distribution over $[0, 1]$, and the marginal $\targetX$ is the uniform distribution over $[1-\epsilon,1].$}
\label{fig:reweight:imposs}
\end{figure}
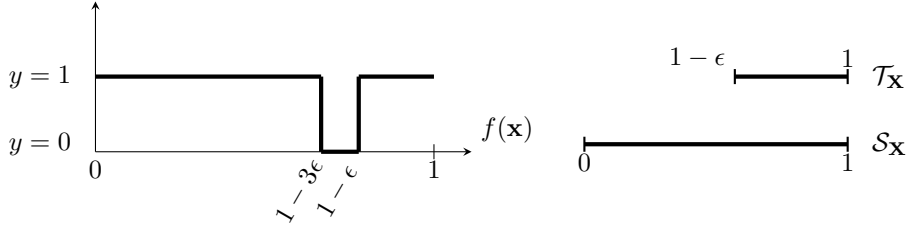

We first consider the following scheme: for some small $\epsilon \in \left(0, \frac{1}{4}\right)$, we assume that the covariate shift assumption holds; \ie, for any $\xbf \in \X$, $\target(y=1\vert \xbf)=\source(y= 1\vert \xbf)=f(\xbf)$. 
We define $f:\X \rightarrow [0,1]$ as follows: for $\xbf \in [1-3\epsilon,1-\epsilon]$, we set $f(\xbf) = 0$, and otherwise we set $f(\xbf) = 1$. 
To define $\source$ and $\target$, we only have to specify their marginals $\sourceX$ and $\targetX$. 
To this end, we let $\sourceX$ be the uniform distribution over $[0,1]$, and we let $\targetX$ be the uniform distribution over $[1-\epsilon, 1]$.
This particular setting is shown in Figure \ref{fig:reweight:imposs}.

\noindent{}The following observations follow from this construction.
\begin{enumerate}
\item For the given construction, the best joint hypothesis that defines $\lambda_\Hcal$ is given by the function $h_{t=1}$;
This function commits $0$ errors on the target distribution and $2\epsilon$ errors on the source distribution, thus giving $\lambda_\Hcal(\source,\target)$ equal to $2\epsilon$.
\item From the definition of \mbox{$\hdh$-divergence}, we obtain that $d_\hdh(\sourceX,\targetX)=1-\epsilon$;
\item $\risk_{\target}(h^*_\source) = 1$, $\risk_{\target}(\Hcal)=0$, and $\risk_{\source}(\Hcal) = \epsilon$ achieved by the threshold functions $h_{t=1-3\epsilon}$, $h_{t=1}$ and $h_{t=1-3\epsilon}$, respectively.
\end{enumerate}
On the other hand, it is possible to find a reweighting distribution that will produce a sample such that $\risk_{\target}(h^*_{\target^\source}) \rightarrow 0$ in the probability when $m$ and $n$ tend towards infinity and $h^*_{\target^\source} = \argmin{h \in \Hcal}{\risk_{\target}(h_{\target^\source}})$. This happens along with the probability of the source error tending to $1$ when $m$ grows to infinity. 
This example is a clear illustration of when a simple reweighting scheme can be efficient for adaptation. This, however, is not the case when we consider different labeling of the target data points. 
Let us now assume that the source distribution remains the same, while for the target distribution $f(\xbf) = 1$ for any $\xbf \in \X$. This slight change gives the following results:
\begin{enumerate}
\item $\lambda_\Hcal(\source,\target) = \epsilon$;
\item $d_\hdh(\sourceX,\targetX)=1-\epsilon$;
\item $\risk_{\target}(h^*_\source) = 0$, $\risk_{\target}(\Hcal)=0$ and $\risk_{\source}(\Hcal) = \epsilon$. 
\end{enumerate}
We can observe that the $\lambda_\Hcal$ term has now become smaller, and that the best source hypothesis achieves a $0$ error on the target distribution. However, the result that we obtain with the reweighting method is completely different: it is not difficult to see that $\risk_{\target}(h^*_{\target^\source}) \rightarrow 1$ in the probability when $m$ and $n$ tend towards infinity, while the error of $h^*_\source$ will tend to zero. 

We conclude by saying that the bound from~\cite{bendavidth} recalled in Section~\ref{chap:div_based}  implies that $\risk_{\target}(h^*_\source)$ is bounded by $\risk_{\target}(\Hcal) + \lambda_\Hcal (\source, \target) + \dhdh{\sourceX}{\targetX}$, and thus it can be hoped that by reweighting the sample $S$ to reflect the distribution $\targetX$, the term $\dhdh{\sourceX}{\targetX}$ in that bound would be diminished. The last example, however, shows that this might not be the case, as $\risk_{\target_\wbf^{L_\X}}$ might be as bad as that bound allows.

\subsection{Impossibility theorems based on sample complexity}
We now present several results that assess the hardness of the domain through the lens of its sample complexity, which is usually defined as the number of training instances required to achieve a low-error classifier for a certain distribution $\D$. This setting in the context of the adaptation problem was studied by \cite{BenDavidU12}, where their first theorem established the sample complexity of solving a domain adaptation problem formulated as follows.
\begin{restatable}[\cite{BenDavidU12}]{theorem}{hardnessLR}
\label{thm:hardnessLR}
For every finite domain $\X$, for every $\varepsilon$ and $\delta$ with \mbox{$\varepsilon + \delta < \tfrac{1}{2}$}, no algorithm can $(\varepsilon,\delta,|S_u|,|T_u|)$-solve the domain adaptation problem for the class $\mathcal{W}$ of triples $(\sourceX, \targetX,f)$ with $C_{\cal B}(\sourceX,\targetX)\geq \frac{1}{2}$, $\dhdh{\sourceX}{\targetX} = 0$, and $\risk_{\target}(\Hcal)=0$ if 
$$|S_u|+|T_u| < \sqrt{(1 - 2(\varepsilon + \delta))\vert \X \vert},$$,
where $\Hcal$ is the hypothesis class that contains only the all-1 and all-0 labeling functions, and $\risk_{\target}(\Hcal) = \underset{h \in \Hcal}{\min} \ \risk_{\target}(h,f)$.
\end{restatable}
This result is interesting in many ways. 
First, the assumptions used in the theorem are extremely simplified, which means that the {\it a-priori} knowledge about the target task is so strong that a zero error classifier for the given hypothesis class can be obtained using only one labeled target instance. 
Secondly, we can also note that the considered setting is extremely favorable for adaptation, as the marginal distributions of the source and target domains are close both in terms of the \mbox{$\hdh$-divergence} and the weight-ratio $C_{\cal B}(\sourceX,\targetX)$. For the latter, this roughly means that the probability to encounter a source point is at least half of the probability of finding it in the target domain. These assumptions further spur the following surprising conclusions:
\begin{enumerate}
\item The sample complexity of domain adaptation cannot be bounded only in terms of the VC dimension of the class that can produce a hypothesis that achieves a zero error on it. This statement agrees well with the previous results, which shows the need for the existence of a good hypothesis for both domains;
\item Some data drawn from the target distribution should be available, to obtain a bound with an exclusive dependency on the VC dimension of the hypothesis class;
\item This result implies that the sample sizes that are needed to obtain useful approximations of the weight-ratio are prohibitively high.
\end{enumerate}
We now provide another result provided by Ben-David and Urner that shows that the same lower bound can be obtained using the Lipschitzness assumption imposed on the labeling function $f$. 
\begin{restatable}[\cite{BenDavidU12}]{theorem}{harddeux}
Let $\X = \left[ 0,1\right]^d, \varepsilon > 0$ and $\delta > 0$ be such that $\varepsilon + \delta < \frac{1}{2}$, let $\lambda > 1$ and let $\mathcal{W}_\lambda$ be the set of triples $(\sourceX, \targetX,f)$  of distributions over $\X$ with $\risk_{\target}(\Hcal)=0$,  $C_{\cal B}(\sourceX,\targetX)\geq \frac{1}{2}$, $\dhdh{\sourceX}{\targetX} = 0$, and \mbox{$\lambda$-Lipschitz} labeling functions $f$. Then no domain adaptation-learner can \mbox{$(\varepsilon,\delta,|S_u|+|T_u|)$-solve} the domain adaptation problem for the class $\mathcal{W}_\lambda$, unless 
$$|S_u|+|T_u| \geq \sqrt{(\lambda+1)^d(1-2(\varepsilon+\delta))}.$$
\end{restatable}

\subsection{Hardness results for sample complexity}
So far we have presented theorems that show what conditions provably lead to the failure of domain adaptation. These results show that even in some extremely simple settings, successful adaptation might require an abundant amount of labeled source data, or at least a reasonable amount of labeled target data. In spite of this, a natural question that might be asked is to what extent the target domain unlabeled data can help to adapt when traded against some labeled source domain data. Before answering this question, we first turn our attention to the sample complexity results presented by \cite{BenDavidSU12}, who investigated the existence of a learning method that can efficiently learn a good hypothesis for a target task provided that the target sample from its corresponding probability distribution is replaced by a (possibly larger) generated sample from a different probability distribution.
The efficiency of such a learning method requires that it does not worsen the generalization guarantee of the learned classifier in the target domain. 
As an example of the considered classifier, we can take a popular nearest-neighbor classifier $h_{\text{NN}}(\xbf)$ that given a metric $\mu$ defined over the input space $\X$, assigns a label to a point $\xbf$ as $h_{\text{NN}}(\xbf) = y(N_S(\xbf))$, with $N_S(\xbf) = \argmindevant{\zbf \in S} \mu(\xbf,\zbf)$ being the nearest neighbor of $\xbf$ in the labeled source sample $S$, and $y(N_S(\xbf))$ is the label of this nearest neighbor.
The theorems obtained are proven under the covariate shift condition and the assumption of a bound on the weight-ratio between the two domains, as explained before. We now present below the first such theorem below.
\begin{restatable}[\cite{BenDavidSU12}]{theorem}{hardquatre}
Let domain $\X = \left[ 0,1\right]^d$ and for some $C > 0$, let $\mathcal{W}$ be a class of pairs of source and target distributions $\{(\source, \target) \vert C_{\cal B}(\sourceX,\targetX)\geq C\}$ with a bounded weight-ratio and their common labeling function $f: \X \rightarrow [0, 1]$, satisfying the $\phi$-probabilistic-Lipschitz property with respect to the target distribution, for some function $\phi$. Then, for all $\lambda$,
$$\esp{S \sim \source^m}\left[ \risk_\target(h_{\text{NN}})\right] \ \leq\  2\risk_{\target}^*(\Hcal) + \phi(\lambda) + 4\lambda\frac{\sqrt{d}}{C}m^{-\frac{1}{d-1}}.$$
\end{restatable}
This theorem suggests that under covariate shift and bounded weight-ratio assumptions, the expected target error of a $\text{NN}$ classifier learned on a sample drawn from the source distribution is bounded by twice the optimal risk over the whole considered hypothesis space, plus several constants related to the nature of the labeling function and the dimension of the input space. Regarding these latter, it can be noted that if the labeling function is $\lambda$-Lipschitz in the standard sense of Lipschitzness, and the labels are deterministic, then we have $\risk_\target^*(\Hcal) = 0$ and $\phi(a) = 0$ for all $a \geq \lambda$. Applying Markov's inequality then yields the following corollary on the sample size bound which further strengthens the previous result.
\begin{corollary} Let domain $\X = \left[ 0,1\right]^d$ and for some $C > 0$, let $\mathcal{W}$ be a class of pairs of source and target distributions $\{(\source, \target) \vert C_{\cal B}(\sourceX,\targetX)\geq C\}$ with a bounded weight-ratio and their common labeling function $f: \X \rightarrow [0, 1]$ satisfying the $\phi$-probabilistic-Lipschitz property with respect to the target distribution, for some function $\phi$. Then, for all $\varepsilon > 0$, $\delta > 0$, $m \geq  \left(\frac{4\lambda\sqrt{d}}{C\varepsilon\delta}\right)^{d+1}$, the nearest neighbor algorithm
applied to a sample of size $m$ has, with probability of at least $1-\delta$, error of at most $\varepsilon$ {\it w.r.t.} the target distribution for any pair $(\source, \target) \in \mathcal{W}$.
\end{corollary}
This corollary provides the first positive result to establish the number of samples required for efficient adaptation in cases where no target data is available to the learner. A natural question that arises is then to quantify the utility of the additional unlabeled target data in the adaptation process, and the conditions required for it to succeed. To answer this question, the authors of \cite{BenDavidU12} considered a particular adaptation algorithm $\cal A$, as summarized below. 
\begin{center}
\vspace{.2cm}
\begin{figure}[!h]
\centering
\begin{minipage}{0.9\textwidth}
\begin{algorithm}[H]
\label{algo:daunlabeled}
\caption{{\bf Input:} An \iid\ sample \mbox{$S_u\!\sim\! \sourceX$} labeled by $f$, an unlabeled \iid\ sample \mbox{$T_u\!\sim\!\targetX$}, and margin parameter $\gamma$.}
\begin{algorithmic}
\STATE \textbf{Step 1}. Partition $[0,1]^d$ into a collection $\cal B$ of boxes (axis-aligned rectangles) with side length $\gamma/\sqrt{d}$.
\STATE \textbf{Step 2}. Obtain sample $S'$ by removing every point in $S_u$, which is sitting
    in a box that is not hit by $T_u$.
\STATE \textbf{Step 3}. Output an optimal risk-minimizing classifier from $\Hcal$ for the sample $S'$.
\end{algorithmic}
\end{algorithm}
\end{minipage}
\end{figure}
\vspace{.2cm}
\end{center}
The following theorem provides lower bounds for both the size of the source labeled and the target unlabeled samples required by algorithm $\cal A$, to learn well when a prior knowledge is assumed to be available to the learner in the form of a hypothesis class that realizes $\targetX$ with margins, as in the definition above. 
\begin{restatable}[\cite{BenDavidU12}]{theorem}{hardtrois}
Let $\X = \left[ 0,1\right]^d$, $\gamma > 0$ be a margin parameter, $\Hcal$ be a hypothesis class of finite VC dimension, and $\mathcal{W}$ be the set of triples $(\sourceX, \targetX,f)$ of source distribution, target distribution, and labeling function with
\begin{enumerate}
\item $C_{\cal I}(\sourceX,\targetX)\geq \tfrac{1}{2}\,$ for the class ${\cal I} = (\hdh)\cap \mathcal{B}$, where $\mathcal{B}$ is a partition of $\left[ 0,1\right]^d$ into boxes of side length $\frac{\gamma}{\sqrt{d}}\,$;
\item $\Hcal$ contains a hypothesis that has $\gamma$-margin on $\target$;
\item the labeling function $f$ is a $\gamma$-margin classifier with respect to $\target$.
\end{enumerate}
Then there is a constant $c>1$, such that for all $\varepsilon > 0$, $\delta > 0$, and for all \mbox{$(\sourceX, \targetX,f) \in \mathcal{W}$}, when given an \iid\ sample $S_u$ from $\sourceX$, labeled by $f$ of size
$$\vert S_u \vert\ \geq\ c\Bigg[\frac{\text{VC}(\Hcal) + \log \frac{1}{\delta}}{C_{\cal I}(\sourceX,\targetX)(1-\varepsilon)\varepsilon}\,  \log\left( \frac{\text{VC}(\Hcal)}{C_{\cal I}(\sourceX,\targetX)(1-\varepsilon)\varepsilon} \right) \Bigg]\,,$$
and an \iid\ sample $T_u$ from $\targetX$ of size 
$$\vert T_u \vert\ \geq\ \frac{1}{\epsilon}\left( 2\left[\frac{\sqrt{d}}{\gamma}\right]^d \ln\left(3\left[\frac{\sqrt{d}}{\gamma}\right]^d\delta\right)\right)\,, 
$$
 then $\Acal$ outputs a classifier $h$ with $\risk_{\target}(h,f) \leq \epsilon$ with probability of at least $1-\delta$.
\end{restatable}
It is worth noting that these bounds follow the standard bounds from statistical learning theory, where the size of the learning sample required for successful learning is given as a function of the VC dimension of the hypothesis class. 
In domain adaptation, this dependency is further extended to the weight-ratio and the accuracy parameters of the learnability model.
Moreover, we observe that this theorem considers the input space that might contain an infinite number of points. 
This assumption can lead to a vacuous bound, as in reality the input space often presents a finite domain, and the dependency of the sample size should be given in its terms. 
The following theorem covers this case.
\begin{theorem}
Let $\X$ be some finite domain, $\Hcal$ be a hypothesis class of finite VC dimension, and $\mathcal{W}= \{(\sourceX, \targetX,f) \vert  \risk_{\target}(\Hcal)=0, C(\sourceX,\targetX)\geq 0\}$ be a class of pairs of source and target distributions with bounded weight-ratio where $\Hcal$ contains the zero-error hypothesis on $\target$. 
Then there is a constant $c>1$, such that for all $\varepsilon > 0$, $\delta > 0$, and all $(\sourceX, \targetX,f) \in \mathcal{W}$, when given an \iid\ sample $S_u$ from $\sourceX$, labeled by $f$ of size
$$\vert S_u \vert \ \geq \ c\Bigg[\frac{\text{VC}(\Hcal) + \log \frac{1}{\delta}}{C(\sourceX,\targetX)(1-\varepsilon)\varepsilon}\, \log\left( \frac{\text{VC}(\Hcal)}{C(\sourceX,\targetX)(1-\varepsilon)\varepsilon} \right) \Bigg]\,,$$
and an \iid\ sample $T_u$ from $\targetX$ of size 
$$\vert T_u \vert \ \geq \  \frac1\epsilon \left(\frac{2\vert \X \vert \ln  3\vert \X \vert}{\delta}\right)\,,
$$
 then algorithm $\cal A$ outputs a classifier $h$ with $\risk_{\target}(h,f) \leq \epsilon$ with probability of at least $1-\delta$.
\end{theorem}
To conclude, we note that both hardness results that state under which conditions the domain adaptation fails, and the results of the analysis of the sample sizes required from the source and target domains for the adaptation to succeed, fall into the category of the so-called impossibility theorems. They essentially draw the limits of the domain adaptation problem under various common assumptions, and provide insights into the hardness of solving this.

\paragraph{The case of agnostic proper domain adaptation}
We presented above an impossibility result for proper domain adaptation that shows that a conservative learner that is fed with a large labeled sample from the source domain might fail to produce a low-error classifier in the target domain, even under high weight-ratio and covariate shift assumptions. Below, we define a two-stage paradigm suggested by \cite{BenDavidSU12} that allows successful learning in this scenario. The proposed two-stage procedure consists of: 1) using a labeled source sample to learn an arbitrary hypothesis with decent performance on the target domain; and 2) applying the learned hypothesis to the unlabeled examples from the target domain, and feeding them to a standard agnostic learner. For the sake of clarity, the definition of an agnostic learning is given below.
\begin{definition}[\cite{BenDavidSU12}]
For $\varepsilon>0, \delta > 0$, $m \in \mathbb{N}$, we say that an algorithm $(\varepsilon, \delta, m)$ (agnostically) learns a hypothesis class $\Hcal$, if for all distributions $\D$, when given an \iid\ sample of size at least $m$, it outputs a classifier of error at most $\risk_\D(\Hcal) + \varepsilon$ with probability of at least $1-\delta$. If the output of the algorithm is always a member of $\Hcal$, we call it an agnostic proper learner for $\Hcal$.
\end{definition}
This definition  can now be used to prove the following theorem for the proposed two-stage procedure.
\begin{restatable}[\cite{BenDavidSU12}]{theorem}{hardsix}
Let $\X$ be some domain and $\mathcal{W}$ be a class of pairs $(\source, \target)$ of distributions over $\X \times \{0, 1\}$ with $\risk_\target(\Hcal) = 0$, such that there is an algorithm ${\cal A}$ and functions $m: (0,1)^2 \rightarrow \mathbb{N}$, $n:(0,1)^2 \rightarrow \mathbb{N}$ such that ${\cal A}(0, \varepsilon, \delta, m(\varepsilon, \delta), n(\varepsilon, \delta))$-solves the domain adaptation learning task for $\mathcal{W}$ for all $\varepsilon, \delta > 0$. Let $\Hcal$ be some hypotheses class for which there exists an agnostic proper learner. Then, the $\Hcal$-proper domain adaptation problem can be $((0, \varepsilon, \delta, m(\varepsilon/3, \delta/2), n(\varepsilon/3, \delta/2))+ m'(\varepsilon/3,\delta/2))$-solved with respect to the class $\mathcal{W}$, where $m'$ is the sample complexity function for agnostically learning $\Hcal$.
\end{restatable}
As in the previous case, the algorithm ${\cal A}$ in the statement of this theorem can be considered to be the nearest neighbor classifier $\text{NN}(\source)$, if the class $\mathcal{W}$ satisfies the conditions from the theorem. To summarize, the presented theorems for the proper domain adaptation learning show that with a domain adaptation algorithm that takes into account the unlabeled instances from the target marginal distribution, it might be possible to solve the proper domain adaptation problem, while in the contrary case, it is provably unsolvable. 

\subsection{Other relevant contributions} 
\paragraph{\cite{RedkoHS19}} In this study, the authors provide a first analysis for consistent estimation of the adaptability term $\lambda$ when some target label data is available. The main construction used in their study is to express the ideal joint hypothesis $h^* = \argmin{h\in \Hcal}{\RS(h,f_\source) +\RT(h,f_\target)}$ as a barycenter of the source and target labeling functions $f_\source$ and $f_\target$. These latter are then considered to be probability measures over $\X$, so that the barycenter is defined over the space of probability distributions without requiring a hypothesis space to be picked in advance.
\paragraph{\cite{CZG19}} In this paper, the authors provide an example similar to that given in \cite{BenDavidLLP10}, to show that small $\Hcal$-divergence between marginal distributions and low source error do not guarantee good performance in the target domain. They further argue that this is mainly explained by the shift in the conditional distributions over the two domains that is accounted for by the inestimable adaptability term. 
\paragraph{\cite{JohanssonSR19}} This paper proceeds in a spirit similar to that of \cite{CZG19}, by first showing an example where finding an invariant representation decreasing the shift between the two domains while minimizing the source error leads to poor performance in the target domain. This is attributed to the unobserved adaptability term and lack of invertability of the learned representation, and it is dealt with by taking into account the performance of a hypothesis in the source domain in regions where the source density is sufficiently high. The authors then provide a tight learning bound based on a weighted source error, a support discrepancy, and an unobservable term that characterizes the invertability of the invariant representation.
\paragraph{\cite{hanneke19}} In this paper, the authors consider a semi-supervised setting where the goal is to learn a hypothesis from a mixture of labeled source and target samples, and to bound the excess risk of this hypothesis, \ie, $\RD(h)-\RD(\Hcal)$ in each domain. The paper further introduces the novel concept of discrepancy between the two domains, called "transfer-exponents", and provides the first minimax-rates, in terms of both source and target sample size and of the latter divergence, similar to the work of \cite{BenDavidU12}.

\subsection{Summary}
In this section, we covered a series of results that establish the conditions required to make a domain adaptation problem solvable. As shown, these necessary conditions might take on different forms, and depend on the value of certain terms presented in the generalization bound and on the size of the available source and target learning samples. The take-away messages of this section can be summarized as follows:
\begin{enumerate}
    \item Solving a domain adaptation problem requires two independent conditions to be fulfilled. First, there is the need to properly minimize the divergence between the source and target marginal distributions. Secondly, there is the need to ensure simultaneously that the {\it a-priori} adaptability of the two domains is high (which is reflected by the small ideal joint error term $\lambda_\Hcal$);
    \item Even under some strong assumptions that make the adaptation problem appear to be easy to solve, there might still be the need for a certain amount of unlabeled source and target data that in the most general case, can be prohibitively large;
    \item A certain amount of labeled source and unlabeled target data can ensure efficient adaptation, and can produce a hypothesis with a small target error. In both cases, this amount depends on the general characteristics of the adaptation problem given by the weight-ratio and the complexity of the hypothesis space represented by its VC dimension;
    \item In proper domain adaptation, ignoring unlabeled target data leads to provably unsolvable adaptation problems, where the domain adaptation learner fails to produce a zero-error hypothesis for the target domain.
\end{enumerate}
All these conclusions provide us with a more general view on the learning properties of the adaptation phenomenon, and essentially provide a list of conditions that need to be verified to make sure that the adaptation problem at hand can be solved efficiently. Apart from that, the established results also provide us with an understanding that some adaptation tasks are harder when compared to others, and that this hardness can be quantified by not one, but several, criteria that take into account both the data distribution and the labeling of instances. Finally, they also show that successful adaptation requires a certain amount of data to be available during the adaptation step, and that this amount might directly depend on the proximity of the marginal distributions of the two domains. This last feature is quite important, as it is added to the dependence on the complexity of the hypothesis class considered previously in the standard supervised learning described in \Cref{chap:sota}.

\section{Learning bounds with integral probability metrics}
\label{chap:ipms}
In the previous sections, we presented several seminal results regarding the generalization bounds for domain adaptation and the impossibility theorems for some of them. We have shown that the basic shape of generalization bounds in the context of domain adaptation remains more or less the same, and mainly differs only in the divergence used to measure the distance between the source and the target marginal distributions. In this section, we consider a large family of metrics on the space of probability measures known as IPMs that present a well-studied topic in probability theory. In particular, we show that depending on the chosen functional class, some instances of IPMs can have interesting properties that are completely different from those shown by both the $\hdh$-divergence and the discrepancy distance seen previously. 

\subsection{Problem set-up}
Integral probability metrics represent a large class of distances defined on the space of probability measures that have found applications in many machine-learning algorithms. The general definition of IPMs can be given as follows. 
\begin{definition}[\cite{zolotarev1984}]
Given two probability measures $\sourceX$ and $\targetX$ defined on a measurable space $\X$, the IPM is defined as 
\begin{align*}
D_{\mathcal{F}}(\sourceX, \targetX)  =\sup_{ f \in \mathcal{F}} \Bigg| \int_\X f d\sourceX - \int_\X f d\targetX  \Bigg|,
\end{align*}
where $\mathcal{F}$ is a class of real-valued bounded measurable functions on $\X$. 
\end{definition}
As mentioned by ~\cite{muller97}, the quantity $D_{\mathcal{F}}(\sourceX, \targetX)$ is a semimetric, and it is a metric if and only if the function class $\mathcal{F}$ separates the set of all signed measures with $\mu(\X) = 0$. It then follows that for any non-trivial function class $\mathcal{F}$, the quantity $D_{\mathcal{F}}(\sourceX, \targetX)$ is zero if $\sourceX$ and $\targetX$ are the same. Several important special cases of IPMs can be obtained by specifically choosing the functional class $\mathcal{F}$. We present those that were used for the analysis of the domain adaptation problem below.

\paragraph{Maximum mean discrepancy} Let $\mathcal{F} =  \{f: \Vert f \Vert_{\mathcal{H}_k} \leq 1 \}$ where $\mathcal{H}_k$ is a RKHS with its associated kernel $k$. Then, the maximum mean discrepancy (MMD) distance is defined as follows:
$$d_{\text{MMD}}(\sourceX,\targetX) =  \underset {\Vert f \Vert_{\mathcal{H}_k} \leq 1} {\sup} \abs{ \int fd(\sourceX-\targetX)} = \Bigg\Vert \int_\X k(\xbf,\cdot) d(\sourceX-\targetX)\Bigg\Vert_{\mathcal{H}_k}.$$
From a practical point of view, we observe that numerous domain adaptation and transfer learning approaches have been based on MMD minimization \cite{PanTKY09,GengTX11,HuangSGBS06,PanKY08,ChenLTW09}, and thus a theoretical analysis of the domain adaptation problem with this is of high scientific interest. 

\paragraph{Wasserstein distance} Let $\mathcal{F} = \{f: \Vert f \Vert_L \leq 1 \}$ where 
$$\Vert f \Vert_{L} = \underset{\xbf \neq \xbf' \in \X} {\sup} \frac{\vert f(\xbf) - f(\xbf') \vert}{c(\xbf,\xbf')}$$
is the Liptschitz semi-norm for real-valued continuous $f$ on $\X$ and some metric $c(\cdot,\cdot): \X \times \X \rightarrow \mathbb{R}_+$.

In this case, the Kantorovich-Rubinstein theorem \cite{opac-b1130971} yields the following result, with the Wasserstein distance $W_1$ defined as follows:
$$W_1(\sourceX,\targetX) =  \underset {\Vert f \Vert_{L} \leq 1} {\sup} \abs{ \int fd(\sourceX-\targetX)} = \inf_{\gamma \in \Pi(\sourceX, \targetX)} \int_{\X \times \X} c(\xbf,\xbf')d\gamma(\xbf,\xbf'),$$
where $\Pi(\sourceX, \targetX)$ is a space of all joint probability measures on $\X \times \X$ with marginals $\sourceX$ and $\targetX$. 

The original optimal transportation problem was introduced by \cite{monge_81} to study the problem of resource allocation. Its modern formulation, which led to the introduction of the Wasserstein distance, is due to \cite{kantorovich}, who proposed a relaxation of the Monge's problem allowing to prove the existence of a unique minimizer for it. Despite being a very powerful tool for comparing and aligning probability distributions, the Wasserstein distance has become an emerging topic in machine learning only recently due to \cite{cuturi:2013}, where an efficient regularization scheme that allowed the solving of the optimal transportation problem was introduced. 

\subsection{Generalization bound with IPMs}
We start this section with a general result that introduces IPMs to the domain adaptation generalization bounds provided by \cite{ZhangZY12}. In this paper, the authors considered a general multi-source scenario where not one, but $K\geq 2$ source domains are available. To be consistent with the rest of the survey, we present the main result of \cite{ZhangZY12} that introduces the IPMs in the context of domain adaptation specified for the case of one source and one target domain below. 
\begin{theorem}
For a labeling function $f \in \mathcal{G}$, let $\mathcal{F} = \{(\xbf,y) \rightarrow \loss(f(\xbf),y)\}$ be a loss function class that consists of the bounded functions with the range $\left[ a,b\right]$ for a space of labeling functions $\cal G$. Let $S = \{(\xbf_1,y_1), \dots, (\xbf_m,y_m) \}$ be a labeled sample drawn from $\source$ of size $m$. Then, given any arbitrary $\xi \geq D_{\mathcal{F}}(\source, \target)$, we have for any $m \geq \frac{8(b-a)}{\xi'^2}$ and any $\epsilon > 0$, with probability of at least $1-\epsilon$, the following holds
\begin{align*}
\sup_{f \in \mathcal{F}} \left\vert \risk^\loss_{\hat{\source}}f - \risk^\loss_{\target}f \right\vert \leq D_{\mathcal{F}}(\source, \target) + \left( \frac{\ln \mathcal{N}_1(\xi'/8, \mathcal{F}, 2m) - \ln(\epsilon/8)}{\frac{m}{32(b-a)^2}}\right)^{\frac{1}{2}},
\end{align*}
where $\xi' = \xi - D_{\mathcal{F}}(\source, \target)$.
\label{trm:zhang_ipm}
\end{theorem}
Here the quantity $\mathcal{N}_1(\xi,\mathcal{F},2m)$ is defined in terms of the uniform entropy number (see \Cref{def:cov_numb}), and it is given by the following equation
\begin{align*}
	\mathcal{N}_1(\xi,\mathcal{F},2m) = \sup_{\{S^{2m}\}} \log N\big(\xi,\mathcal{F},\ell_1(\{S^{2m}\})\big),
	\end{align*}
where for the source sample $S$ and its associated ghost sample $S' = \{(\xbf'_1,y'_1), \dots, (\xbf'_m,y'_m) \}$ drawn from $\source$, the quantity  
$S^{2m} = \lbrace S, S'\rbrace$ and the metric $\ell_1$ are a variation of the $\ell_1$ metric defined for some $f \in \mathcal{F}$ based on the following norm
\begin{align*}
\Vert f\Vert_{\ell_1(\{S^{2m}\}))} = \frac{1}{m}\sum_{i=1}^{m}\Big( \vert f(\xbf_i,y_i)\vert +  \vert f(\xbf'_i,y_i)\Big).
\end{align*}

It can be noted that there are several peculiarities related to this result. First, it is different from other generalization bounds provided before, as the divergence term here is defined for the joint distributions $\source$ and $\target$, and not for the marginal distributions $\sourceX$ and $\targetX$. Note that, in general, the joint target distribution $\target$ cannot be estimated in the classical scenario of unsupervised domain adaptation, as this can be done only when target labels are known, thus making the application of this bound quite uninformative in practice. Secondly, the proposed bound is very general, as it does not specify explicitly the functional class $\cal F$ considered in the definition of the IPM. On the one hand, this allows this bound to be adjusted to any instance of IPMs that can be obtained by choosing the appropriate functional class, while on the other hand, it also requires the uniform entropy number for this to be determined. Finally, the authors established a link between the discrepancy distance seen before and the $D_{\mathcal{F}}(\source, \target)$ that allows us to obtain a bound with a more "traditional" shape. More precisely, the authors proved that the following inequality holds in the case of one source and one target domain for any $\loss$ and functional class $\cal F$:
\begin{align*}
    D_{\mathcal{F}}(\source, \target) \leq \discl{\loss}{\sourceX}{\targetX} + \sup_{g \in \mathcal{G}} \abs{\esp{\xbf \sim \targetX}[\loss(g(\xbf),f_\target(\xbf)] - \esp{\xbf \sim \targetX}[\loss(g(\xbf),f_\source(\xbf)]}.
\end{align*}
Note that the second term of the right-hand side is basically a disagreement between the labeling functions $f_\source$ and $f_\target$ that is zero only when they are equal. Using this inequality, it can be shown that the proposed theorem can be reduced to the following shape:
\begin{equation}
    \sup_{f \in \mathcal{F}} \vert \risk^\loss_{\hat{\source}}f - \risk^\loss_{\target}f \vert \leq \discl{\loss}{\sourceX}{\targetX} + \lambda + \left( \frac{\ln \mathcal{N}_1(\xi'/8, \mathcal{F}, 2m) - \ln(\epsilon/8)}{\frac{m}{32(b-a)^2}}\right)^{\frac{1}{2}},
    \label{eq:ipm_stand}
\end{equation}
where $\lambda = \sup_{g \in \mathcal{G}} \abs{\esp{\xbf \sim \targetX}[\loss(g(\xbf),f_\target(\xbf)] - \esp{\xbf \sim \targetX}[\loss(g(\xbf),f_\source(\xbf)]}$, and the last term is the complexity term that depends on the covering number of the space $\cal F$, similar to the bounds based on the algorithmic robustness presented by \Cref{chap:sota}. To this end, \Cref{eq:ipm_stand} now looks similar to the generalization bounds from the previous sections.

To show that for a finite complexity term the difference between the empirical source risk and the target risk never exceeds the divergence between the two domains with the increasing number of available source examples, the authors proved the following theorem.
\begin{theorem}
For a labeling function $f \in \mathcal{G}$, let $\mathcal{F} = \{(\xbf,y) \rightarrow \loss(f(\xbf),y)\}$ be a loss function class that consists of the bounded functions with the range $\left[ a,b\right]$ for a space of labeling functions $\cal G$. If the following holds
\begin{align*}
\lim_{m \rightarrow \infty}  \frac{\ln \mathcal{N}_1(\xi'/8, \mathcal{F}, 2m)}{\frac{m}{32(b-a)^2}}< \infty\,,
\end{align*}
with $\xi' = \xi - D_{\mathcal{F}}(\source, \target)$, then we have for any $\xi \geq D_{\mathcal{F}}(\source, \target)$,
\begin{align*}
\lim_{m \rightarrow \infty} \Prob{}{\{\sup_{f \in \mathcal{F}} \vert \risk^\loss_{\hat{\source}}f - \risk^\loss_{\target}f \vert > \xi\}} = 0.
\end{align*}
\label{trm:zhang_ipm2}
\end{theorem}
It can be noted here that the probability of event $\{\sup_{f \in \mathcal{F}} \vert \risk^\loss_{\hat{\source}}f - \risk^\loss_{\target}f \vert > \xi\}$ is taken with respect to the threshold $\xi \geq D_{\mathcal{F}}(\source, \target)$, while in standard learning theory this guarantee is usually stated for any $\xi > 0$ given that $\lim_{m \rightarrow \infty} \frac{\ln \mathcal{N}_1(\xi, \mathcal{F}, m)}{m}< \infty$. This highlights an important difference between the classic generalization bounds for supervised learning and the result given by \Cref{trm:zhang_ipm}.

As we mentioned above, the general setting for generalization bounds with IPMs proposed by Zhang {\it et al.} suffers from two major drawbacks: (1) the function class in the definition of the IPM is not specified, making it intractable to compute; (2) the proposed bounds are established for joint distributions rather than marginal distributions, making them not very informative in practice. To this end, we present below two different lines of research that tackle these drawbacks, and establish the generalization bounds for domain adaptation by explicitly considering a particular function class with a divergence term that takes into account the discrepancy between the marginal distributions of the source and target domains. These lines lead to two important particular cases of IPMs that were used to derive generalization bounds in domain adaptation: the Wasserstein distance and the MMD. We take a closer look at both of these in what follows. 

\subsection{Learning bounds with the Wasserstein distance}
Despite many important theoretical insights presented previously, the above-mentioned divergence measures, such as the $\hdh$-divergence and the discrepancy, do not directly take into account the geometry of the data distribution when estimating the discrepancy between two domains. Recently, \cite{courty14workshop} proposed to tackle this drawback by solving the domain adaptation using the Wasserstein distance. To justify domain adaptation algorithms based on the minimization of the Wasserstein distance, the generalization bounds for the three domain adaption settings involving this latter were presented by \cite{RedkoHS16}. According to \cite{Villani09}, the Wasserstein distance is relatively strong and can be combined with smoothness bounds to obtain convergences in other distances. As mentioned by the authors, this important advantage of the Wasserstein distance leads to tighter bounds in comparison to other state-of-the-art results, and it is more computationally attractive, as explained below.

To proceed, let $\mathcal{F} =  \{f \in \mathcal{H}_k: \Vert f \Vert_{\mathcal{H}_k} \leq 1 \}$, where $\mathcal{H}_k$ is a RKHS with its associated kernel $k$. Let $\loss_{h,f}:\xbf \rightarrow \loss(h(\xbf),f(\xbf))$ be a convex loss-function defined $\forall h,f \in \mathcal{F}$, and assume that $\loss$ obeys the triangle inequality. As before, $h(\xbf)$ corresponds to the hypothesis and $f(\xbf)$ to the true labeling functions. Considering that $(h,f) \in \mathcal{F}^2$, the loss function $\loss$ is a non-linear mapping of the RKHS $\mathcal{H}_k$ for the family of $\loss_q$ losses defined previously\footnote{If $(h,f) \in \mathcal{F}^2$ then $h-f \in \mathcal{F}$, which implies that $\loss(h(\xbf),f(\xbf)) = \vert h(\xbf) - f(\xbf) \vert^q$ is a nonlinear transform for $h-f \in \mathcal{F}$.}. Using results from \cite{Saitoh}, it can be shown that $\loss_{h,f}$ also belongs to the RKHS $\mathcal{H}_{k^q}$, admitting the reproducing kernel $k^q$, and that its norm obeys the following inequality:
$$\vert \vert \loss_{h,f} \vert \vert_{\mathcal{H}_{k^q}}^2 \leq \vert \vert h - f \vert \vert_{\mathcal{H}_k}^{2q}.$$
This result gives us two important properties of $\loss_{f,h}$ that are used further:
\begin{enumerate}
\item the function $\loss_{h,f}$ belongs to the RKHS, which allows us to use the reproducing property via some feature map $\phi(\xbf)$ associated to kernel $k^q$;
\item the norm $\vert \vert \loss_{h,f} \vert \vert_{\mathcal{H}_{k^q}}$ is bounded. 
\end{enumerate}
Thus, the error function defined above can be also expressed in terms of the inner product in the corresponding Hilbert space, {\it i.e}\footnote{For simplicity, we further write $\loss$ meaning $\loss_{f,h}$.},
\begin{align*}
    \RD^{\loss} (h, f_\D) &= \esp{\xbf \sim \DX} [\loss(h(\xbf),f_\D(\xbf))]= \esp{\xbf \sim \DX} [\langle\phi(\xbf),\loss\rangle_{\mathcal{H}_{k^q}}].
\end{align*}

Now the following lemma that relates the Wasserstein metric with the source and target error functions for an arbitrary pair of hypotheses can be proved. 
\begin{restatable}[\cite{RedkoHS16}]{lemma}{mmdw}
Let $\sourceX, \targetX \in  \mathcal{P}\left(\X\right)$ be two probability measures on $\mathbb{R}^d$. Assume that the cost function $c(\xbf,\xbf') = \Vert \phi(\xbf) - \phi(\xbf') \Vert_{\mathcal{H}_{k_\loss}}$, where $\mathcal{H}$ is a RKHS equipped with kernel $k_\loss: \X \times \X \rightarrow \mathbb{R}$ induced by $\phi: \X \rightarrow \mathcal{H}_{k_\loss}$ and $k_\loss(\xbf, \xbf') = \langle \phi(\xbf), \phi(\xbf') \rangle_{\mathcal{H}_{k_\loss}}$. Assume further that the loss function $\loss_{h,f}:\xbf \longrightarrow \loss(h(\xbf),f(\xbf))$ is convex, symmetric, bounded, obeys triangle equality, and has the parametric form $\vert h(\xbf) - f(\xbf) \vert^q$ for some $q > 0$. Assume also that the kernel $k_\loss$ in the RKHS $\mathcal{H}_{k_\loss}$ is square-root integrable {\it w.r.t.} both $\sourceX,\targetX$ for all  $\sourceX,\targetX \in \mathcal{P}(\X)$ where $\X$ is separable and $0\leq k_\loss(\xbf,\xbf') \leq K, \forall \ \xbf,\xbf' \in \X$. If $\norm{\loss}_{\mathcal{H}_{k_\loss}}\leq 1$, then the following holds
$$\forall (h,h')\in \mathcal{H}_{k_\loss}^2,\quad  \RT^{\loss_q}(h,h')\leq \RS^{\loss_q}(h,h') + W_1(\sourceX,\targetX).$$
\label{trm:mmd_w}
\end{restatable}
This lemma makes use of the Wasserstein distance to relate the source and target errors. The assumption made here is to specify for the cost function that $c(\xbf,\xbf') = \Vert \phi(\xbf) - \phi(\xbf') \Vert_{\mathcal{H}}$. While it might appear too restrictive, this assumption is, in fact, not that strong. Using the properties of the inner-product, we have
\begin{align*}
\Vert \phi(\xbf) - \phi(\xbf') \Vert_{\mathcal{H}} &= \sqrt{\langle \phi(\xbf) - \phi(\xbf'), \phi(\xbf) - \phi(\xbf') \rangle_{\mathcal{H}}} = \sqrt{k(\xbf,\xbf) -2k(\xbf,\xbf')+k(\xbf,\xbf')}.
\end{align*}

As the authors noted, it is possible to further show that for any given positive-definite kernel $k$ there is a distance $c$ (used as a cost function in our case) that generates this, and {\it vice versa} (see Lemma 12 from \cite{sejdinovic2013equivalence}).
 
The following generalization bound was proven by the authors using a result that showed the convergence of the empirical measure $\hat{\mu}$ to its true associated measure {\it w.r.t.} the Wasserstein metric provided by \cite{BolleyGV07}.
\begin{theorem}
Under the assumptions of Lemma \ref{trm:mmd_w}, let $S_u$ and $T_u$ be two samples of size $N_S$ and $N_T$ drawn {\it i.i.d.} from $\sourceX$ and $\targetX$, respectively. Let $\hatsourceX = \frac{1}{N_S}\sum_{i=1}^{N_S} \delta_{\xbf_i^S}$ and $\hattargetX = \frac{1}{N_T}\sum_{i=1}^{N_T} \delta_{\xbf_i^T}$ be the associated empirical measures. Then for any $d'>d$ and $\varsigma' < \sqrt{2}$, there exists some constant $N_0$ depending on $d'$, such that for any $\delta > 0$ and $\min(N_S,N_T) \geq N_0 \max(\delta^{-(d'+2)},1)$ with probability of at least $1-\delta$ for all $h$, we have
\begin{align*}
\RT^{\loss_q}(h)\leq \RS^{\loss_q}(h) &+ W_1(\hatsourceX, \hattargetX) + \sqrt{2\log\left(\frac{1}{\delta}\right)/\varsigma'}\left(\sqrt{\frac{1}{N_S}}+\sqrt{\frac{1}{N_T}}\right) + \lambda \,,
\end{align*}
where $\lambda$ is the combined error of the ideal hypothesis $h^*$ that minimizes the combined error of $\RS^{\loss_q}(h)+\RT^{\loss_q}(h)$.
\label{trm:wass2}
\end{theorem}
A first immediate consequence of this theorem is that it justifies the use of the optimal transportation in the domain adaptation context when combined with the minimization of the source error, and assuming the joint error given by the $\lambda$ term is small. For this latter, \cite{courty14workshop} proposed a class-labeled regularization term added to the original optimal transport formulation to restrict source examples of different classes to be transported to the same target example, by promoting group sparsity in the matrix $\gamma$ due to $\Vert \cdot \Vert^p_q$ with $q = 1$ and $p = \frac{1}{2}$. In some way, this regularization term influences the capability term, by ensuring the existence of a good hypothesis that will be discriminant on both source and target domain data. 

\paragraph{Semi-supervised case}
To remain consistent with the previous sections, we also provide the generalization bound for the Wasserstein distance in the semi-supervised setting below. 
\begin{restatable}[\cite{RedkoHS16}]{theorem}{irdeux}
Let $S_u$, $T_u$ be unlabeled samples of size $N_S$ and $N_T$ each, drawn independently from $\sourceX$ and $\targetX$, respectively. 
Let $S$ be a labeled sample of size $m$ generated by drawing $\beta\, m$ points from $\targetX$ ($\beta \in [0,1]$) and $(1-\beta)\,m$ points from $\sourceX$ and labeling them according to $f_\source$ and $f_\target$, respectively. If $\hat{h} \in \Hcal$ is the empirical minimizer of $\RSemp^\alpha(h)$ on $S$ and $h_T^* = \argmin{h \in \Hcal} \RT^{\loss_q}(h)$, then for any $\delta \in (0,1)$ with probability of at least $1-\delta$ (over the choice of samples),
$$\RT^{\loss_q}(\hat{h})\ \leq\ \RT^{\loss_q}(h_T^*) + c_1 + 2(1-\alpha)(W_1(\hatsourceX, \hattargetX)+ \lambda + c_2),$$
where
\begin{align*}
&c_1 = 2 \sqrt{\frac{2K\left(\frac{(1-\alpha)^2}{1-\beta}+\frac{\alpha^2}{\beta}\right)\log(2/\delta)}{m}} +4 \sqrt{K/m} \left( \frac{\alpha}{m\beta \sqrt{\beta} } + \frac{(1-\alpha)}{m(1-\beta)\sqrt{1-\beta} }\right),\\
&c_2 = \sqrt{2\log\left(\frac{1}{\delta}\right)/\varsigma'}\left(\sqrt{\frac{1}{N_S}}+\sqrt{\frac{1}{N_T}}\right).
\end{align*}
\label{trm:our4}
\end{restatable}
In line with the results obtained previously, this theorem shows that the best hypothesis that takes into account both source and target labeled data (\ie, $0 \leq \alpha < 1 $) performs at least as good as the best hypothesis learned on target data instances alone ($\alpha = 1$). This result agrees well with the intuition that semi-supervised domain adaptation approaches should be at least as good as unsupervised ones.

\subsection{Generalization bound with MMD}
Based on the results with the Wasserstein distance, we now introduce learning bounds for the target error where the divergence between the task distributions is measured by the MMD distance. As before, we start with a lemma that relates the source and target errors in terms of the introduced discrepancy measure for an arbitrary pair of hypotheses. Then, we show how the target error can be bounded by the empirical estimate of the MMD plus the complexity term.  
\begin{lemma}[\cite{Redko15}] Let $\mathcal{F} =  \{f \in \mathcal{H}_k: \Vert f \Vert_{\mathcal{H}_k} \leq 1 \}$ where $\mathcal{H}_k$ is a RKHS with its associated kernel $k$. Let $\loss_{h,f}:\xbf \rightarrow \loss(h(\xbf),f(\xbf))$ be a convex loss-function with a parametric form $\vert h(\xbf) - f(\xbf) \vert^q$ for some $q > 0$, and defined $\forall h,f \in \mathcal{F}$ such that $\loss$ obeys the triangle inequality. Then, if $\Vert l \Vert_{{\mathcal{H}_{k^q}}} \leq 1$, we have :
\begin{align*}
\forall (h,h')\in{\cal F},\quad \RT^{\loss_q} (h , h')\leq \RS^{\loss_q} (h , h') + d_{\text{MMD}}(\mathcal{\sourceX},\mathcal{\targetX}).
\end{align*}
\label{trm:our1}
\end{lemma}
This lemma is proved in a similar way to \Cref{trm:mmd_w} from \cite{RedkoHS16}, as presented before in this section. Using this and the result that relates the true and the empirical MMD distances \cite{Song08}, we can prove the following theorem.
\begin{theorem}
With the assumptions from \Cref{trm:our1}, let $S_u$ and $T_u$ be two samples of size $m$ drawn {\it i.i.d.} from $\sourceX$ and $\targetX$, respectively. Then, with probability of at least $1-\delta (\delta \in (0,1))$ for all $h \in \mathcal{F}$, the following holds:
\begin{align*}
\RT^{\loss_q} (h) \leq \RS^{\loss_q} (h) + d_{\text{MMD}}(\hatsourceX,\hattargetX)+ \frac{2}{m}\left( \esp{\xbf \sim \sourceX} \left[ \sqrt{\tr(K_{\source})} \right] + \esp{\xbf \sim \targetX} \left[ \sqrt{\tr(K_{\target})} \right]\right) + 2 \sqrt{\frac{\log(\frac{2}{\delta})}{2m}} + \lambda,
\end{align*} 
where $d_{\text{MMD}}(\hatsourceX,\hattargetX)$ is an empirical counterpart of $d_{\text{MMD}}(\sourceX,\targetX)$, $K_\source$ and $K_\target$ are the kernel functions calculated on samples from $\sourceX$ and $\targetX$, respectively, and $\lambda$ is the combined error of the ideal hypothesis $h^*$ that minimizes the combined error of $\RS^{\loss_q}(h)+\RT^{\loss_q}(h)$.
\label{trm:our2}
\end{theorem}
We can see that this theorem is similar in shape to \Cref{trm:wass2} and \Cref{trm:bd2}. The main difference, however, is that the complexity term does not depend on the Vapnik-Chervonenkis dimension. In our case, the loss function between two errors is bounded by the empirical MMD between distributions and two terms that correspond to the empirical Rademacher complexities of $\mathcal{H}$ {\it w.r.t.} the source and target samples. In both theorems, $\lambda$ has the role of the combined error of the ideal hypothesis. Its presence in the bound comes from the use of triangle inequality for the classification error. 

This result is particularly useful, as an unbiased estimate of the squared MMD distance $d^2_{\text{MMD}}(\hatsourceX,\hattargetX)$ can be calculated in linear time. We also note that the bound obtained can be further simplified with the use of, for instance, Gaussian, exponential or Laplacian kernels, to calculate the kernel functions $K_{\mathcal{S}}$ and $K_{\mathcal{T}}$, as these have 1s on the diagonal, thus facilitating the calculation of the trace. Finally, it can be seen that the bound from Theorem~\ref{trm:our2} has the same terms as Theorem~\ref{trm:bd2}, while the MMD distance is estimated as in \Cref{trm:mnsr2}.

\paragraph{Semi-supervised case}
Similar to the case considered by \cite{bendavidth}, we can also derive similar bounds for the MMD distance in the case of combined error. 
To this end, we present the following analog of Theorem \ref{theo:combined}. 
\begin{theorem}
With the assumptions from \Cref{trm:our1}, let $S_u$, $T_u$ be unlabeled samples of size $m'$, each drawn independently from $\sourceX$ and $\targetX$, respectively. 
Let $S$ be a labeled sample of size $m$ generated by drawing $\beta\, m$ points from $\targetX$ ($\beta \in [0,1]$) and $(1-\beta)\,m$ points from $\sourceX$, and labeling them according to $f_\source$ and $f_\target$, respectively. If $\hat{h} \in \Hcal$ is the empirical minimizer of $\risk^\alpha(h)$ on $S$ and $h_T^* = \argmin{h \in \Hcal} \RT^{\loss_q}(h)$, then for any $\delta \in (0,1)$, with probability of at least $1-\delta$ (over the choice of samples),
$$\RT^{\loss_q}(\hat{h})\ \leq\ \RT^{\loss_q}(h_T^*) + c_1 + c_2,$$
\begin{align*}
c_1 &= 2 \sqrt{\frac{2K\left(\frac{(1-\alpha)^2}{1-\beta}+\frac{\alpha^2}{\beta}\right)\log\frac{2}{\delta}}{m}}+2\left( \sqrt{\frac{\alpha}{\beta}} + \sqrt{\frac{1-\alpha}{1-\beta}}\right)\sqrt{\frac{K}{m}}\,,\\
c_2 &= \hat{d}_{\text{MMD}}(S_u, T_u) + \frac{2}{m'}\, \esp{\xbf \sim \sourceX}\!\! \sqrt{\tr(K_{\source})} + \frac{2}{m'}\, \esp{\xbf \sim \targetX}\!\! \sqrt{\tr(K_{\target})} + 2 \sqrt{\frac{\log\frac{2}{\delta}}{2m'}} + \lambda.
\end{align*}
\label{trm:our5}
\end{theorem}
Several observations can be made from this theorem. First of all, the main quantities that define the potential success of domain adaptation according to \cite{bendavidth} (\ie, the distance between the distributions and the combined error of the joint ideal hypothesis) are preserved in the bound. This is an important point that indicates that the two results are not contradictory or supplementary. Secondly, rewriting the approximation of the bound as a function of $\alpha$ and omitting additive constants can lead to a similar result as for Theorem~\ref{theo:combined}. This observation might indicate the existence of a strong connection between these. 

The generalization guarantees obtained for domain adaptation based on the MMD distance allow another step forward to be made in domain adaptation theory, and the results presented in the previous sections to be extended in two different ways. Similar to discrepancy-based results, the bounds with the MMD distance allow any arbitrary loss function to be considered, and thus applications of domain adaptation other than binary classification can be studied. On the other hand, similar to the entropic-regularized Wasserstein distance, the MMD distance has some very useful estimation guarantees that are unavailable for both the $\hdh$ and $\disc$ divergences. This feature can be very important in accessing both the \textit{a-priori} hardness of adaptation and its \textit{a-posteriori} success, to understand whether a given adaptation algorithm manages to correctly reduce the discrepancy between the domains. 

\subsection{Relationship between the Wasserstein and the the MMD distances}
Here, we have just presented two results that introduced the Wasserstein and the MMD distances to the domain adaptation generalization bounds for both semi-supervised and unsupervised cases. As both results are {built} on the same construction, there might be the need to explore the link between the Wasserstein and the MMD distances. To do this, we first observe that in some particular cases, the {latter} can be bounded by the {former}. 
Indeed, if we assume that the ground metric in the Wasserstein distance is $c(\xbf,\xbf') = \Vert \phi(\xbf)-\phi(\xbf')\Vert_\mathcal{H}$, then the following results can be obtained: 
\begin{align*}
\Bigg\Vert \int_{\X} f d(\sourceX - \targetX) \Vert_\mathcal{H} & = \Bigg\Vert \int_{\X \times \X} (f(\xbf) - f(\xbf')) d\gamma(\xbf,\xbf') \Bigg\Vert_\mathcal{H}\\
& \leq \int_{\X \times \X} \Vert f(\xbf) - f(\xbf') \Vert_{\mathcal{H}} d\gamma(\xbf,\xbf')\\
& = \int_{\X \times \X} \Vert \left\langle f(\xbf), \phi(\xbf)\right\rangle - \left\langle f(\xbf'), \phi(\xbf')\right\rangle \Vert_{\mathcal{H}} d\gamma(\xbf,\xbf')\\
& \leq \Vert f \Vert_{\mathcal{H}} \int_{\X \times \X} \Vert \phi(\xbf) - \phi(\xbf') \Vert_{\mathcal{H}} d\gamma(\xbf,\xbf').
\end{align*}
Now taking the supremum over $f$ {\it w.r.t.} $\mathcal{F} =  \{f: \Vert f \Vert_{\mathcal{H}} \leq 1 \}$, and the \textit{infimum} over $\gamma \in \Pi(\sourceX, \targetX)$, this gives
\begin{align}
    d_{\text{MMD}}(\sourceX, \targetX) \leq W_1(\sourceX,\targetX).  
    \label{eq:mmd_wass}
\end{align}
This result holds under the hypothesis that $c(\xbf,\xbf') = \Vert \phi(\xbf)-\phi(\xbf')\Vert_\mathcal{H}$. On the other hand, in \cite{gao2014minimum}, the authors showed that $W_1(\sourceX,\targetX)$ with this particular ground metric can be further bounded, as follows
$$W_1(\sourceX,\targetX) \leq \sqrt{d^2_{\text{MMD}}(\sourceX, \targetX)+C},$$
where $C =\Vert \mu[\sourceX]\Vert_\mathcal{H} + \Vert \mu[\targetX] \Vert_\mathcal{H}$. This result is quite strong for multiple reasons. First, it allows the squared MMD distance to be introduced to the domain adaptation bounds using \cite[Lemma 1]{RedkoHS16}, which leads to the following result for two arbitrary hypotheses $(h, h') \in \mathcal{H}^2$
$$ \RT(h , h') \leq \RS (h , h') + \sqrt{d^2_{\text{MMD}}(\sourceX, \targetX)+C}.$$

\noindent On the other hand, the unified inequality 
\begin{equation}
    d_{\text{MMD}}(\sourceX, \targetX) \leq W_1(\sourceX,\targetX) \leq \sqrt{d^2_{\text{MMD}}(\sourceX, \targetX)+\Vert \mu[\sourceX]\Vert_\mathcal{H} + \Vert \mu[\targetX] \Vert_\mathcal{H}}
    \label{eq:ineq:mmd_wass}
\end{equation}
suggests that the MMD distance establishes an interval bound for the Wasserstein distance. This point is very interesting, because originally the calculation of the Wasserstein distance (also known as the Earth Mover's distance) requires the solving of a linear programming problem that can be quite time consuming due to the computational complexity of $\mathcal{O}(n^3\log(n))$, where $n$ is the number of instances. 

This result, however, is true only under the assumption that $c(\xbf,\xbf') = \Vert \phi(\xbf)-\phi(\xbf')\Vert_\mathcal{H}$. While in most applications, the Euclidean distance $c(\xbf,\xbf') = \Vert \xbf - \xbf'\Vert$ is used as a ground metric, this assumption can represent an important constraint. Luckily, it can be circumvented due to the duality between the RKHS-based and distance-based metric representations studied by \cite{sejdinovic2013equivalence}). Let us first rewrite the ground metric as
\begin{align*}
\Vert \phi(\xbf)-\phi(\xbf') \Vert_{\mathcal{H}} = \sqrt{\langle \phi(\xbf)-\phi(\xbf'), \phi(\xbf)-\phi(\xbf') \rangle_{\mathcal{H}}} = \sqrt{k(\xbf,\xbf) -2k(\xbf,\xbf')+k(\xbf',\xbf')}.
\end{align*}
Now, to obtain the standard Euclidean distance in the expression of the ground metric, we can pick a kernel given by the covariance function of the fractional Brownian motion, \ie, 
$k(\xbf,\xbf') = \frac{1}{2}(\Vert \xbf\Vert^2 + \Vert \xbf'\Vert^2 - 2\Vert \xbf - \xbf'\Vert^2).$
Inserting this expression into the definition of $c(\xbf,\xbf')$ gives the desired Euclidean distance, and thus allows the Wasserstein distance to be calculated with the standard ground metric. 

\subsection{Other relevant contributions} 
\paragraph{\cite{zhang_bridging_2019}} In this work, the authors generalized the seminal bounds to the multi-class setting, and introduced a classification margin $\beta>0$ into their results. This was done by introducing a definition of the error function $\RD^\beta$ that takes into account the classification margin, as follows:
$$\RD^\beta = \esp{\xbf \sim \D}{[l^\beta(h(\xbf),f_\D(\xbf))]},$$
where$l^\beta$ is the ramp loss (\cite[Section 15.2.3]{shalev2014understanding}), defined as:
\begin{equation}\label{eq:ramp_loss}
l^{\beta}(t) := \left\{  \begin{array}{r l} 1-\frac{t}{\beta}, & \text{ if } 0\leq t\leq \beta\\
     																			\text{[}t<0\text{]}, & \text{ otherwise}\end{array} \right.
\end{equation}

Their main contribution for the case of binary classification with labels encoded in $\{-1,1\}$ can then be stated as follows:
 \begin{align}
     \RT(h) \leq \RS^\beta(h) + \sup_{h' \in \Hcal}\abs{\RS^\beta(\sgn{h},h')-\RT^\beta(\sgn{h},h')} + \lambda^{(\beta)},
     \label{zhang_divergence}
 \end{align}
 where 
 $$\lambda^{(\beta)} = \inf_{h \in \Hcal} \RS^\beta(h) + \RT^\beta(h).$$
The alignment term in Equation \eqref{zhang_divergence} was termed the margin disparity discrepancy. As can be noted, this involves a supremum over one hypothesis instead of two, making it lower than $\Hcal \Delta \Hcal$-divergence defined previously, which corresponds to the case of $\beta = 0$ with the definition of the error given above. This also offers new insights into the domain adaptation problem, by introducing the margin violation rate and scoring functions that give the confidence level of belonging to a class of interest, rather than functions with binary output. However, as they bound the 0-1 loss on the target domain, \ie, $\err{0,0}{\target}{h,f}$, their bound does not indicate the behavior of the margin violation rate on this latter. For $\lambda^{(\beta)}$, this remains conceptually similar to the $\lambda$ term of the other bounds, with the only difference consisting in the definition of the error terms. 

\paragraph{\cite{dhouib_margin_2020}} This work provides a generalization bound using a translated version of the ramp loss given in Equation \eqref{eq:ramp_loss} and defined as $l^{\rho,\beta} := l^\beta(\cdot - \rho)$ for some $\rho>0$. The authors first prove a bound that is analogous to \Cref{zhang_divergence}, but concerning the margin violation loss $\RT^{\rho, 0}(h)$ on the target domain, as follows:
 \begin{align}
     \RT^{\rho, 0}(h) \leq \RS^{\frac{\rho + \beta}{\alpha},0}(h) + \sup_{h' \in \Hcal'}\abs{\RS^{\rho,\beta}(h,h')-\RT^{\rho,\beta}(h,h')} + \lambda^{(\alpha)},
     \label{dhouib_divergence}
 \end{align}
where 
$$\lambda^{(\alpha)} = \inf_{h \in \Hcal'} \RS(h) + \RT(h) + \Prob{\xbf \sim \sourceX}{\left[\abs{h(\xbf)} < \alpha\right]}.$$
Compared to the bound from \Cref{zhang_divergence}, this bound is more informative on the separation quality between classes in the target domain, assessed by the margin violation risk $\RT^{\rho, 0}(h)$. Also, the divergence term is continuous in both $h$ and $h'$ for $\beta>0$, which makes it more suitable for optimization algorithms. The non estimable term $\lambda^{(\alpha)}$ is non symmetric w.r.t to $\target$ and $\source$ as it involves an absolute margin violation risk only for $\sourceX$. Finally, hypothesis space $\Hcal'$ used to define the divergence and the $\lambda^{(\alpha)}$ term on the one hand, and the one concerning $h$, \ie $\Hcal$, are not necessarily equal.

\paragraph{\cite{ShenQZY18,CourtyFHR17}} Several studies have presented generalization bounds for domain adaptation based on the Wasserstein distance, similar to those presented in this section. To this end, \cite{ShenQZY18} gave a learning bound with the exact same form as the bound in \Cref{trm:wass2}, but without imposing any additional assumptions on the ground metric used in the definition of the Wasserstein distance. On the other hand, \cite{CourtyFHR17} proposed a learning bound for an adaptation scenario between joint source and target probability distributions $\source$ and $\target$, similar to that of \cite{ZhangZY12}. Their bound introduced $W(\source,\target)$ with an additional term related to the probabilistic transfer Lipschitzness assumption introduced in the latter paper for the labeling function with respect to the optimal coupling. Also, the work of \cite{dhouib_margin_2020} mentioned above proposed a generalization DA bound with an adversarial (minimax) version of the Wasserstein distance between the marginal distributions analyzed extensively in \cite{dhouib_advOT_2020}. 

Finally, we also note that the study of \cite{JohanssonSR19} mentioned in the previous section also introduces learning bounds for domain adaptation based on the concept of IPM. 

\subsection{Summary}
In this section, we presented several theoretical results that use IPMs as a measure of divergence between the marginal source and the target domain distributions in the domain adaptation generalization bounds. We argued that this particular choice of a distance provides a number of advantages compared to the $\hdh$-distance and the discrepancy distances considered before. First, both the Wasserstein distance and the MMD distance can be calculated from available finite samples in a computationally attractive way, due to linear time estimators for their entropy-regularized and quadratic versions, respectively. Secondly, the Wasserstein distance allows geometrical information to be taken into account when calculating the divergence between the two domain distributions, while the MMD distance is calculated based on the distance between the embeddings of two distributions in some (possibly) richer space. This feature is relatively interesting, as it provides more flexibility when it comes to incorporating the prior knowledge into the domain adaptation problem on the one hand, and allows a potentially richer characterization of the divergence between the domains, on the other. This might explain the abundance of domain adaptation algorithms based on the MMD distance, and some recent domain adaptation techniques developed based on optimal transportation theory. Finally, we note that in general, the presented bounds are similar in shape to those described in \Cref{chap:div_based}, and they preserve their main terms, thus remaining consistent with these. This shows that despite the large variety of ways that can be used to formally characterize the generalization phenomenon in domain adaptation, the intuition behind this process and the main factors defining its potential success remain the same.  

\section{PAC-Bayesian theory for domain adaptation}
\label{chap:dalc}
In this section, we recall the results from \cite{dalc,pbda,long_pbda}, where PAC-Bayesian theory was used to theoretically understand domain adaptation through the weighted majority vote learning point of view. 

\subsection{Problem set-up}
In the traditional PAC-Bayesian setting, we consider a $\prior$ distribution over the hypothesis set $\Hcal$, and the objective is to learn a $\posterior$ distribution over $\Hcal$, by taking into account the information captured by the learning sample $S$.
In the domain adaptation setting, the goal is different, and it consists of learning the \mbox{$\posterior$-weighted} majority vote 
$$\forall \xbf\in\X,\quad \BQ(\xbf) = \sign{\esp{ h \sim \posterior} h(\xbf)},$$, 
with the best performance on the target domain $\target$. Note that, here, we consider the $0-1$ loss function.
As in the nonadaptation setting, PAC-Bayesian domain adaptation generalization bounds do not directly upper-bound $\risk^\zoloss_\target(\BQ)$, but upper-bound the expectation according to $\posterior$ of the individual risks of the functions from $\Hcal$: $\espdevant{h \sim \posterior} \risk^\zoloss(h)$,
which is closely related to $\BQ$ (see Equation~\eqref{eq:gibbsrelation}).
Let us introduce a tight relation between $\RD(\BQ)$ and $\espdevant{h \sim \posterior} \risk^\zoloss(h)$, known as the C-bound \cite{Lacasse06}, and defined for all distribution $\D$ on $\XY$ as
\begin{align}
\label{eq:cbound}
\risk^\zoloss_\D(\BQ)\ \leq\ 1 - \frac{\displaystyle \left(1-2 \esp{h \sim \posterior}\risk^\zoloss_\D(h)\right)^2}{\displaystyle 1-2 \dq}.
\end{align}
where $$\dq\ =\ \esp{(h,h')\sim\posterior^2} \esp{\xbf\sim\DX} \zoloss\big(h(\xbf),h'(\xbf)\big)$$
is the expected disagreement between pairs of voters on the marginal distribution $\DX$.
It is important to highlight that the expected disagreement $\dq$ is closely related to the concept of expected joint error $\eq$ between pairs of voters:
$$\eq\ =\  \esp{(h,h')\sim\posterior^2} \esp{(\xbf,y)\sim \D}\zoloss\big(h(\xbf),y)\big)\times \loss_{0-1}\big(h'(\xbf),y)\big).
$$
Indeed, for all distribution $\D$ on $\XY$, we have
\begin{align}
\label{eq:rde}
\esp{h \sim \posterior}\risk^\zoloss_\D(h)\ = \ \frac12 \dq +\eq.
\end{align}
In the following, we present the two PAC-Bayesian generalization bounds for domain adaptation presented in \cite{pbda,dalc}, through the point of view of  \cite{catoni2007pac}.

\subsection{In the spirit of Ben-David et al. and Mansour et al.}
The authors of \cite{pbda} proposed to define a divergence measure that follows the idea underlying the C-bound of Equation~\eqref{eq:cbound}.
More precisely, if $\espdevant{h\sim\posterior}\risk^\zoloss_\source(h)$ and $\espdevant{h\sim\posterior}\risk^\zoloss_\target(h)$ are similar, then $\risk^\zoloss_\source(\BQ)$ and $\risk^\zoloss_\target(\BQ)$ are similar when $\dqS$ and $\dqT$ are also similar.
Thus, the domains $\source$ and $\target$ are close according to $\posterior$ if the expected disagreement over the two domains tends to be close.
This intuition led the authors to the following domain disagreement pseudometric.
\begin{definition}[Domain disagreement~\cite{pbda}]
\label{def:disagreement}
Let $\Hcal$ be a hypothesis class.
For any marginal distributions $\sourceX$ and $\targetX$ over~$\X$, and any distribution $\posterior$ on $\Hcal$, the domain disagreement $\disPB{\sourceX}{\targetX}$ between~$\sourceX$ and~$\targetX$ is defined by
\begin{align*}
\disPB{\sourceX}{\targetX}  \ \eqdef  \ 
\Big|\,\dqT  - \dqS \,\Big|.
\end{align*}
\end{definition}
It is worth noting that the value of $\disPB{\sourceX}{\targetX}$ is always lower than the \mbox{$\hdh$-distance} between $\sourceX$ and $\targetX$. 
Indeed, for every $\Hcal$ and $\posterior$ over $\Hcal$, we have
\begin{align*}
\tfrac{1}{2}\, \dhdh{\sourceX}{\targetX} \ & =\ \sup_{\substack{(h,h')\in\mathcal{H}^2}} \left|
\esp{\xbf\sim\sourceX} \zoloss\big(h(\xbf),h'(\xbf)\big) - \esp{\xbf\sim\target} \zoloss\big(h(\xbf),h'(\xbf)\big)
\right| \\
&\geq\ 
\esp{(h,h')\sim\posterior^2} \left| \esp{\xbf\sim\sourceX} \zoloss\big(h(\xbf),h'(\xbf)\big) - \esp{\xbf\sim\target} \zoloss\big(h(\xbf),h'(\xbf)\big)
\right|\\
&\geq \Big| \, \dqT  - \dqS \, \Big|\\
&=\    \disPB{\sourceX}{\targetX}.
\end{align*}
Using this domain divergence, the authors proved the following domain adaptation bound.
\begin{restatable}[\cite{pbda}]{theorem}{pbda}
\label{thm:pacbayesdabound}
\label{theo:pbda}
Let ${\cal H}$ be a hypothesis class. We have 
\begin{align*}
\nonumber \forall \posterior&\mbox{ on }\Hcal,\  \esp{h\sim\posterior}\risk^\zoloss_\target(h) \ \leq \  \esp{h\sim\posterior}\risk^\zoloss_\source(h) +  \frac{1}{2}\disPB{\sourceX}{\targetX} + \lambda_\posterior\,, 
\end{align*}
where $\lambda_\posterior$ is the deviation between the expected joint errors between pairs for voters on the target and source domains, defined as
 \begin{align}\label{eq:lambda_rho}
 \lambda_\posterior
 \ =\ 
 \Big|\, \eqT - \eqS \,\Big|.
\end{align}
\end{restatable}
The above theorem can be used to prove different kinds of PAC-Bayesian generalization bounds. Below, we present only one such generalization bound, which was used to derive an adaptation algorithm in~\cite{pbda}. 
\begin{theorem}\label{theo:PBcatoni_pbda}
 For any domains $\source$ and $\target$ over $\XY$, any set of voters $\Hcal$, any prior distribution $\prior$ over $\Hcal$, any $\delta \in (0,1]$, any real numbers $\omega > 0$ and $a > 0$, with a probability of at least $1-\delta$ over the random choice of $S \times  T_u  \sim (\source \times  \targetX)^m $, for every posterior distribution $\posterior$ on $\Hcal$, we have
 \begin{align*} 
 \esp{h\sim\posterior}\risk^\zoloss_\target(h)  \ \leq\  &\omega'\,  \esp{h\sim\posterior}\risk^\zoloss_S(h)   +  a'\, \tfrac{1}{2}\, \disPB{S}{T_u}\\
 & +  \left( \frac{\omega'}{\omega} + \frac{a'}{a} \right)  \frac{\KL{\posterior}{\prior}+\ln\frac{3}{\delta}}{m} + \lambda_\posterior + \tfrac{1}{2} (a' -  1)
   \,,
 \end{align*}
where $\disPB{S}{T_u}$ is the empirical estimate of the the domain disagreement;
$\lambda_\posterior$ is defined by Equation~\eqref{eq:lambda_rho}; $\displaystyle \omega'\eqdef\tfrac{\omega}{1 -e^{-\omega}}$ \, and \, $\displaystyle a'\eqdef \tfrac{2a}{1 -e^{-2a}}$.
\end{theorem}

Similarly to the bounds of Theorems~\ref{trm:bd1} and~\ref{trm:mnsr1}, this bound can be seen as a trade-off between different quantities. 
The terms $\espdevant{h\sim\posterior}\risk^\zoloss_S(h)$ and $\disPB{S}{T}$ are akin to the first two terms of the bound of Theorem~\ref{trm:bd1}: $\espdevant{h\sim\posterior}\risk^\zoloss_S(h)$ is the \mbox{$\posterior$-average} risk over $\Hcal$ on the source sample, and $\disPB{S}{T_u}$ measures the \mbox{$\posterior$-average} disagreement between the marginals, although it is specific to the current model depending on $\posterior$. The last term $\lambda_\posterior$ measures the deviation between the expected joint target and source errors of the individual hypothesis from $\Hcal$ (according to $\posterior$). A successful domain adaptation is possible if this deviation is low, although when no labels in the target sample are available, this term cannot be controlled or estimated. 

Despite the same underlying philosophy, the authors note that {this} bound is in general incomparable with those ones of Theorems~\ref{trm:bd1} and~\ref{trm:mnsr1} due to the dependence of $\disPB{S}{T}$ and $\lambda_\posterior$ on the learned posterior.

\subsection{A different philosophy}
In \cite{dalc}, the authors introduce another domain divergence to provide an original bound for the PAC-Bayesian setting.
{They} take advantage of Equation~\eqref{eq:rde}, which expresses the risk of the Gibbs classifier in terms of two {quantities}:
\begin{align} \label{eq:rde_target}
		  \esp{h\sim\posterior}\risk^\zoloss_\target(h)  \ =\ \tfrac12\dqT + \eqT\,
\end{align}
It can be noted that the latter expression consists of half of the expected disagreement, which does not require labeled data to be estimated, and the inestimable expected joint error. To deal with the latter, the authors designed a divergence to link $\eqT$ to $\eqS$, called the \mbox{$\beta$-divergence}, which is defined by 
\begin{equation} \label{eq:bq}
\forall q>0,\quad \bq \ = \ \left[\,\esp{(\xbf,y)\sim \source} \left(  \frac{\target(\xbf,y)}{\source(\xbf,y)} \right)^q\, \right]^{\frac1q}.
\end{equation}
The \mbox{$\beta$-divergence} is parametrized by the value of $q>0$, and allows well-known distribution divergence to be recovered, such as the \mbox{$\chi^2$-distance} and the R{\'e}nyi divergence mentioned at the end of Section \ref{chap:div_based}.
When $q\to \infty$, we have
\begin{equation} \label{eq:binf}
\binf   \ =  \sup_{(\xbf,y)\in\scriptsupport(\source)} \left(  \frac{\target(\xbf,y)}{\source(\xbf,y)} \right) ,\end{equation}
where $\support(\source)$ denotes the support of the domain $\source$.
This \mbox{$\beta$-divergence} leads to the following bound.
\begin{restatable}[\cite{dalc}]{theorem}{dalc}
\label{theo:dalc}
		Let $\Hcal$ be a hypothesis space, $\source$ and $\target$ be the source and target domains on $\XY$, and $q>0$ be some positive constant. Then, for all posterior distributions $\posterior$ on $\Hcal$, we have
		\begin{align*} 
 \esp{h\sim\posterior}\risk^\zoloss_\target(h)  \ \leq \  \frac12 \,\dqT +
\bq{\times}
\Big[ \eqS \Big]^{1-\frac1q}
+ \eta_{\target\setminus\source},
		\end{align*}
		where $$\eta_{\target\setminus\source} = \Pr{(\xbf,y)\sim \target}  \Big((\xbf,y)\notin \support(\source)\Big) \ \sup_{h\in\Hcal} \risk_\TminusS(h)\,$$
		with $\TminusS$ the distribution of $(\xbf,y){\sim}\target$ conditional to $(\xbf,y)\in\support(\target){\setminus}\support(\source)$.
\end{restatable}
The last term of the bound, $\eta_{\target\setminus\source}$, which cannot be estimated without target labels, captures the worst possible risk for the target area not included in $\support(\source)$, similar to the idea used by \cite{JohanssonSR19}.
Note that we have $$\eta_{\target\setminus\source}\leq \Pr{(\xb,y)\sim \target}\big((\xb,y)\notin \support(\source)\big).$$ 
An interesting property of Theorem~\ref{theo:dalc} is that when domain adaptation is not required (\ie, $\source = \target$), the bound is still sound and nondegenerate. Indeed, in this case we have
\begin{align*} 
\RS(\GQ) \ =\ \RT(\GQ) \ &\leq\  \tfrac12\, \dqT +  1\times\left[\eqS\right]^1+0 =\  \tfrac12\, \dqS +  \eqS\ =\ \RS(\GQ)\, .
		\end{align*}
Below, we present the PAC-Bayesian generalization bound obtained from the above theorem for the case $q{\to}\infty$. 
\begin{theorem}
\label{theo:PBcatoni_dalc}
 For any domains $\source$ and $\target$ over $\XY$, any set of voters $\Hcal$, any prior distribution $\prior$ over $\Hcal$, any $\delta \in (0,1]$, any real numbers $b > 0$ and $c > 0$,  with a probability of at least $1-\delta$ over the random choices of $S\sim(\source)^\ms$ and $T_u  \sim (\targetX)^\mt $, for every posterior distribution $\posterior$ on $\Hcal$, we have
		\begin{align*}
	 \esp{h\sim\posterior} \risk^\zoloss_\target(h) \ \leq\ &c'\,\tfrac12\,\dqTemp  + b'\,\eqSemp + \eta_{\target\setminus\source}+ \left(\frac{c'}{\mt\times c} + \frac{b'}{\ms\times b} \right)     \left(2\,\KL{\posterior}{\prior} + \ln \tfrac{2}{\delta}\right)\, ,
		\end{align*}
where $\dqTemp$ and $\eqSemp$ are the empirical estimations of the target voters' disagreement and the source joint error, and $\displaystyle b'=\tfrac{b}{1-e^{-b}}\,\binf$, and $\displaystyle c'=\tfrac{c}{1-e^{-c}}$.
	\end{theorem}
Similarly to the first bound, the above theorem upper-bounds the target risk by a trade-off of different terms given by the following atypical quantities:
\begin{enumerate}
\item The expected disagreement $\dqTemp$ that captures second degree information about the target domain;
\item The divergence between the domains, captured by the $\beta_q$-divergence is not an additive term any more: it weights the influence of the expected joint source error $\eqSemp$ where the parameter $q$ allows different instances of the $\beta_q$-divergence to be considered;
\item The term $\eta_{\target\setminus\source}$ quantifies the worst feasible target error on the regions where the source domain is not informative for the target task.
\end{enumerate}

\subsection{Comparison of the two domain adaptation bounds}
The main difference between the bounds of Theorems~\ref{theo:pbda} and~\ref{theo:dalc} lies in the estimable terms that the latter relies on.
In Theorem~\ref{theo:dalc}, the nonestimable terms are the \mbox{$\beta$-divergence} $\bq$ and the term $\eta_{\TminusS}$. 
Contrary to the noncontrollable term $\lambda_\posterior$ of Theorem~\ref{theo:pbda}, these terms do not depend on the \emph{learned} posterior distribution~$\posterior$: for every $\posterior$ on $\Hcal$, $\bq$ and $\eta_{\TminusS}$ are constant
values that measure the relation between the domains for the considered task. 
Moreover, the \mbox{$\beta$-divergence} is not an additive term but a multiplicative one (as opposed to {\small$\disPB{\sourceX}{\targetX} +  \lambda_\posterior$} in Theorem~\ref{theo:pbda}), which is an important contribution of this new perspective. This is similar to the studies of \cite{MansourMR09} and \cite{DhouibR18}, who also introduced such a multiplicative dependence.
Consequently, $\bq$ can be viewed as a hyperparameter, which allows us to tune the trade-off between the target voters' disagreement $\dqT$ and the source joint error $\eqS$. 

Note that, when $\eqT \geq \eqS$, we can upper-bound the term $\lambda_\posterior$ of Theorem~\ref{theo:pbda} by using the same trick as in the proof of Theorem~\ref{theo:dalc}. 
This leads to
\begin{align*}
\eqT  \,\geq\, \eqS   \quad \Longrightarrow \quad \lambda_\posterior\ &=\  \eqT  - \eqS  \leq\ \bq \times \big[ \eqS  \big]^{1-\frac1q}  + \eta_{\TminusS} - \eqS .
\end{align*}
Thus, in this particular case, we can rewrite the Theorem~\ref{theo:pbda} statement for all $\posterior$ on $\Hcal$, as
\begin{align*}
\esp{h\sim\posterior}\risk^\zoloss_\target(h) \leq  \esp{h\sim\posterior}\risk^\zoloss_\source(h) + \frac{1}{2}\disPB{\sourceX}{\targetX} +
\bq \times  \big[ \eqS  \big]^{1-\frac1q}  - \eqS  
{+} \eta_{\TminusS}.
\end{align*}
It turns out that, if $\dqT   \geq \dqS$
in addition to $\eqT  \geq \eqS $,
the above statement reduces to that of Theorem~\ref{theo:dalc}. 
In all other cases, Theorem~\ref{theo:dalc} is tighter, thus confirming that following the seminal works of Section~\ref{chap:div_based}, with introduction of absolute values in Theorem~\ref{theo:pbda}, gives a very rough approximation. Finally, one of the key points of the generalization bounds of Theorems~\ref{theo:PBcatoni_pbda} and~\ref{theo:PBcatoni_dalc} is that they suggest algorithms for tackling majority vote learning in the domain adaptation context.
Similar to what was done in traditional supervised learning~\cite{LangfordS02,AmbroladzePS06}, \cite{pbda,dalc,long_pbda} specialized these theorems to linear classifiers, and derived adaptation algorithms based on this specialization. 

\subsection{Other relevant contributions}
\paragraph{\cite{McNamaraB17}} In this study, the authors made use of the PAC-Bayesian framework to derive a generalization bound for fine tuning in deep learning in a spirit close to that of analysing a domain adaptation problem. Their considered setting corresponded to a scenario where there is the need to adapt a network trained for a given domain to a similar one. The authors obtained a bound that does not directly involve the concept of divergence between the domains, but a function that measures a transferability property between the two domains.

\subsection{Summary}
In this section, we recalled the two domain adaptation analyses for the PAC-Bayesian framework presented in \cite{pbda,dalc,long_pbda} for models taking the form of a majority vote over a set of classifiers. More precisely, the first result of this section follows the underlying philosophy of the seminal works of Ben-David {\it et al.} and Mansour {\it et al.} of Section~\ref{chap:div_based}, by upper-bounding the target risk by a source risk and a domain divergence measure suitable for the PAC-Bayesian setting. This divergence is expressed as the average deviation between the disagreement over a set of classifiers on the source and target domains, contrary to \mbox{$\hdh$-divergence} and discrepancy distance, which are defined in terms of the worst-case deviation. Then, we recalled another domain adaptation bound that takes advantage of the inherent behavior of the target risk in the PAC-Bayesian setting. 
The upper bound obtained is different from the original one, as it expresses a trade-off between the disagreement on the target domain only, the joint errors of the classifiers on the source domain only, and a term that reflects the worst-case error in regions where the source domain is noninformative.
Contrary to the first bound and those of the previous sections, the divergence is not an  additive term, but is a factor that weights the importance of the source information. These analyses were combined with PAC-Bayesian generalization bounds of Section~\ref{chap:sota}, and involved an additional term that measures the deviation of the learned majority vote to the {\it a-priori} knowledge we have on the majority vote.

\section{Domain adaptation theory based on algorithmic properties}
\label{chap:algorithmic}
In this section, we first review the work of \cite{MansourS14}, where they derived a domain adaptation generalization bound in terms of the algorithmic robustness of \cite{XuM10} recalled in Section~\ref{chap:sota}. Then, we present the works of \cite{KuzborskijO13} based on a closely related concept of algorithmic stability. Note that this last contribution is proved for a setting different from the domain adaptation problem considered so far, as in this case there is no access to the source examples, but rather to a hypothesis learned from them.

\subsection{Robust domain adaptation}
\paragraph{Definition of \texorpdfstring{$\lambda$-shift}{,}}
\cite{MansourS14} used the concept of algorithmic robustness~\cite{XuM10} to define the \mbox{$\lambda$-shift} that encodes prior knowledge of the deviation between the source and target domains.
The goal of their definition was to capture the proximity of the loss associated to a hypothesis on the source and target domains in the regions defined by partitioning the joint space $\X\times\Y$. As there is usually no access to target labels, the authors proposed to consider the conditional distribution of the label in a given region, and the relation to its sampled value over the given labeled sample $S$. To proceed, let $\rho$ be a distribution over the label space $\Y$, and let $\sigma^y$ and $\sigma^{-y} = 1 - \sigma^y$ denote the probability of a given label $y\in\Y$ and the total probability of the other labels, respectively. The definition of the $\lambda$-shift is then given as follows.
\begin{definition}[\cite{MansourS14}]
\label{def:lambda-shift}
Let $\sigma$ and $\rho$ be two distributions over $\Y$.
$\rho$ is the $\lambda$-shift with respect to $\sigma$, denoted by $\rho\in \lambda(\sigma)$, if for all $y\in \Y$ we have $\rho^y\leq \sigma^y + \lambda\sigma^{-y}$ and $\rho^y\geq \sigma^y(1-\lambda)$. If for some $y\in \Y$ we have $\rho^y = \sigma^y + \lambda \sigma^{-y}$, we say that $\rho$ is \mbox{strict-$\lambda$-shift} with respect to $\sigma$.
\end{definition}
Note that, for the sake of simplicity, for $\rho\in\lambda(\sigma)$, the upper bound and the lower bound of the probability $\rho^y$ are respectively denoted by:
$$\bar{\lambda}^y(\sigma) = \sigma^y + \lambda(1-\sigma^y),\qquad\mbox{and}\qquad\underline{\lambda}^y(\sigma) = \sigma^y (1-\lambda)\, .$$

The above definition means that \mbox{$\lambda$-shift} between two distributions on $\Y$ implies a restriction on the deviation between the probability of a label on the distributions: this shift might be at most a $\lambda$ portion of the probability of the other labels or of the probability of the label. 
Note that $\lambda=1$, respectively $\lambda=0$, corresponds to the no restriction and the total restriction cases, respectively.

\paragraph{Learning bounds based on algorithmic robustness}
To analyze the domain adaptation setting, the authors assumed that $\X\times\Y$ can be partitioned into $M$ disjoint subsets, defined as $\X\times\Y = \bigcup_{i,j} \X_i\times\Y_j$, where the input space is partitioned as $\X=\bigcup_{i=1}^{M_{\X}}$, and the output space as $\Y=\bigcup_{j=1}^{M_\Y} \Y_j$ and $M=M_{\X}M_{\Y}$.
Note that, an \mbox{$(M,\epsilon)$-robust} algorithm outputs a hypothesis that has an $\epsilon$ variation in the loss in each region $\X_i\times\Y_j$. We now present the following theorem.
\begin{theorem}[\cite{MansourS14}]
Let $\cal A$ be an \mbox{$(M,\epsilon)$-robust} algorithm with respect to a loss function $\loss:\X\times\Y$, such that $0 \leq \loss(h(\xbf,y) \leq M_l$, for all $(\xbf,y) \in (\X \times \Y)$ and $h\in\Hcal$.
If $\source$ is \mbox{$\lambda$-shift} of $\target$ with respect to the partition of $\X$ for any $\delta \in (0,1]$, the following bound holds with probability of at least $1-\delta$, over the random draw of the sample $S$ from $\source$, and of the sample $T$ from $\target$ of size $m$,
\begin{align*}
\forall h\in \Hcal,\  \risk^\loss_{\target}(h)\ \leq \ \sum_{i=1}^{M_\xbf} T(\X_i) \loss_S^\lambda(h,\X_i) + \epsilon + M_\loss\sqrt{\frac{2M\ln 2 + 2 \ln \frac1\delta}{m}} \, ,
\end{align*}
where  $T(\X_i) = \frac1m \big|\left\{\xbf \in T\cap\X_i\right\}\big|$ is the ratio of target points in the region $\X_i$, and
$$\forall i \in\{1,\ldots,M_{\X}\},\quad \loss_S^\lambda(h,\X_i) \leq \max_{y\in\Y}\left\{ \loss_i(h,y) \bar{\lambda}^y(\source_i) + \sum_{y'\ne y}  \loss_i(h,y') \underline{\lambda}^{y'}(\source_i) \right\}, $$
with 
$$\loss_i(h,y) = \left\{\begin{array}{lr}
\max_{\xbf\in S\cap \X_i\times y} \loss(h(\xbf),y)&\quad\mbox{if $S\cap\X_i\times y\neq\emptyset$}\\
M_\loss&\quad\mbox{otherwise.}
\end{array}\right.$$
\end{theorem}

The main difference between this domain adaptation result and the original robustness bound of Theorem~\ref{theo:robustness} of Section~\ref{chap:sota} is seen in the first term.
In the latter case, which is an upper bound on the source risk, the first term $\frac{1}{m} \sum_{(x,y)\in S} \loss(h_S(x),y)$ simply corresponds to the empirical error of the model learned on the source sample.
In the former bound, which upper-bounds the target risk, the first term $\sum_{i=1}^{M_\xbf} T(\X_i) \loss_S^\lambda(h,\X_i)$ depends also on the empirical risk on the source sample, which is a combination of the \mbox{$\lambda$-shifted} source risk of each region weighted by the ratio of target points in the region. This is reminiscent of the multiplicative dependence between the source error and the divergence term already mentioned in previous sections. 

\subsection{Hypothesis transfer learning}
\label{sec:HTL}
In this section we review theoretical results for the {\it hypothesis transfer learning (HTL)} setting where only a
hypothesis learned in the source domain, and not the source
(labeled) data, is available in addition to a {\it small} training sample from the target domain. As a direct consequence of this, HTL does not introduce any assumptions about the relatedness of the source and target distributions, and it has an advantage in that it avoids the need to store abundant source data. 

More formally, let $h_{\text{src}}\in\Hcal_\source$ be a hypothesis learned from labeled source data, and let $T=\{(\xbf_i,y_i)\}_{i=1}^m\sim(\target)^m$ be a (labeled) target sample. The goal of HTL is then to learn a target model using $h_{\text{src}}$ and $T$ that is better than the one we can learn from $T$ only. This goal is formalized using the following definition of a HTL algorithm $\cal A$:
$$
\Acal : (\X\times\Y)^m \times \Hcal_\source \mapsto \Hcal\, ,
$$
where $\Acal$ maps any (labeled) target sample $T\sim(\target)^m$ and a source hypothesis $h_{\text{src}}\in\Hcal_\source$ onto a target hypothesis $h\in\Hcal$. We now use this formalization to present several key definitions in HTL.
\begin{definition}[Usefulness and Collaboration~\cite{Kuzborskij18}]
  A hypothesis $h_{\text{src}}\in\Hcal_\source$ is useful for $\Acal$ with respect to the distribution $\source$ and a training sample $S$ of size $m$ if
$$
\esp{S\sim(\source)^m}[\risk_{\D}(\Acal(S,h_{\text{src}}))] < \esp{S\sim(\source)^m} [\risk_{\D}(\Acal(S,\mathbf{0}))].
$$
A hypothesis $h_{\text{src}}\in\Hcal_\source$ and a distribution $\D$ collaborate \cite{BenDavidU13workshop} for $\Acal$, with respect to a training sample $S$ of size
$m$, if
$$
\esp{S\sim(\source)^m}[\risk_{\source}(\Acal(S,h_{\text{src}}))]< \min \left\{\risk_{\source}(\Acal(\emptyset,h_{\text{src}}),\esp{S\sim(\source)^m} [\risk_{\D}(\Acal(S,\mathbf{0}))]\right\}.
$$
\end{definition}
This definition provides two interesting properties for a hypothesis
transfer learning algorithm. The concept of usefulness corresponds to
the case where the algorithm $\Acal$ allows a model to be inferred with a
lower risk by using the source hypothesis. The collaboration refers to
the case where the access to both the source hypothesis $h_{\text{src}}$ and the sample $S$ used together helps to increase the performance in
comparison to the case where they are used separately. If any one of these
two properties is not satisfied, then the resulting learning procedure leads to higher target error. The authors further analyzed a regularized least squares algorithm (RLS) for HTL, as presented below.
\paragraph{A biased RLS algorithm for HTL}
We first begin with a quick recap of the classic RLS algorithm. For a learning sample $T=\{(\xbf_i,y_i)\}_{i=1}^m\sim(\target)^m$ such that $y_i \in [-B,B]$ with $B\in \mathbb{R}$ and $\xbf_i\in
\mathbf{R}^d$ with $\|\xbf\|\leq 1$, the RLS algorithm aims
to solve the following optimization problem:
$$
\min_{\wbf\in\mathbf{R}^d}\left\{\frac{1}{m}\sum_{i=1}^m
(\wbf^T\xbf_i - y_i)^2+\lambda \|\wbf\|^2\right\}.
$$
It is well-known that RLS has useful theoretical properties and its solution can be expressed in a closed form. Now, we consider a source hypothesis of the form $h_{\text{src}}(\xbf)=\xbf^T \wbf_0$, where $\wbf_0$ corresponds to the parameters of $h_{\text{src}}$ in the same space as $\wbf$. 
In \cite{OrabonaCCFS09}, the authors suggested to use a biased regularization with respect to $\wbf_0$, as 
$$\min_{\wbf\in\mathbf{R}^d}\left\{\frac{1}{m}\sum_{i=1}^m
(\wbf^T\xbf_i - y_i)^2+\lambda \|\wbf-\wbf_0\|^2\right\}.$$
In this formulation, we can see that the source
hypothesis represented by $\wbf_0$ acts as a bias that tends to
make the learned model closer to $\wbf_0$ if the learning sample
is compatible with it. Following the result of \cite{KuzborskijO13}, we present a more general version, where the target hypothesis to be learned is defined by
\begin{equation}
\label{eq:HTL_TLS}
h_T(\xbf) = \trunc_C\left(\xbf^\top\hat{\wbf}_T\right) + h_{\text{src}}(\xbf)\,,
\end{equation}
where 
$$
\hat{\wbf}_T = \argmin{\wbf}\frac{1}{m} \sum_{i=1}^m\left(\wbf^\top\xbf_i - y_i + h_{\text{src}}(\xbf_i)\right)^2 + \lambda \|\wbf\|^2\,,
$$
and the truncation function $\trunc_C(a)$ is defined as 
$$
\trunc_C(a)=\min\left[\max\left(a, -C\right), C\right].
$$
This formulation is a generalization of the usual biased RLS algorithm that allows consideration of any type of source model $h_{\text{src}}$. In particular, we can retrieve the usual formulation when $C=\infty$ and  $h_{\text{src}}(\xbf)=\xbf^\top\wbf_0$, where $\wbf_0$ and $\wbf_T$ belong to the same space.

From the theoretical standpoint, the goal of the authors was then to bound the expected
risk associated with this algorithm, in terms of the characteristics of
the source model $h_{\text{src}}$. The proposed result is based upon the {\it leave-one-out} risk over a sample $T$, defined as 
$$\RTemp^{\text{loo}}(\Acal,T)=\frac{1}{m}\sum_{i=1}^m\loss(\Acal_{T^{\backslash  i}},(\xbf_i,y_i))\,,$$
where $\Acal_{T^{\backslash i}}$ represents the model learned by
algorithm $\Acal$ from sample $T$, without the example $(\xbf_i,y_i)$. The first result related to HTL can be now presented in the following theorem.
\begin{theorem}[\cite{KuzborskijO13}]
Set $\lambda \geq \frac{1}{m}$. If $C\geq B +\|h_{\text{src}}\|_\infty$, then
for any hypothesis learned by the algorithm presented in
Equation~\eqref{eq:HTL_TLS}, with probability of at least $1-\delta$
over any sample $T$ of size $m$ {\it i.i.d.} from $\target$, we have
$$
\RT(h_T)-\RTemp^{\text{loo}}(h_T,T)=\mathcal{O}\left(C\frac{\sqrt[4]{\RT(h_{\text{src}})\trunc_{C^2}\left(\frac{\RT(h_{\text{src}})}{\lambda}\right)+\RT^2(h_{\text{src}})}}{\sqrt{m}\delta\lambda^{3/4}}\right).
$$
\noindent If $C=\infty$, then we have
$$
\RT(h_T)-\RTemp^{\text{loo}}(h_T,T)=\mathcal{O}\left(\frac{\sqrt{\RT(h_{\text{src}})}(\|h_{\text{src}}\|_{\infty}+B)}{\sqrt{m}\delta\lambda}\right).
$$
\end{theorem}
According to \cite{Kuzborskij18}, we can draw
the following implications.
\begin{enumerate}
\item For the null source hypothesis, \ie, $h_{\text{src}}=\boldsymbol{0}$, we fall into a classic supervised learning setting, while for $C=\infty$, the generalization bound is bounded by $\mathcal{O}\left(\frac{B}{\sqrt{m}\lambda}\right)$, similar to the results obtained for classic RLS
algorithms~\cite{BousquetE02};
\item If $h_{\text{src}}\neq\boldsymbol{0}$ and $\frac1\lambda\RT(h_{\text{src}})$ tend to zero, then the target true risk converges to the leave-one-out risk. This means that when the source hypothesis is good enough on the target domain, then transfer learning helps to learn a better hypothesis on the target domain, even with small training samples.
\item If $\frac1\lambda \risk(h_{\text{src}})$ is high, then more target labeled data are needed to provide a reliable hypothesis on the target. The domains are then considered to be unrelated, so the source hypothesis does not bring any useful information. 
\end{enumerate}

\paragraph{Multi-source scenario}
Here, we consider the setting of \cite{KuzborskijO17}, where the source hypothesis is expressed as a weighted combination of different source hypotheses
 $$h_{\text{src}}^{\boldsymbol{\beta}}(\xbf)=\sum_{i=1}^n \beta_ih_{\text{src}}^{i}(\xbf),$$
and where the target hypothesis is defined as
$$h_{\boldsymbol{w},\boldsymbol{\beta}}(\xbf)= \langle
\boldsymbol{w},\xbf \rangle + h_{\text{src}}^{\boldsymbol{\beta}}(\xbf).$$
The relevance of the different source hypotheses is then characterized
by their associated weight given by the vector $\boldsymbol{\beta}$.

Let $\ell: \Y\times \Y\rightarrow
\mathbb{R}_+$ be an $H$-smooth loss function, such that $\forall y_1, y_2 \in \Y, \ |\nabla_{y_1}\ell(y_1,y)-\nabla_{y_2}\ell(y_2,y)|\leq H|y_1-y_2|$, and let $\Omega:\mathcal{H}\rightarrow \mathbb{R}_+$ be a
$\sigma$-strongly convex function with respect to a norm
$\|\cdot\|$ and to a hypothesis space $\mathcal{H}$. Given a target training set
$T=\{(\xbf_i,y_i)\}_{i=1}^m$, $\lambda\in\mathbb{R}_+$, $n$ source
hypotheses $\{h_{\text{src}}^{i}\}_{i=1}^n$ and a parameter vector
$\boldsymbol{\beta}$ verifying $\Omega(\boldsymbol{\beta})\leq \rho$,
the transfer algorithm generates a target hypothesis
$h_{\hat{\boldsymbol{w}},\boldsymbol{\beta}}$ such that
\begin{align*}
\hat{\boldsymbol{w}}=\argmin{\wbf\in\mathcal{H}}\left\{\frac{1}{m}\sum_{i=1}^m\ell(\langle \boldsymbol{w},\xbf_i\rangle + h_{\text{src}}^{\boldsymbol{\beta}}(\xbf_i),y_i)+\lambda \Omega(\wbf)\}\right\}.
\end{align*}

In this formulation, the loss function is only minimized with
respect to $\wbf$, and not specifically with respect to
$\boldsymbol{\beta}$. 
However, it is assumed that $
\Omega(\boldsymbol{\beta})\leq \rho$ makes $\boldsymbol{\beta}$ constrained by a
strongly convex function, which allows regularized
algorithms to be covered that consider an additional regularization with respect to
$\boldsymbol{\beta}$. As in the previous analysis, the key quantity $\RT(h_{\text{src}}^{\boldsymbol{\beta}})$ that measures the relevance of the source hypothesis on the target domain will have a crucial role in the analysis of the generalization properties of $h_{\hat{\boldsymbol{w}},\boldsymbol{\beta}}$. To illustrate the types of algorithms covered by this analysis, we can
consider the least-squares-based regularization that given source hypotheses
$\{\wbf_{\text{src}}^i\}\subset \mathcal{H}$, the parameters
$\boldsymbol{\beta}\in\mathbb{R}^n$ and $\lambda\in\mathbb{R}_+$ outputs the target hypothesis
$$
h(\xbf)= \langle
\hat{\boldsymbol{w}},\xbf \rangle \,,
$$
where
\begin{equation}\label{HTL:LSBR}
\hat{\boldsymbol{w}}=\argmin{\wbf\in\mathcal{H}} \left\{\frac{1}{m}\sum_{i=1}^m(\langle
\boldsymbol{w},\xbf_i \rangle - y_i)^2+\lambda\|\wbf-\sum_{j=1}^n\beta_j\wbf_{\text{src}}^j\|_2^2\right\}.
\end{equation}
The problem defined by Equation~\eqref{HTL:LSBR} presents a special case of the classic regularized empirical risk minimization (ERM), and can be interpreted
as the minimization of the empirical error on the target sample while keeping
the solution close to the (best) linear combination of source
hypotheses. Note that while such a formulation is limited to a linear combination of the source hypotheses in the same space as the target predictor, it can be
generalized by allowing the source hypotheses to be treated as "black
box" predictors. The results presented below correspond to generalization bounds for such an RLS multi-source algorithm.
\begin{theorem}[\cite{KuzborskijO17}]\label{th:TLERM}
Let $h_{\hat{\boldsymbol{w}},\boldsymbol{\beta}}$ be a hypothesis output by a regularized ERM algorithm from an $m$-sized training set $T$ {\it i.i.d.} from the target domain $\target$, $n$ source hypotheses
$\{h_{\text{src}}^{i}:\|h_{\text{src}}^{i}\|_{\infty}\leq 1\}_{i=1}^n$, any source weights $\boldsymbol{\beta}$ obeying $\Omega(\boldsymbol{\beta})\leq \rho$ and $\lambda\in\mathbb{R}_+$. 
Assume that the loss is bounded by $M$: $\ell(h_{\hat{\boldsymbol{w}},\boldsymbol{\beta}}(\xbf),y)\leq M$
for any $(\xbf,y)$ and any training set. Then, denoting
$\kappa=\frac{H}{\sigma}$ and assuming that $\lambda\leq \kappa$, we have with probability of at least $1-e^{-\eta}$, for all $\eta\geq 0$ 
\begin{align*}
\RT(h_{\hat{\boldsymbol{w}},\boldsymbol{\beta}}) &\leq\ 
\RTemp(h_{\hat{\boldsymbol{w}},\boldsymbol{\beta}})+\mathcal{O}\left(\frac{\RT^{\text{src}}\kappa}{\sqrt{m}\lambda}+\sqrt{\frac{\RT^{\text{src}}\rho\kappa^2}{m\lambda}}+\frac{M\eta}{m\log\left(1+\sqrt{\frac{M\eta}{u^{\text{src}}}}\right)}\right)\\
&\leq\ \RTemp(h_{\hat{\boldsymbol{w}},\boldsymbol{\beta}})+\mathcal{O}\left(\frac{\kappa}{\sqrt{m}}\left(\frac{\RT^{\text{src}}}{\lambda}+\sqrt{\frac{\RT^{\text{src}}\rho}{\lambda}}\right)+\frac{\kappa}{m}\left(\frac{\sqrt{\RT^{\text{src}}M\eta}}{\lambda}+\sqrt{\frac{\rho}{\lambda}}\right)\right)\,,
\end{align*}
where \mbox{$u^{\text{src}}=\RT^{\text{src}}\left(m+\frac{\kappa\sqrt{m}}{\lambda}\right)+\kappa\sqrt{\frac{\RT^{\text{src}}m\rho}{\lambda}}$} and $\RT^{\text{src}}=\RT(h_{\text{src}}^{\boldsymbol{\beta}})$ is the risk of the
source hypothesis combination.
\end{theorem}
The following conclusions can be drawn from this result.
\begin{enumerate}
\item If $\RT^{\text{src}}$ is high, then $h_{\text{src}}^{\boldsymbol{\beta}}$ has no use for transfer, and would only hurt the performance in the target domain;
\item If $m=\mathcal{O}(1/\RT^{\text{src}})$, then a small value $\RT^{\text{src}}$ allows a faster convergence rate of $\mathcal{O}(\sqrt{\rho}/m\sqrt{\lambda})$ when making use of the information coming from the source hypotheses combination.
\end{enumerate}

\paragraph{Comparison with standard theory of domain adaptation}
Recall that the seminal results presented in Section~\ref{chap:div_based} have the following general form
$$
\RT(h)\leq \RS(h)+ d(\sourceX,\targetX) +\lambda\,,
$$
where $d$ is some divergence between the source and target marginal distributions and $\lambda$ refers to the adaptation capability of the hypothesis class $\mathcal{H}$ from where $h$ is taken. 

In general, domain adaptation bounds cannot be directly compared to the result of Theorem~\ref{th:TLERM}, even though the term $R^{\text{src}}$ can be interpreted as $\hdh$-divergence by defining 
$\mathcal{H}=\{\xbf \mapsto \langle \boldsymbol{\beta},
\boldsymbol{h}_{\text{src}}(\xbf)\ \rangle|\ \Omega(\boldsymbol{\beta})\leq \tau\}$
where 
$\boldsymbol{h}_{\text{src}}(\xbf) =
[h_{\text{src}}^1(\xbf),\ldots,h_{\text{src}}^n(\xbf)]^{\top}$, and fixing $h = h_{\text{src}}^{\boldsymbol{\beta}} \in\mathcal{H}$, such that
$$
R^{\text{src}}=\risk_\target(h_{\text{src}}^{\boldsymbol{\beta}})\leq
\risk_\source(h_{\text{src}}^{\boldsymbol{\beta}}) +\dhdh{\sourceX}{\targetX}+\lambda_\Hcal.
$$
If we insert this inequality into the result presented above, then for any hypothesis
$h$ and $\lambda\leq 1$, and $\rho\leq 1/\lambda$, we have
\begin{equation}
\RT(h)\leq
\RSemp(h)+\mathcal{O}\left(\frac{\risk_\source(h_{\text{src}}^{\boldsymbol{\beta}})
    +\dhdh{\sourceX}{\targetX}+\lambda_{\mathcal{H}}}{\sqrt{m}\lambda}+\frac{1}{m\lambda}\right).
\label{eq:domain adaptation-HTL}
\end{equation}
The two results agree that the divergence between the domains has to be small to generalize well. 
The divergence is actually
controlled by the choice of $\boldsymbol{h}_{\text{src}}$, while the complexity of the hypothesis class $\mathcal{H}$ is controlled by $\tau$. 
In traditional domain adaptation, a hypothesis $h$ performs well on the target domain only if it performs well on the source domain, under the condition that $\mathcal{H}$ is expressive enough to ensure adaptation, or in other words that the $\lambda_\Hcal$ term should be small. 
In HTL, however, this condition can be relaxed, as highlighted by Equation~\eqref{eq:domain adaptation-HTL}, which implies that a good source model has to perform well on its own domain. 
Additionally, while in traditional domain adaptation the $\lambda$-term is assumed to be small -- otherwise there is no hypothesis that can perform well on both domains at the same time, and the adaptation cannot be effective -- in HTL, the transfer can still be beneficial even for large $\lambda$, due to the availability of the labeled target samples.

\subsection{Other relevant contributions}
\paragraph{\cite{li2007bayesian}} The authors of this study investigated HTL from the Bayesian perspective, by proposing a PAC-Bayesian study and deriving bounds that capture the relationship between domains by an additive KL-divergence term, which is classic in a PAC-Bayesian setting. In the particular case of logistic regression, they showed that the divergence term is upper-bounded by $\|h-h_{\text{src}}\|^2$, which motivated the biased regularization term in logistic regression and the interest of incorporating the source hypothesis into the adaptation model.
\paragraph{\cite{MorvantKAIS12}} As in \cite{DhouibR18}, the authors of this paper considered learning with a particular family of similarity functions introduced in \cite{Balcan2008_2}, and provided a generalization bound for this using the algorithmic robustness framework.
\paragraph{\cite{HabrardIJAIT13}} This paper presented a study on iterative self-labeling for domain adaptation, where at each iteration a hypothesis $h$ is learned from the current sample $S$,  some target samples are pseudo-labeled from $T_u$ by $h$, and these are incorporated into the source sample $S$ to progressively modify the current classifier. Their analysis suggested that such a procedure theoretically solves a domain adaptation problem when the hypothesis obtained at each iteration improves upon the hypothesis obtained without self-labeling.
\paragraph{\cite{PerrotH15ICML}} The theoretical results of this paper made use of an extension of the concept of algorithmic stability (see Subsection~\ref{sec:UniformStability}) to similarity learning, and provided generalization bounds for this in the HTL framework presented above. In particular, instead of learning a set of weights $\wbf$ that parameterize the hypothesis function $h$, the authors learned a similarity matrix $\M$ that is regularized with respect to a similarity matrix $\MS$ learned in a related source domain.
\paragraph{\cite{HabrardKAIS16}} In this study, the authors analyzed a setting that consisted of learning $N$ weak hypotheses\footnote{A weak hypothesis for $\D$ is a hypothesis such that $\RD(h)=\frac{1}{2}-\varepsilon$, where $\varepsilon>0$ is a small constant.} using the labeled source sample and reweights them differently by taking into account the data from the unlabeled target domain. Their theoretical analysis proves that the proportion of target examples having a margin $\gamma$ decreases exponentially with the number of iterations, but does not benefit from any generalization guarantees given by an upper bound on the risk with respect to the target distribution.

\paragraph{\cite{simonHTL}} In this study, the authors considered an extension of the original HTL setting through a general form of transfer defined by transformation functions that can be provided as input to the HTL algorithm. These transformation functions include, for instance, the offset transfer and scale transfer, thus generalizing the study of \cite{KuzborskijO13}.

Also, we note that the study of \cite{hanneke19} mentioned in Section 4 analyzed the HTL-based adaptation approach, and showed its efficiency in improving the target performance.

\subsection{Summary}
In this section, we have presented theoretical results that allow algorithmic properties of adaptation algorithms to be taken into consideration. First, we recalled how the algorithmic robustness can be extended to the domain adaptation setting, with relaxation of the covariate-shift assumption. Secondly, we focused on a different domain adaptation setting called hypothesis transfer learning, where there is no access to source samples, but to source model(s) given by the learned hypotheses. In this setting, we presented theoretical results obtained in the case of regularized ERM-based algorithms that rely on the algorithmic stability framework. 

In general, we can highlight several important differences of this framework with respect to the results seen in the previous sections. These are the following:
\begin{enumerate}
    \item Contrary to the divergence-based bounds, the learning guarantees presented in this section do not include a term that measures the discrepancy between the marginal distributions of the two domains. This is as expected, as in the HTL scenario we do not have access to a learning sample from the source domain, but only to a hypothesis learned on it;
    \item The potential success of adaptation in the HTL framework depends on the performance of the source hypothesis on the target distribution, and allows a better hypothesis to be learned, even on small samples when some assumptions are fulfilled;
    \item Contrary to the majority of the results seen so far, the adaptability term is absent from the bounds related to the HTL setting, as in this case, the learner has access to some target labeled data. 
\end{enumerate}

\section{Conclusions and discussion}
\label{chap:conclusions}
In this survey, we have presented an overview of the existing theoretical guarantees that have been proven for the domain adaptation problem, a learning setting that extends traditional learning paradigms to the case where the model is learned and deployed on samples coming from different, yet related, probability distributions. The cited theoretical results often take the shape of learning bounds, where the goal is to relate the error of a model on the training domain (also called the source domain) to that of the test domain (also called the target domain). To this end, we note that the results presented are highly intuitive, as they explicitly introduce the dependence of the relationship between the two errors mentioned above to the similarity of their data-generating probability distributions and that of their corresponding labeling functions. Consequently, this two-way relatedness between the source and target domains characterizes both the unsupervised proximity of two domains, by comparing their marginal distributions, and the possible labelings of their samples, by looking for a good model with a low error with respect to these. This general trade-off is preserved, in one way or another, in the majority of published results on the subject, and thus this can be considered as a cornerstone of modern domain adaptation theory. 

As any survey that gives an overview of a certain scientific field, this {one} would have been incomplete without identification of the problems that remain open. In the context of domain adaptation theory, these problems can be arguably split into two main categories, where the first is related to the domain adaptation problem itself, and the second is related to other learning scenarios similar to domain adaptation. For the first category, one important open problem is that of characterizing the \textit{a-priori} adaptability of the adaptation given by the joint error term. Indeed, this term is often assumed to be small for domain adaptation to be possible, although only one previous study \cite{RedkoHS19} suggested an actual way for its consistent estimation from a handful of labeled target data. On the other hand, domain adaptation has been recently extended to open-set and heterogeneous settings, where for the former both source and target domains are allowed to have nonoverlapping classes, while for the latter the input space of the two domains might differ. To the best of our knowledge, there are still no theoretical results that analyze these scenarios. This point brings us to the second category of open problems related to learning scenarios similar to that of domain adaptation, such few-shot learning problems, where there is the need to learn on a sample that contains no or only a few examples of certain classes appearing in the test data. Intuitively, this problem is tightly related to domain adaptation and might naturally inherit some of its theoretical guarantees, although there have been no studies that make this link explicit in the literature to date.

Finally, this survey has not discussed such closely related topics as multitask learning, learning-to-learn, and lifelong learning, to name but a few. This particular choice was made to remain focused on one particular problem, as this is vast enough on its own. We also admit that there are certainly other relevant papers that provide guarantees for domain adaptation that are not included in this survey\footnote{If your paper does not appear in this survey, but seems relevant to its contents, please let us know, and we will try to include it in the revised versions.}. This field, however, is so large and recent advances have been published at such a great pace that it is simply not possible to keep up with it and to report all possible results, without breaking the general structure and the narrative of our survey. 

\bibliographystyle{apalike}
\bibliography{references}

\end{document}